\documentclass[dvipsnames]{article} % For LaTeX2e
\usepackage{iclr2027_conference,times}

\usepackage{hyperref}
\usepackage{url}
\usepackage{mdframed}

\usepackage{microtype}
\usepackage{graphicx}
\usepackage{subfigure}
\usepackage{booktabs} % for professional tables
\usepackage{enumitem}
\usepackage{amsfonts}
\usepackage{amsmath}
\usepackage{amssymb}
\usepackage{mathtools}
\usepackage{amsthm}
\usepackage{multirow}
\PassOptionsToPackage{dvipsnames}{xcolor}
\usepackage{placeins}
\usepackage{bbm}
\usepackage{array}
\usepackage{stackrel}
\usepackage{algpseudocode}
\usepackage{algorithm}
\usepackage{colortbl}
\usepackage{tabu}
\usepackage{enumitem}
\usepackage{caption}
\usepackage{wrapfig}

\usepackage{fancybox}
\usepackage{makecell}

\usepackage{xspace}
\newcommand\ie{i.\,e.\xspace}
\newcommand\eg{e.\,g.\xspace}
\newcommand\Eg{E.\,g.\xspace}

\newcommand\wlogenerality{w.\,l.\,o.\,g.\xspace}
\newcommand{\E}{\mathbb{E}}
\newcommand{\R}{\mathbb{R}}

\newcommand{\ind}{\perp\!\!\!\perp}

\DeclareMathOperator*{\argmin}{arg\,min}
\newcommand*\diff{\mathop{}\!\mathrm{d}}

\newcommand\op[1]{\mathrm{#1}}  % Operator	

\newcommand{\abs}[1]{\left\lvert #1 \right\rvert}

\newcommand\Var{\op{Var}}

\newcommand{\norm}[1]{\left\lVert#1\right\rVert}

\usepackage{scalerel,stackengine,amsmath}

\usepackage{pifont}

\newcolumntype{P}[1]{>{\centering\arraybackslash}p{#1}}

\newcommand{\green}[1]{\textcolor{ForestGreen}{#1}}

\usepackage{times,tcolorbox}

\usepackage{pifont}
\usepackage{tikz}
\usepackage{graphicx}

\newcommand*\circled[1]{%
\tikz[baseline=(char.base)]{
  \node[shape=circle, draw=NavyBlue!60, fill=NavyBlue!10, thick, inner sep=1pt] (char) {\scriptsize\textsf{#1}};
}}

\newtheorem{definition}{Definition}
\newtheorem{theorem}{Theorem}

\newtheorem{rem}{Remark}

\newtheorem*{examplea}{\textit{Example A}}
\newtheorem*{exampleb}{\textit{Example B}}
\newtheorem*{examplec}{\textit{Example C}}

\newcounter{examplearef}
\newcounter{examplebref}
\newcounter{examplecref}

\newtheorem{innercustomrem}{Remark}
\newenvironment{numrem}[1]
  {\renewcommand\theinnercustomrem{#1}\innercustomrem}
  {\endinnercustomrem}

\newtheorem{innercustomtheorem}{Theorem}
\newenvironment{numtheorem}[1]
  {\renewcommand\theinnercustomtheorem{#1}\innercustomtheorem}
  {\endinnercustomtheorem}

\definecolor{tabblue}{HTML}{2077B4}
\definecolor{tabred}{HTML}{FF0000}
\definecolor{yellow}{HTML}{FFFF88}
\definecolor{orange}{HTML}{FFCC99}

\newtcbox{\myovalbox}{colback=yellow,boxrule=0.1pt,arc=3pt,
  boxsep=0pt,left=0.5pt,right=0.5pt,top=0.5pt,bottom=0.5pt,}

\title{Debiased Prior-Data Fitted Networks for Amortized Bayesian Semi-Parametric Inference}

\author{
     
    \hspace{-8pt} Valentyn Melnychuk\thanks{Equal contribution} , Emil Javurek\footnotemark[1] , \& Stefan Feuerriegel
    
    \setcounter{footnote}{0} \\
    LMU Munich \& Munich Center for Machine Learning \\
    Munich, Germany\\
    \texttt{melnychuk@lmu.de,emil.javurek@lmu.de} \\
}

\iclrfinalcopy
\begin{document}

\maketitle

\begin{abstract}
    Prior-data fitted networks (PFNs) have recently gained popularity as an end-to-end approach to amortized Bayesian inference. Yet, their statistical behavior for semi-parametric targets, such as means under missingness or causal effects, remains poorly understood.  In this work, we develop theoretical foundations for debiased PFNs to perform inference on finite-dimensional targets in general semi-parametric settings. First, we show that, despite being Bayes-optimal with respect to the target, an idealized PFN can exhibit a posterior concentration rate that implicitly depends on the concentration rates of (latent) nuisance functions and can become slower as the prior becomes more diverse. Then, to speed up the target posterior concentration rates, we introduce two generic debiasing strategies based on efficient influence functions: (i)~a label correction and (ii)~a prior correction.
    Finally, we show that the debiased PFNs (i) \& (ii) possess concentration rates that asymptotically match a convergence rate given by traditional frequentist bias-corrected estimators. In this way, we establish a \emph{novel connection between amortized Bayesian inference with PFNs and frequentist semi-parametric efficiency theory}. Under additional assumptions, this allows us to guarantee local and global optimality of PFN-based estimators (namely, the Bernstein--von Mises property and minimax optimality, respectively). Therefore, our work provides guidelines for designing debiased PFNs for semi-parametric inference with principled and provable asymptotic guarantees.  
\end{abstract}

%%%%%%%%%%%%%%%%%%%%%%%%%%%%%%%%%%%%%%%%%%%%%%%%%%%%%%%%%%%%%%%%%%%%%%%%%%%%%%%%%%%%%%%%%%
\vspace{-0.1cm}
\section{Introduction} 
\vspace{-0.1cm}
%%%%%%%%%%%%%%%%%%%%%%%%%%%%%%%%%%%%%%%%%%%%%%%%%%%%%%%%%%%%%%%%%%%%%%%%%%%%%%%%%%%%%%%%%%

% PFNs intro
\emph{Prior-data fitted networks} (PFNs) have emerged as a powerful approach for building tabular foundation models \citep{muller2022transformers,hollmann2023tabpfn,hollmann2025accurate}. PFNs combine pretraining with synthetic data and in-context learning \citep{brown2020language,dong2024survey} to perform an approximate \emph{amortized Bayesian inference} (ABI) \citep{gershman2014amortized,le2017inference}. 
Hence, a PFN (usually a transformer-based neural network) is trained once on datasets sampled from a prior over data-generating mechanisms, and then, at inference time, it maps a new dataset directly into a target posterior distribution. 

% Modern applications of PFNs
PFNs have been applied widely across various machine learning (ML) tasks \citep{muller2025position}. For example, PFNs were adopted to tasks like time-series forecasting \citep{dooley2023forecastpfn, ansari2024chronos,taga2025timepfn}, survival analysis \citep{seletkov2026survival,qi2026survivalpfn,bohm2026survpfn}, Bayesian optimization \citep{muller2023pfns4bo}, and explainability \citep{fonseca2026explainerpfn,brockschmidt2026confoundingshap}, among others. Since recently, PFNs were also used for \emph{semi-parametric inference tasks}\footnote{In this work, we use the term ``semi-parametric inference'' to describe the estimation of a finite-dimensional target in the presence of an infinite-dimensional nuisance component (= nuisance function) \citep{bickel1993efficient}.} such as causal inference \citep{robertson2025pfn,balazadeh2025causalpfn,mahajan2025amortized,ma2026foundation,javurek2026amortizing,reuter2026use,melnychuk2026frequentist}. These developments point toward a broader future role of PFNs as general-purpose engines for automating any statistical inference task.

% Asymptotic stats theory gap for PFNs: main question, what are the concentration rates and are they optimal? Asymptotic properties are important 
Surprisingly, the statistical properties of PFN-based inference remain only partially explored. Some theoretical aspects of the PFNs have been studied \citep{nagler2023statistical,ma2025optimal,nagler2025uncertainty,hulskamp2026uncertainty,ching2026efficient}. However, several core properties of PFN-based inference (and, especially, semi-parametric inference) remain \emph{unclear}:
\begin{mdframed}[linecolor=black!60,linewidth=0.6pt,roundcorner=4pt,backgroundcolor=black!2,innerleftmargin=6pt,innerrightmargin=6pt,innertopmargin=3pt,innerbottommargin=3pt]
    \circled{1} At what rates do na\"ive PFN-based posteriors concentrate around finite-dimensional targets? 
    \\ \circled{2} Can debiasing strategies improve these rates for a specific inference target?
    \\ \circled{3} Do the proposed debiasing strategies yield posterior concentration rates that match the convergence rates of well-adopted debiased frequentist estimators, and are they asymptotically optimal?
\end{mdframed}
\vspace{-0.2cm}
Our work studies the above research questions, and thereby adds to the theoretical foundations for PFN-based inference through the lens of \emph{asymptotic properties}. Our contributions\footnote{Code is available at \mbox{\url{https://anonymous.4open.science/r/Ortho-PFN-Anonymized-E4B7}}.} are three-fold:

\circled{1}~We start by restating that a PFN aims to exactly recover the Bayes-optimal posterior wrt. its prior \citep{nagler2023statistical}. Importantly, however, such Bayes optimality \emph{does not by itself guarantee favorable concentration properties} for a particular finite-dimensional target. We show that the corresponding target posterior concentration rate is governed by two closely related features of the underlying semi-parametric inference task. First, even though a PFN can directly output a posterior for the finite-dimensional target, the concentration rate implicitly depends on the posterior concentration rates of (latent) nuisance functions. 
Second, informally, these rates can become slower as the prior becomes more diverse.

\begin{figure}
    \vspace{-0.4cm}
    \setlength{\fboxsep}{0pt}
    \centering
    \includegraphics[width=\linewidth]{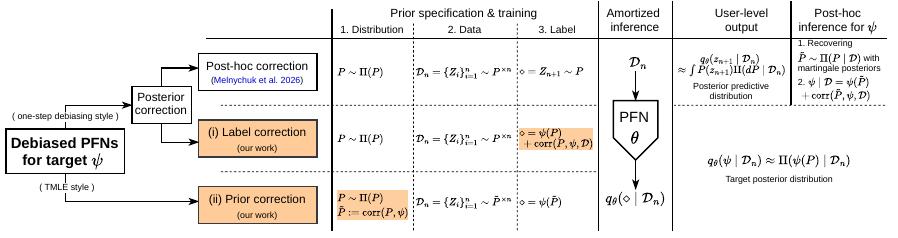}
    \vspace{-0.7cm}
    \caption{\textbf{Overview of our proposed debiasing strategies:} (i) label correction and (ii) prior correction (marked with \colorbox{orange}{orange}). Compared to the existing approach of \citet{melnychuk2026frequentist}, our debiasing strategies (i) and (ii) are applied at the prior specification \& training stages of PFNs and, therefore, do \emph{not} require post-hoc inference to be performed by a downstream user.}
    \label{fig:visual-abstract}
    \vspace{-0.4cm}
\end{figure}

% (2) Techniques to speed up the concentration rates: (i) prior correction & (ii) target label correction
\circled{2}~Then, to speed up the target posterior concentration rate, we carefully tailor two generic debiasing strategies \citep{ray2020semiparametric,yiu2025semiparametric} specifically for PFNs (see Fig.~\ref{fig:visual-abstract}): (i)~a \emph{one-step posterior label correction} that modifies the PFN's target label using an efficient influence function; and (ii)~a \emph{prior correction along a least-favorable direction} that alters the PFN's prior in a target-specific direction. Importantly, unlike the post-hoc approach of \citet{melnychuk2026frequentist}, our strategies (i)--(ii) are applied at the pretraining stage and thus do \emph{not} introduce additional correction steps at test time.

% (3) Properties of the debiased PFNs
\circled{3}~Finally, we show that the concentration rates of the resulting debiased PFNs asymptotically match the convergence rates achieved by well-established frequentist bias-corrected estimators. Hence, the corresponding target posteriors also concentrate at the same rates. This allows us to establish a \textbf{direct connection between PFN-based ABI and frequentist semi-parametric efficiency theory}. Furthermore, under additional assumptions, our debiasing strategies ensure local and global optimality guarantees in the form of semi-parametric Bernstein--von Mises property \citep{freedman1999wald} and minimax optimality \citep{balakrishnan2026fundamental}, respectively.

% Running examples: (a) sample mean; (b) sample mean under missingness; (c) WATE 
To instantiate our theory, we later use three running examples of inferential targets: (A)~\emph{mean value} (classical statistics setting), (B)~\emph{mean value under missingness at random (MAR)} (prediction-powered inference setting \citep{angelopoulos2023prediction,van2026calibeating}), and (C)~\emph{weighted average treatment effect} (WATE) (causal inference setting \citep{hirano2003efficient}).

In sum, our main contribution is providing \emph{guidelines for debiasing PFNs} for semi-parametric inference tasks with principled and provable asymptotic properties.  

%%%%%%%%%%%%%%%%%%%%%%%%%%%%%%%%%%%%%%%%%%%%%%%%%%%%%%%%%%%%%%%%%%%%%%%%%%%%%%%%%%%%%%%%%%
\vspace{-0.1cm}
\section{Related Work} \label{sec:related-work}
\vspace{-0.1cm}
%%%%%%%%%%%%%%%%%%%%%%%%%%%%%%%%%%%%%%%%%%%%%%%%%%%%%%%%%%%%%%%%%%%%%%%%%%%%%%%%%%%%%%%%%%

In the following, we discuss existing works on (a) the statistical properties of PFNs and (b) semi-parametric inference with PFNs. In Appendix~\ref{app:extended-rw}, we provide further details on related works.   

\textbf{Statistical properties of PFNs.} \citet{nagler2023statistical} has laid general statistical foundations of PFNs and argued whether transformers are suitable for consistent ABI. Multiple subsequent works then studied downstream aspects of the pretrained PFNs: \eg, the application of PFNs for downstream inference \citep{nagler2025uncertainty,ng2025tabmgp}, and the error decomposition of PFN predictions \citep{hulskamp2026uncertainty,fortini2026principled}. Other works explored fundamental limits of PFNs pretraining itself: \citet{kim2024transformers,ching2026efficient} showed that transformer-based PFNs can in principle achieve minimax rates for H\"older smooth functions; \citet{ma2025optimal} demonstrated the adaptivity of PFN wrt. different smoothness levels; and \citet{rueter2025can} empirically showed the learnability of latent variables with PFNs. Our work falls into the latter category: We establish the asymptotic properties of \emph{idealized PFNs used as semi-parametric inference engines}.  

\textbf{Semi-parametric inference with PFNs.} Recent work has adapted PFNs to causal inference \citep{balazadeh2025causalpfn,robertson2025pfn,ma2026foundation,javurek2026amortizing}; yet, these works primarily fit PFNs in a na\"ive way and, at best, only establish estimation consistency (without further discussing the asymptotic properties). The closest to our theoretical work is \citet{melnychuk2026frequentist}, who show that na\"ive PFN-based inference for the ATE can suffer from prior-induced confounding bias \citep{hahn2018regularization,linero2023and} and who apply a \emph{post-hoc} one-step posterior correction. Our work differs from the former: We characterize posterior concentration rates for general finite-dimensional semi-parametric targets, and we incorporate debiasing \emph{directly} into PFN pretraining, thereby avoiding \emph{post-hoc} correction at test time. \textbf{To the best of our knowledge, our work is the first to offer theoretical foundations for PFNs used in semi-parametric inference}.
% \textbf{Statistical inference with PFNs.}

%%%%%%%%%%%%%%%%%%%%%%%%%%%%%%%%%%%%%%%%%%%%%%%%%%%%%%%%%%%%%%%%%%%%%%%%%%%%%%%%%%%%%%%%%%
\vspace{-0.1cm}
\section{Preliminaries} \label{sec:preliminaries}
\vspace{-0.1cm}
%%%%%%%%%%%%%%%%%%%%%%%%%%%%%%%%%%%%%%%%%%%%%%%%%%%%%%%%%%%%%%%%%%%%%%%%%%%%%%%%%%%%%%%%%%

\textbf{Notation.} Let $Z \in \mathcal{Z} \subseteq \R^d$ denote a random variable with probability distribution $P = P(Z) \in \mathcal{P}(\mathcal{Z})$, where $\mathcal{P}(\mathcal{Z})$ is a class of probability distributions defined on $\mathcal{Z}$. With a slight abuse of notation, we also denote the density/probability mass function of observing $Z=z$ as $P(z)$. For $P\in\mathcal{P}(\mathcal{Z})$, we observe an i.i.d. dataset $\mathcal{D}_n=(Z_1,\ldots,Z_n)\sim P^{\otimes n}$. Similarly, we denote processes (distributions in the infinite-dimensional/functional spaces) with $\Pi$: \eg, $\Pi(P) $ is a prior process and $\Pi(P \mid \mathcal{D}_n)$ is a posterior process over the distributions of $Z$. Also, for any $P$-measurable function $f$, we write an $L_p(P)$-norm of $f$ as $\norm{f}_{L_p(P)} :=(\E_P[\abs{f(Z)}^p])^{1/p}$; a mean of $f$ as $Pf:=\E_P[f(Z)]=\int f(z) \diff{P}(z)$; a sample mean of $f$ as $P_nf:=n^{-1}\sum_{i=1}^n f(Z_i)$; and the mean wrt. Bayesian bootstrap process as $\operatorname{BB}_n\{f(Z)\}=\sum_{i=1}^n W_if(z_i)$ where $(W_1, \dots, W_n)$ is a sample from a Dirichlet distribution $\operatorname{Dir}(n; 1, \dots,1)$. Similarly, we also define $L_p$-norm for a function $f$ as $\norm{f}_{L_p}:= (\int \abs{f(z)}^p \diff{z})^{1/p}$. For a sequence of random vectors $X_n$ and a positive deterministic sequence $a_n$, we write $X_n=o_{P}(a_n)$ if $\lVert X_n\rVert_{2}/a_n\to 0$ in $P^{\otimes n}$-probability, and $X_n=O_{P}(a_n)$ if $\lVert X_n\rVert_2/a_n$ is bounded in $P^{\otimes n}$-probability. Finally, $a \lesssim b$ means that $a \le C \cdot b$ for some $C \ge 0$.

\textbf{Problem setup.} In our work, we are interested in a statistical inference for a finite-dimensional statistical functional (estimand) $\psi:\mathcal{P}(\mathcal{Z})\to \Psi \subseteq \mathbb{R}$ (\wlogenerality $\psi(P)$ is one-dimensional, and $\Psi$ is convex). Let $\eta(P)\in\mathcal H$ denote a collection of nuisance functions, that is, components of $P$ used for inference on $\psi(P)$, where $\mathcal H$ is a function space equipped with norm $\norm{\cdot}_{\mathcal H}$ \citep{kennedy2024semiparametric}. For a fixed distribution $P_0\in\mathcal{P}(\mathcal{Z})$, we write $\psi_0:=\psi(P_0)$ as the target statistical estimand. When $\psi(P)$ is pathwise differentiable at $P$, we denote its \emph{efficient influence function} (EIF) under the nonparametric model by $\phi_\psi(Z; P)$ (see Appendix~\ref{app:background} for details). 
Importantly, pathwise-differentiable estimands $\psi(P)$ admit a \emph{von--Mises expansion}:
\begingroup\makeatletter\def\f@size{9}\check@mathfonts
\begin{equation} \label{eq:von-mises}
    \psi(P)-\psi(P_0)= (P - P_0)\phi_\psi(\cdot; P) + R_{2, \psi}(P, P_0) = - P_0\phi_\psi(\cdot; P) + R_{2, \psi}(P, P_0), 
\end{equation}
\endgroup
where $R_{2, \psi}(P, P_0)$ is the second-order remainder term. 

\textbf{Inferential targets.} In our work, we consider the following three running examples of estimands:

\begin{examplea}[Mean value] \refstepcounter{examplearef}\label{ex:a}
    We start with a trivial estimand,\footnote{While estimation of the mean does not involve nuisance functions, we consider it as the simplest running example that illustrates the main ideas before turning to genuinely semi-parametric settings.} a mean value of $Z \in [0, 1]; \psi(P) = \E_P[Z]= \int z P(z) \diff{z}$. Then, the corresponding nuisance functions, the EIF, and the second-order remainder are 
    \begingroup\makeatletter\def\f@size{9}\check@mathfonts
    \begin{equation}
        \eta(P) = [P(z)]; \,\, \phi_\psi(Z; \green{P}) = Z - \psi(\green{P}); \,\, R_{2, \psi}(P, P_0) =0.
    \end{equation}    
    \endgroup
\end{examplea}

\begin{exampleb}[Mean value under missingness at random (MAR)] \refstepcounter{examplebref}\label{ex:b}
    Mean value under MAR is a classical semi-parametric target \citep{angelopoulos2023prediction,van2026calibeating}. Here, $ Z = (X, R, RY)$ with covariates $X \in [0, 1]^{d_x}$, labeling indicator $ R \in \{0,1\}$, and the (partially missing) outcome $RY \in [0, 1]$. Here, the inference target is $\psi = \E_P[Y]$. Under standard MAR assumptions (\ie, conditional independence, $R \ind Y \mid X$; and strong overlap, $\kappa < \pi(x) < 1 - \kappa$, with $\kappa \in (0, 1/2)$), $\psi(P) = \E_P[\mu(X)] = \int \mu(x) P(x) \diff x$, where $\mu(x) = \E_P[RY \mid X=x, R=1]$ is a conditional outcome mean, and $\pi(x) = \mathbb{P}(R=1 \mid X =x)$ is a labeling score. Then, the corresponding nuisance functions, the EIF, and the second-order remainder are \citep{robins1995analysis}:
    \begingroup\makeatletter\def\f@size{8}\check@mathfonts
    \begin{equation}
        \eta(P) = [\mu(x), P(x)]; \,\, \phi_\psi(Z; \green{P}) = \frac{R}{\green{\pi(}X\green{)}}(Y-\green{\mu(}X\green{)}) + \green{\mu(}X\green{)} - \psi(\green{P}); \,\, R_{2, \psi}(P, P_0) = P_0 \bigg[\frac{\pi - \pi_0}{\pi} (\mu - \mu_0) \bigg]. 
    \end{equation}
    \endgroup
\end{exampleb}

\begin{examplec}[Weighted average treatment effect (WATE)] \refstepcounter{examplecref}\label{ex:c}
    Let $Z=(X,A,Y)$, with propensity score $\pi(x)=\mathbb{P}(A=1\mid X=x)$ and outcome regressions $\mu_a(x)=\E_P[Y\mid A=a,X=x]$. For a user-specified weight $w(\cdot;\pi)\geq0$, the WATE is $\psi_w(P)=\E_P[w(X;\pi)(\mu_1(X)-\mu_0(X))]/\E_P[w(X;\pi)]$, which includes the ATE $(w\equiv1)$ and overlap-weighted ATE $(w=\pi(1-\pi))$ as special cases.
    We defer further details to Appendix~\ref{app:ex-c}.
\end{examplec}

\textbf{Amortized Bayesian inference with PFNs.} A PFN amortizes inference across synthetically generated datasets. Specifically, for a fixed dataset $\mathcal{D}_n$, let $q_\theta(z_{n+1}\mid \mathcal{D}_n)$ denote a conditional density on $\mathcal{Z}$ parameterized by a neural network with parameters $\theta \in \Theta$ (typically, a transformer \citep{vaswani2017attention}). Then, a \textbf{posterior-predictive PFN} \citep{muller2022transformers} \emph{with fixed $n$} aims to minimize a prior-predictive negative log-likelihood (NLL) risk wrt. a new observation $Z_{n+1} \sim P$:
\vspace{-0.1cm}
\begingroup\makeatletter\def\f@size{9}\check@mathfonts
{\setlength{\belowdisplayskip}{0.1cm}
\begin{align} \label{eq:pfn-nll-ppd}
    \mathcal{R}^{\text{PP}}_{\mathrm{NLL}}(\theta; \Pi,n) := \E_{P\sim\Pi} \E_{\mathcal{D}_n\sim P^{\otimes n}} \E_{Z_{n+1} \sim P}\left[ -\log q_\theta\big(Z_{n+1}\mid \mathcal{D}_n\big)\right],
\end{align}}
\endgroup
where $\Pi = \Pi(P)$ is a prior process over the distributions $P$. 

\textbf{Bayesian self-consistency.} The main reason why the prior-predictive risk in Eq.~\eqref{eq:pfn-nll-ppd} performs ABI is the property of \emph{Bayesian self-consistency} \citep{dawid1982well}: the conditional distribution of a held-out observation $Z_{n+1}$ generated under the prior process $\Pi$, given the observed dataset $\mathcal{D}_n$, coincides with the Bayesian posterior-predictive distribution. More formally, a PFN $q_{\theta^*}(\cdot\mid \mathcal{D}_n)$ that minimizes the population risk in Eq.~\eqref{eq:pfn-nll-ppd} also minimizes a Kullback-Leibler (KL) divergence wrt. a ground-truth posterior-predictive density $\pi(z_{n+1} \mid \mathcal{D}_n):=\int P(z_{n+1}) \Pi(\diff P \mid \mathcal{D}_n)$ \citep{nagler2023statistical}:
\begingroup\makeatletter\def\f@size{9}\check@mathfonts
\vspace{-0.1cm}
{\setlength{\belowdisplayskip}{0.1cm} \begin{align}
    \theta^* = \argmin_{\theta \in \Theta} \mathcal{R}^{\text{PP}}_{\mathrm{NLL}}(\theta; \Pi,n) = \argmin_{\theta \in \Theta} \E_{P\sim\Pi} \E_{\mathcal{D}_n\sim P^{\otimes n}} \operatorname{KL}[\pi(\cdot \mid \mathcal{D}_n) \, || \, q_\theta(\cdot \mid \mathcal{D}_n)].
\end{align}}
\endgroup
Therefore, the statistical properties of PFN-based ABI are mainly guided by (a)~the specification of the prior process $\Pi$ (which, in turn, specifies the prior $\Pi(P)$, the implicit log-likelihood $\ell_n(P):=\log \prod_{i=1}^n P(Z_i)$, and, thus, the posterior $\Pi( \diff P \mid \mathcal{D}_n) \propto \exp\{\ell_n(P)\} \Pi(\diff P)$) and (b)~the amortization errors due to limited expressivity of the PFN $q_\theta$. In the following, we primarily focus on the aspect (a).

%%%%%%%%%%%%%%%%%%%%%%%%%%%%%%%%%%%%%%%%%%%%%%%%%%%%%%%%%%%%%%%%%%%%%%%%%%%%%%%%%%%%%%%%%%
\vspace{-0.1cm}
\section{Semi-Parametric Inference with PFNs} 
\vspace{-0.1cm}
%%%%%%%%%%%%%%%%%%%%%%%%%%%%%%%%%%%%%%%%%%%%%%%%%%%%%%%%%%%%%%%%%%%%%%%%%%%%%%%%%%%%%%%%%%

In the following, we develop theoretical foundations for amortized Bayesian semi-parametric inference with PFNs along our three research questions: \circled{1}~In Sec.~\ref{subsec:slow-rates}, we first show that, despite being Bayes-optimal, an idealized PFN can suffer from slow posterior concentration rates. \circled{2}~~In Sec.~\ref{subsec:debiasing-strategies}, we propose two remedies against the slow concentration rates of PFNs: (i)~label correction and (ii)~prior correction. \circled{3}~In Sec.~\ref{subsec:debiased-theory}, we formalize the favorable asymptotic properties of the debiased PFNs. We provide detailed theoretical statements and all the proofs in Appendix~\ref{app:theory}.

\vspace{-0.1cm}
\subsection{Bayes-Optimality and Slow Concentration Rates} \label{subsec:slow-rates}
\vspace{-0.1cm}

The core idea of ABI with PFNs can be straightforwardly applied to semi-parametric inference. Specifically, a \textbf{PFN for target $\mathbf{\psi}$} with fixed $n$ is given by a conditional density $q_\theta(\psi \mid \mathcal{D}_n)$ in the parameter space $\Psi$ that minimizes the following NLL risk wrt. a label $\psi$:
\begingroup\makeatletter\def\f@size{9}\check@mathfonts
\begin{align} \label{eq:pfn-nll-semiparam}
    \mathcal{R}_{\mathrm{NLL}}^\psi(\theta; \Pi,n) := \E_{P\sim\Pi} \E_{\mathcal{D}_n\sim P^{\otimes n}} \left[ -\log q_\theta\big(\psi(P) \mid \mathcal{D}_n\big)\right].
\end{align}
\endgroup
Again, similarly, to the posterior-predictive PFNs (see Sec.~\ref{sec:preliminaries}), the minimization of the risk in Eq.~\eqref{eq:pfn-nll-semiparam} is equivalent to the KL-projection of $q_\theta(\psi \mid \mathcal{D}_n)$ on the ground-truth posterior distribution\footnote{More precisely, $\Pi(\psi(P) \mid \mathcal{D}_n)$ is the marginalization of the full posterior $\Pi(P \mid \mathcal{D}_n)$.} $\Pi(\psi(P) \mid \mathcal{D}_n)$. This yields the following useful definition.

\begin{wrapfigure}{r}{0.54\textwidth}
    \vspace{-0.85cm}
    \centering
    \includegraphics[width=0.97\linewidth]{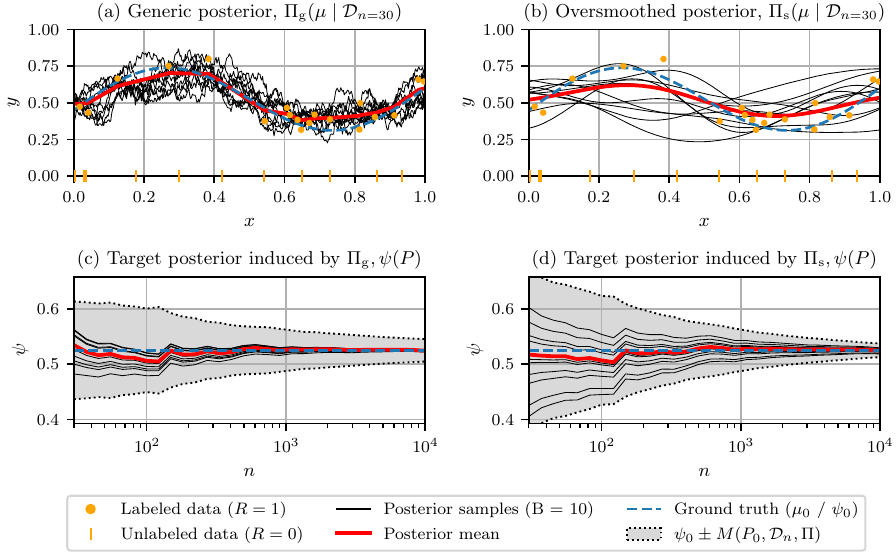}
    \vspace{-0.3cm}
    \caption{\textbf{Propagation of $\mu(x)$ nuisance errors into the target} (based on Example~\ref{ex:b-slow} with $d_x = 1$, $P(x) = P_0(x) = \mathrm{Unif}[0, 1]$, and $\pi_0(x) = 0.2 + {0.7}/({1 + \exp\{-6(x - 0.4)\}})$). Here, \textbf{(a)}~a generic posterior, $\Pi_{\mathrm{g}}(\mu\mid \mathcal{D}_{n=30})$, is a Gaussian process with a Mat\'ern kernel ($\beta = 0.5$). Then, \textbf{(b)}~an oversmoothed posterior, $\Pi_{\mathrm{s}}(\mu \mid \mathcal{D}_{n=30})$, is a Gaussian process with a squared-exponential kernel ($\beta = \infty$), random intercept, and inflated noise level. \textbf{(c)},\,\textbf{(d)}~The corresponding posteriors of $\psi = \int \mu(x) \diff{P_0}(x)$ for growing $n$ (each line follows a fixed posterior quantile across $n$) and the bands $\psi_0 \pm M(P_0, \mathcal{D}_n, \Pi)$. 
    In (d) (unlike (c)), the errors of $\mu(x)$ barely cancel in $P_0(\mu - \mu_0)$, and the target posterior concentrates at nearly the same rate as $M(P_0, \mathcal{D}_n, \Pi)$ (\ie, the oversmoothed posterior is rate-close to the worst-case posterior).
    }
    \label{fig:worst-case-posterior}
     \vspace{-1.28cm}
\end{wrapfigure}

\begin{definition}[Idealized PFN] \label{def:ideal-pfn}
    A PFN $q_{\theta^*}$ for target $\psi$ with fixed $n$ is called idealized if $\theta^* = \argmin_{\theta \in \Theta} \mathcal{R}_{\mathrm{NLL}}(\theta; \Pi,n)$ yields $Q_{\theta^*}(\psi \mid \mathcal{D}_n) = \Pi(\psi(P) \mid \mathcal{D}_n)$ for every $\mathcal{D}_n$ from a prior process $\Pi$ (here: $Q_{\theta^*}$ is a distribution corresponding to the density $q_{\theta^*}$). Thus, the PFN class is sufficiently expressive to capture the ground-truth posterior.
\end{definition}

\underline{\textbf{Bayes-optimality.}} Equipped with the definition of idealized PFNs, we can use both the ground-truth posterior and its PFN approximation interchangeably. Also, it allows us to formalize a Bayes decision-theoretic property, \ie, Bayes-optimality, of PFNs for target $\psi$ (see Appendix~\ref{app:background} for details). 

\begin{rem}[Bayes-optimality of the idealized PFN] \label{rem:bayes-optimality}
    The posterior mean of an idealized PFN, i.e., $\hat{\psi}_{\theta^*}(\mathcal{D}_n) := {\int \psi \,  q_{\theta^*}(\psi \mid \mathcal{D}_n) \diff{\psi}}$, is Bayes-optimal under the squared $l_2$ risk $\mathcal{R}_{l_2^2}(\hat{\psi}; \Pi,n) := \E_{P\sim\Pi} \E_{\mathcal{D}_n\sim P^{\otimes n}} [\hat{\psi}(\mathcal{D}_n) - \psi(P)]^2$:
    \begingroup\makeatletter\def\f@size{9}\check@mathfonts
    % \vspace{-0.1cm}
    {\begin{gather}
        \begin{split}
            & \mathcal{R}_{l_2^2}(\hat{\psi}_{\theta^*}; \Pi,n) = \inf_{\hat{\psi}: \mathcal{Z}^{n} \to \Psi} \mathcal{R}_{l_2^2}(\hat{\psi}; \Pi,n) \\
            &= \E_{P\sim\Pi} \E_{\mathcal{D}_n\sim P^{\otimes n}}[\Var_{\Pi}[\psi(P) \mid \mathcal{D}_n]],
        \end{split}
    \end{gather}}
    \endgroup
    where the last term expresses the expected posterior uncertainty about the target.
\end{rem}

Remark~\ref{rem:bayes-optimality} justifies the optimality of $q_{\theta^*}$ from a Bayesian perspective. However, from a frequentist lens, we might be interested in how quickly $\Pi(\psi(P) \mid \mathcal{D})$ and, hence, $q_{\theta^*}(\psi \mid \mathcal{D})$) \emph{concentrate locally around} some \textbf{fixed} $\psi_0 = \psi(P_0)$ sampled from a prior. It turns out that the concentration rates of the target $\psi$ implicitly depend on the concentration rate of the associated nuisance functions $\eta(P)$, even though those nuisance functions are not explicitly modeled by the idealized PFN $q_{\theta^*}$.  

\begin{theorem}[Slow local $L_2$-concentration rates (short version)] \label{thrm:slow-rates}
    Assume that the target $\psi$ admits a von--Mises expansion from Eq.~\eqref{eq:von-mises} wrt. to any pair of $P, P_0$ independently drawn from $\Pi(P)$. Then, for a fixed target $\psi_0 = \psi(P_0)$ where $P_0 \in \operatorname{supp} \Pi $, and $\mathcal{D}_n \sim P_0^{\otimes n}$, the target $L_2$-concentration error, $\norm{\psi - \psi_0}_{L_2(q_{\theta^*}, \mathcal{D}_n)} := ({\int (\psi - \psi_0)^2 q_{\theta^*}(\psi \mid \mathcal{D}_n) \diff{\psi} })^{1/2}$, can be upper-bounded: 
    \begingroup\makeatletter\def\f@size{8}\check@mathfonts
    \vspace{-0.2cm}
    {\setlength{\belowdisplayskip}{-0cm} \begin{align} \label{eq:slow-rates}
        \norm{\psi - \psi_0}_{L_2(q_{\theta^*}, \mathcal{D}_n)} \lesssim \underbrace{\bigg[ \int \norm{\eta(P) - \eta(P_0)}_\mathcal{H}^2 \,  \Pi({\diff P \mid \mathcal{D}_n})\bigg]^{1/2}}_{F_\psi(P_0, \mathcal{D}_n, \Pi)} + \underbrace{\bigg[\int R_{2, \psi}^2 (P, P_0) \, \Pi({ \diff P \mid \mathcal{D}_n}) \bigg]^{1/2}}_{R_{2,\psi}(P_0, \mathcal{D}_n, \Pi)},
    \end{align}}
    \endgroup
    where $F_{\psi}(P_0, \mathcal{D}_n, \Pi)$ and $R_{2,\psi}(P_0, \mathcal{D}_n, \Pi)$ are posterior-averaged first-order error and second-order remainder, respectively.
    Furthermore, assuming the $L_2$-concentration of the latent nuisance posteriors (\ie, $\int \norm{\eta(P) - \eta(P_0)}_{\mathcal{H}}^2 {\Pi(\diff P \mid \mathcal{D}_n)} \to 0$ with $n \to \infty$),  the upper-bound in Eq.~\eqref{eq:slow-rates} is rate-sharp: Unless explicitly corrected, there exist a worst-case posterior $\Pi(P\mid \mathcal{D}_n)$ under which the target $L_2$-concentration error inherits the rate of the latent nuisance functions, namely, the rate of the first-order term $F_{\psi}(P_0, \mathcal{D}_n, \Pi)$.   
\end{theorem}

Theorem~\ref{thrm:slow-rates} shows that, despite the Bayesian optimality of the idealized PFNs, the target $L_2$-concentration error can linearly depend on \textbf{the concentration rate of the latent nuisance functions}.\footnote{Notably, Bayes-optimality does not extend to the squared posterior $L_2$-concentration risk itself. Namely, under the squared $L_2$-concentration risk, the optimal Bayesian model is achieved by a Dirac-delta distribution centered at the posterior mean of the idealized PFN. Nevertheless, the full posterior remains Bayes-optimal under the NLL risk.} Hence, it can also be interpreted as a generalization of the previously discovered issue of \emph{prior-induced bias} \citep{hahn2018regularization} (for a more formal connection, we refer to Remark~\ref{rem:pib} in Appendix~\ref{app:theory}). In our examples, Eq.~\eqref{eq:slow-rates} can be instantiated as follows. 

\begin{examplea}[Mean value]
    For a fixed pair of distributions $P$ and $P_0$, we have the following:
    \begingroup\makeatletter\def\f@size{9}\check@mathfonts
    \vspace{-0.1cm}
    {\setlength{\belowdisplayskip}{0.1cm} \begin{equation}
        \abs{\psi -\psi_0} = \abs{\int_0^1 z \, (P(z) - P_0(z))\diff{z}}  \le \norm{z - 1/2}_{L_2} \cdot \norm{P - P_0}_{L_2(z)} = 1/\sqrt{12} \cdot {\norm{P - P_0}_{L_2(z)}},
    \end{equation}}
    \endgroup
    and, thus,  $\norm{\eta(P) - \eta(P_0)}_\mathcal{H} = \norm{P - P_0}_{L_2(z)}$ is the leading first-order error term. Furthermore, this bound is locally rate-sharp and is attained when the posterior $\Pi(P \mid \mathcal{D}_n)$ assigns asymptotically non-negligible mass to a region around the path $P_t(z)=P_0(z)+t(z-1/2)$ for sufficiently small $t > 0$ (along such a path both nuisance errors and $\abs{\psi(P_t)-\psi_0}$ are of order $t$).
\end{examplea}

\begin{exampleb}[Mean value under MAR] \refstepcounter{examplebref}\label{ex:b-slow}
    Similarly, for a fixed pair of distributions $P$ and $P_0$,
    \begingroup\makeatletter\def\f@size{9}\check@mathfonts
    \vspace{-0.1cm}
    {\setlength{\belowdisplayskip}{0.1cm} \begin{align} \label{eq:ex-b-slow-rates}
        \abs{\psi -\psi_0} & = \abs{(P - P_0)\mu + P_0\Big[\frac{\pi_0}{\pi}(\mu - \mu_0) \Big] + R_{2,\psi}(P, P_0)} \\[-0.4em]
         & \le \norm{\mu}_{L_2(x)} \cdot \norm{P - P_0}_{L_2(x)} + \frac{1}{\kappa} \cdot \norm{\mu - \mu_0}_{L_2(P_0)} + \abs{R_{2,\psi}(P, P_0)},
    \end{align}}
    \endgroup
    and, thus,  $\norm{\eta(P) - \eta(P_0)}_\mathcal{H} = \norm{P - P_0}_{L_2(x)} + \norm{\mu - \mu_0}_{L_2(P_0)}$ is the leading first-order error term. Again, this bound is locally rate-sharp and is attained when the posterior $\Pi(P\mid\mathcal D_n)$ assigns asymptotically non-negligible mass to a region around a path $P_t(x)=P_0(x)+t h_0(x)$, $\mu_t(x)=\mu_0(x)+t$, and $\pi_t(x)=\pi_0(x)$, where $h_0$ is any valid mean-zero direction satisfying $\int \mu_0(x)h_0(x)\diff{x}\neq0$ (along such a path both nuisance errors and $\abs{\psi(P_t)-\psi_0}$ are of order $t$).
\end{exampleb}

\begin{figure}
    \vspace{-0.4cm}
    \centering
    \includegraphics[width=\linewidth]{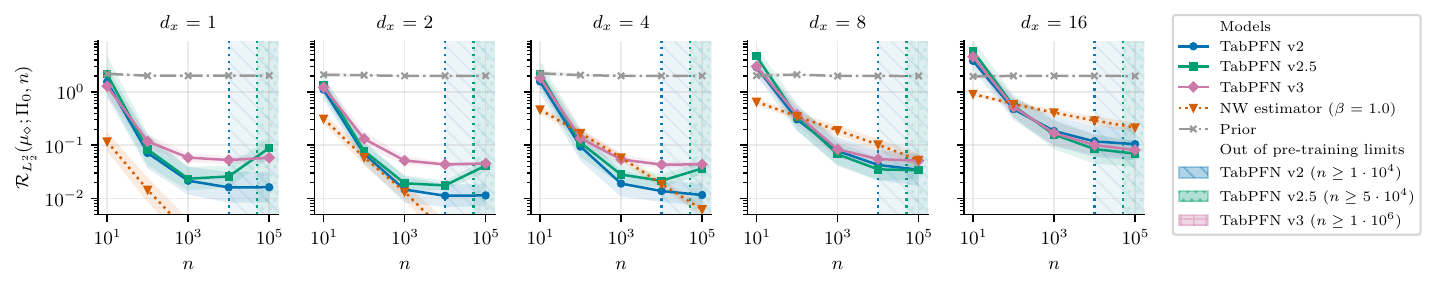} 
    \vspace{-0.8cm}
    \caption{\textbf{Slowing concentration rates for different versions of TabPFN ($\diamond \in \{\mathrm{v2}, \mathrm{v2.5}, \mathrm{v3}\}$)}. Here, we show $\Pi_0$-prior-averaged squared $L_2$-concentration errors of the conditional outcome mean $\mu(x)$ posteriors, {$\mathcal{R}_{L_2^2}(\mu_\diamond; \Pi_0,n) := \E_{P_0\sim\Pi_0} \E_{\mathcal{D}_n\sim P_0^{\otimes n}} [M(P_0, \mathcal{D}_n, \Pi_{\diamond})^2]$}, depending on dataset sizes $n$ and covariate dimensionality $d_x$ (prior-averaging is based on a sample of $M=24$ ground-truth $P_0$, each with $D =1$ dataset; lower is better). For reference, we also show the error of a frequentist non-parametric estimator, namely, the $\Pi_0$-prior-averaged squared $L_2$-error of Nadaraya–Watson regression with a Mat\'ern kernel ($\beta = 1.0$) (= NW estimator ($\beta = 1.0$)). Note that the ground-truth $\Pi_0$ is based on the adapted prior of the TabPFN $\mathrm{v1}$ \citep{hollmann2023tabpfn} ($\mu_0(\cdot) \sim \Pi_{\mathrm{v1}}(\mu) \text{ and }P_0(y\mid x) = N(\mu_0(x), \sigma^2 = 0.3^2)$) and is assumed to be in support of the priors of the higher versions.}
    \label{fig:tabpfn-versions-regression}
    \vspace{-0.4cm}
\end{figure}

The latter two examples are particularly relevant for Bayesian semi-parametric inference: Here, the averaged errors of the conditional expectation ($P_0(\mu_t - \mu_0) = t$) implicitly propagate into the error of the target \textbf{in the same order} $t$. Yet, for a posterior given by a generic non-parametric Bayesian model (and idealized PFNs can be viewed as such),  the squared error $\{P_0(\mu-\mu_0)\}^2$ is only upper-bounded by $\|\mu-\mu_0\|_{L_2(P_0)}^2$, with equality achieved for specific, worst-case posteriors $\mu = \mu_t$. 
Thus, in general, the $L_2$-concentration rate of $\mu$, $M(P_0, \mathcal{D}_n, \Pi)^2 := \int \norm{\mu-\mu_0}_{L_2(P_0(x))}^2 {\Pi}(\diff P \mid \mathcal{D}_n)$, informs us of the \emph{worst-case} $L_2$-concentration rate of target $\psi$ (as we show in Fig.~\ref{fig:worst-case-posterior}).

\underline{\textbf{Slow concentration rates.}} It turns out that, in fully non-parametric Bayesian models, the $L_2$-concentration rate of $\mu$, $M(P_0, \mathcal{D}_n, \Pi)$, can be arbitrarily slow. For example, for many non-parametric Bayesian models with a more diverse prior $\Pi(P)$, the posterior concentration rate slows down\footnote{Yet, this is not true in general: Classical non-parametric Bayesian theory \citep{ghosal2017fundamentals} characterizes posterior contraction through a trade-off between the amount of prior mass assigned to small neighborhoods of the data-generating distribution and the complexity of the effective support of the prior. Still, in our work, we consider the worst-case scenario, namely when the greater prior diversity leads to worse posterior concentration.} (\eg, for Gaussian processes with a rougher Mat\'ern kernel \citep{van2011information}). 

Our findings have also implications for existing PFNs such as TabPFN \citep{hollmann2023tabpfn,hollmann2025accurate,grinsztajn2025tabpfn,grinsztajn2026tabpfn}. In Fig.~\ref{fig:tabpfn-versions-regression}, we empirically show that more recent versions of TabPFN ($\diamond =\mathrm{v2} \to \mathrm{v2.5} \to \mathrm{v3}$) have slower rates of the $L_2$-concentration error, $M(P_0, \mathcal{D}_n, \Pi_{\diamond})$. These insights motivate our debiasing strategies for PFNs, with the aim that \emph{a more diverse prior does not result in much slower target concentration rate}.

\vspace{-0.1cm}
\subsection{Debiasing Strategies} \label{subsec:debiasing-strategies}
\vspace{-0.1cm}

In the following, we introduce two debiasing strategies for PFNs.

\underline{\textbf{(i)~Label correction.}}
We start by tailoring (i)~one-step posterior corrections \citep{yiu2025semiparametric} specifically for PFNs. Unlike \citet{melnychuk2026frequentist}, we aim to perform bias correction not post-hoc but at the \emph{training} stage of the PFN for target $\psi$. For this, we use a key property of Bayesian self-consistency (see Sec.~\ref{sec:preliminaries}): A joint sample from a prior process $(\mathcal{D}_n, \psi(P))$ can be interpreted as a joint sample of random $ \mathcal{D}_n$ and a conditional (posterior) target $\psi(P) \mid \mathcal{D}_n$. Therefore, in the case of the PFNs, the one-step posterior correction takes the form of label $\psi$ correction.  

\begin{mdframed}[linecolor=black!60,linewidth=0.6pt,roundcorner=4pt,backgroundcolor=black!2,innerleftmargin=6pt,innerrightmargin=6pt,innertopmargin=3pt,innerbottommargin=3pt]
    A \textbf{label correction} modifies the risk of a PFN with target $\psi$ by performing a one-step posterior correction wrt. $\psi(P)$: 
    \vspace{-0.2cm}
    \begingroup\makeatletter\def\f@size{8}\check@mathfonts
    {\setlength{\belowdisplayskip}{-0.2cm}
    \begin{align} \label{eq:label-corr}
        \tilde{\mathcal{R}}_{\mathrm{NLL}}^{\psi}(\theta; \Pi,n) := \E_{P\sim\Pi} \E_{\mathcal{D}_n\sim P^{\otimes n}} \E_W \bigg[ -\log q_\theta\big(\tilde{\psi}(P) \mid \mathcal{D}_n \big)\bigg], \,\, \tilde{\psi}(P) := \psi(P) + \underbrace{\operatorname{BB}_n\{{\phi_\psi(Z; P) \}}}_{\text{corr}(P, \psi, \mathcal{D}_n)}, 
    \end{align}}
    \endgroup
    where $W$ are the weights of $\operatorname{BB}_n \{ \cdot \}$, sampled independently of $P$ and $\mathcal{D}_n$.  
\end{mdframed}

\textbf{Construction.} Hence, once the EIF for a target $\psi$ is derived, we automatically get the label correction (given that we can effectively infer the corresponding nuisance functions from the full prior $P \sim \Pi(P)$). Let us now revisit our running examples.

\begin{examplea}[Mean value]
    For the mean value, the label correction effectively replaces the initial label, $\psi(P)$, with the Bayesian bootstrap, thereby removing the first-order error wrt. $P$:
    \begingroup\makeatletter\def\f@size{8}\check@mathfonts
    \vspace{-0.1cm}
    \begin{equation} \label{eq:mean-label-corr}
        \tilde{\psi}(P) = \psi(P) + \operatorname{BB}_n\{Z-\psi(P)\} = \operatorname{BB}_n\{Z\}.
    \end{equation}
    \endgroup
\end{examplea}

\begin{exampleb}[Mean value under MAR] \refstepcounter{examplebref}\label{ex:b-label-corr}
    Similarly, the label correction is:
    \begingroup\makeatletter\def\f@size{8}\check@mathfonts
    \vspace{-0.1cm}
    {\setlength{\belowdisplayskip}{0.1cm} \begin{equation} \label{eq:mar-label-corr}
        \tilde{\psi}(P) = {\psi}(P) + \operatorname{BB}_n\bigg\{{\frac{R}{{\pi(}X{)}}(Y-{\mu(}X{)})} + {{\mu(}X{)} - {\psi}(P)}\bigg\} = \operatorname{BB}_n\bigg\{\frac{R}{{\pi(}X{)}}(Y-{\mu(}X{)}) + {\mu(}X{)}\bigg\}.
    \end{equation}}
    \endgroup
\end{exampleb}

% Interpretation 
\textbf{Interpretation.} In both Eq.~\eqref{eq:mean-label-corr}-\eqref{eq:mar-label-corr}, the label correction effectively replaces the original target label with a Bayesian-bootstrap one-step estimator. This is done by design such that the first-order error term (\ie, $F_\psi(P_0, \mathcal{D}_n, \Pi)$ in Eq.~\eqref{eq:slow-rates}) gets replaced with \emph{the empirical term that concentrates around zero with the parametric rate} $O_{P_0}(n^{-1/2})$.

\underline{\textbf{(ii)~Prior correction.}}
An alternative approach to speed up the posterior concentration rates is an approach of prior corrections in the least-favorable direction \citep{ray2020semiparametric,ray2019debiased,yang2015semiparametric}. Existing constructions are typically expressed through target-specific parameterizations of nuisance functions (\eg, Gaussian process priors for $\mu(x)$ for the mean under MAR). Here, on the other hand, we formulate a general, distribution-level approach, suitable for PFNs. Specifically, we perturb each prior draw $P\sim\Pi(P)$ as follows.

\begin{mdframed}[linecolor=black!60,linewidth=0.6pt,roundcorner=4pt,backgroundcolor=black!2,innerleftmargin=6pt,innerrightmargin=6pt,innertopmargin=3pt,innerbottommargin=3pt]
    A \textbf{prior correction} modifies the prior of a PFN with target $\psi$ by augmenting prior draws $P \sim \Pi(P)$ with an independent perturbation along a least-favorable direction\footnotemark $\operatorname{LF}_{\psi}(P; \lambda)$: 
    \vspace{-0.2cm}
    \begingroup\makeatletter\def\f@size{9}\check@mathfonts
    {\setlength{\belowdisplayskip}{-0cm} \begin{align} \label{eq:prior-corr}
        \tilde P :=& \underbrace{\operatorname{LF}_{\psi}(P; \lambda)}_{_{\text{corr}(P, \psi)}}, \quad 
       \begin{cases} 
            \lambda \sim  N(0, \sigma_n^2) \text{ and } \lambda \ind P, \\
            \operatorname{LF}_{\psi} \text{ is such that } \operatorname{LF}_{\psi}(P; 0) = P \text{ and } \operatorname{S}[\operatorname{LF}_{\psi}(P; \lambda)] \big\lvert_{\lambda=0} = \phi_\psi(\cdot; P),
        \end{cases}
    \end{align}}
    \endgroup
    where $\sigma^2_n > 0$ is such that $n \sigma^2_n \to \infty$; $\operatorname{LF}_{\psi}(\cdot; \lambda): \mathcal{P} \to \mathcal{P}$ is a distributional transformation; and $\operatorname{S}[\cdot]: \mathcal{P} \to L_2(\mathcal{P})$ is a score operator (\ie, for a distribution $P_\lambda$ parametrized by $\lambda$ with density $P(z; \lambda)$, $S[P_\lambda](z) = \frac{\partial}{\partial \lambda} \log P(z; \lambda)$). Furthermore, the draws of $P$ and $\lambda$ jointly induce a \textbf{corrected prior process} $\widetilde{\Pi}(\tilde{P})$, that is assumed to have the same support, $\operatorname{supp} \widetilde{\Pi} = \operatorname{supp} \Pi$. 
\end{mdframed}
\footnotetext{This direction is called \emph{least favorable} because, among local perturbations of the data-generating distribution $P$, it represents the directions in which estimation of $\psi$ is locally hardest. Notably, the least favorable directions should not be confused with the worst-case posteriors from Theorem~\ref{thrm:slow-rates}: Our prior correction enriches the prior along the least-favorable directions to enable bias removal, while the worst-case posteriors can arise when the prior lacks sufficient flexibility in these directions, leaving first-order bias uncorrected.}

\textbf{Construction.} We immediately see that the perturbation $\operatorname{LF}_\psi$ is given as a possibly non-unique solution to the operator equation $\operatorname{LF}_\psi(P; \cdot) = (S[\cdot] \lvert_{\lambda=0} )^{-1} \phi_\psi(\cdot ; P)$. This non-uniqueness gives us a certain freedom in how to perform the correction; for example, we can either correct the full distribution or only the relevant nuisance function components. In the following running examples, we carefully adapt the existing approaches from the TMLE literature \citep{van2006targeted}, specifically for the debiased ABI with PFNs. 

\begin{examplea}[Mean value]
    Here, one of the alternatives for the prior correction is the exponential tilting \citep{diaz2015targeted}. Then, the corrected prior density of $P$ from $P \sim \Pi(P)$ is
    \begingroup\makeatletter\def\f@size{9}\check@mathfonts
    % \vspace{-0.1cm}
    {\setlength{\belowdisplayskip}{0cm} \begin{equation}
        \tilde{P}(z) := {\exp\{ \lambda (z - \psi(P))\}} \big/ {C_\psi(P, \lambda)} \, \cdot P(z), \quad C_\psi(P, \lambda)= P[\exp\{ \lambda (z - \psi(P))\}],
    \end{equation}}
    \endgroup
    where $C_\psi(P, \lambda)$ is a normalization constant, and it is easy to verify that $\tilde{P} = P$ when $\lambda = 0$ and $S[\tilde{P}](Z) \lvert_{\lambda = 0} = Z - \psi(P) = \phi_\psi(Z; P)$. Hence, the corrected prior gives each prior draw an additional degree of freedom, $\lambda$, exactly in the direction that moves its mean (we will explain soon why this helps to reduce the first-order error posterior). 
\end{examplea}

\begin{exampleb}[Mean value under MAR] \refstepcounter{examplebref}\label{ex:b-prior-corr}
    Assume, the density of a prior draw, $P(z)$, factorizes as $P(z) = P(x) \pi(x)^r(1-\pi(x))^{1-r}P(y\mid x,R=1)^r$, and the outcome distribution is heteroscedastic normal $P(y\mid x,R=1) = N(\mu(x); \sigma^2(x))$. Now, the prior correction jointly (namely, with the same $\lambda$) perturbs $P(x)$ and $\mu(x)$ along their corresponding least-favorable directions:
    \begingroup\makeatletter\def\f@size{9}\check@mathfonts
    \vspace{-0.1cm}
    {\setlength{\belowdisplayskip}{0.1cm}\begin{gather}
        \begin{split}
            \tilde{P}(x) &:= {\exp\{ \lambda (\mu(x) - \psi(P))\}} \big/{C_\psi(P, \lambda)} \, \cdot P(x), \quad C_\psi(P, \lambda) = P[\exp\{ \lambda (\mu(x) - \psi(P))\}]; \\
            \tilde{\mu}(x) &:= \mu(x) + \lambda \cdot \sigma^2(x) / \pi(x),
        \end{split}
    \end{gather}}
    \endgroup
    where $C_\psi(P, \lambda)$ is a normalization constant, and, again, it can be easily verified that the correction satisfies the definition in Eq.~\eqref{eq:prior-corr}, in particular, because $S[\tilde{P}](X) \lvert_{\lambda = 0} = \mu(X) - \psi(P)$, $S[\tilde{P}](Y \mid X, R=1) \lvert_{\lambda = 0} = ({Y - \mu(X)}) / {\pi(X)}$, and the full observed data score sums to the EIF (with a factor $R$). Here, the prior correction acts differently wrt. different prior nuisance functions (\ie, it does the exponential tilting wrt. $P(x)$ and the location shift \citep{gruber2010targeted} wrt. $\mu(x)$). 
\end{exampleb}

\begin{figure}[t]
    \vspace{-0.4cm}
    \centering
    \includegraphics[width=\linewidth]{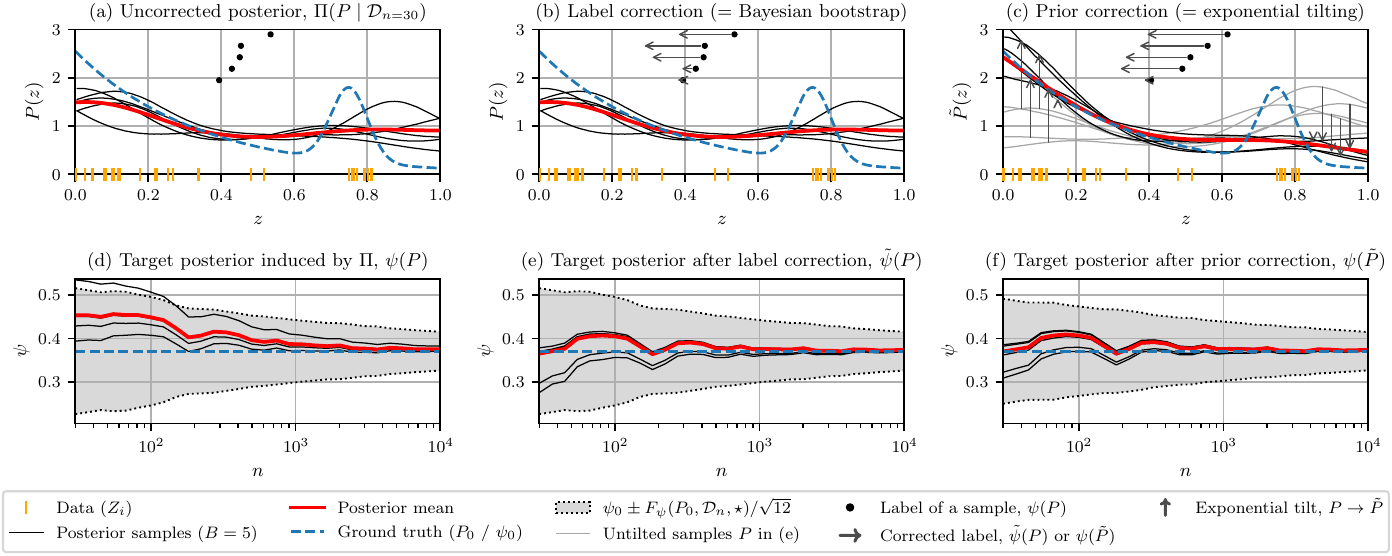}
    \vspace{-0.6cm}
    \caption{\textbf{Debiasing strategies for the mean value} (based on Example~\ref{ex:a} with $Z \in [0, 1]$ and $\psi(P) = \E_P[Z]$). 
    \textbf{(a)}~Posterior draws of $P$ under a non-parametric prior $\Pi$ (a logistic Gaussian process with a squared-exponential kernel ($\beta = \infty$), $P(z) \propto \exp\{f(z)\}$) and the corresponding target labels $\psi(P)$. \textbf{(b)}~(i)~Label correction: The posterior of $P$ is unchanged, but the label of every draw is replaced with $K=1$ sample Bayesian bootstrap (shown with arrows). \textbf{(c)}~(ii)~Prior correction: Posterior draws $\tilde{P}$ under the corrected prior $\widetilde{\Pi}$ (black), each the exponential tilt of an untilted draw $P$ (gray, vertical arrows) that moves its label from $\psi(P)$ to $\psi(\tilde{P})$ (horizontal arrows), 
    \textbf{(d)}-\textbf{(f)}~The resulting posteriors of the target for growing $n$ (each line follows a fixed posterior quantile across $n$) and the bands $\psi_0 \pm F_\psi(P_0, \mathcal{D}_n, \star)/\sqrt{12}$, where $\star \in \{\Pi, \widetilde \Pi \}$. Without a correction, the uncorrected posterior draws pull $\psi$ towards $1/2$. 
    Both corrections remove this first-order error: The posteriors after the corrections are better centered and concentrate at a faster rate around $\psi_0$.
    }
    \label{fig:debiasing-example-a}
     \vspace{-0.4cm}
\end{figure}

% Interpretation
\textbf{Interpretation.} Informally speaking, in all the examples, the prior correction alters the original prior such that it puts enough probability mass along directions that are \emph{locally most relevant for changing the target $\psi$}. This allows to nudge the joint posterior $\widetilde\Pi(P,\lambda\mid\mathcal D_n)$ (and, thus, $\widetilde{\Pi}(\psi(\tilde{P}) \mid \mathcal{D}_n)$) toward what is supported by the data and, thereby, to correct for the first-order error term.
The prior correction also has an important connection to the sharp counterexamples from Theorem~\ref{thrm:slow-rates}: There, local perturbations of the nuisance functions induce a change in $\psi$ of the same order, causing the target posterior to inherit the first-order nuisance error. The prior correction enriches the prior along such target-relevant directions, so that posterior updating can correct this first-order error directly.

\underline{\textbf{Combined strategies.}} Interestingly, both correction strategies aim at the same goal (= reduction of the first-order error), yet they operate in different spaces: While (i)~the label correction works in the target parameter space $\Psi$, (ii)~the prior correction modifies the prior in the distributional space $\mathcal{P}$. Notably, both strategies can be combined: \Eg, in Example~\ref{ex:b-label-corr}, the label correction can be used to correct the marginal distribution $P(x)$, and the prior correction is then applied only wrt. the conditional outcome mean $\mu(x)$. In this way, both our strategies provide a very flexible toolkit to debias PFNs (depending on the ease of implementation, either (i) or (ii) can be chosen for each underlying nuisance function).   

\vspace{-0.1cm}
\subsection{Asymptotic Properties of Debiased PFNs} \label{subsec:debiased-theory}
\vspace{-0.1cm}

Our debiasing strategies allow us to significantly improve the previous results of Theorem~\ref{thrm:slow-rates}. 

\begin{theorem}[$L_2$-concentration rates of debiased PFNs (short version)] \label{thrm:debiased-rates}
    Assume the conditions of Theorem~\ref{thrm:slow-rates} and that either debiasing strategy (i)-(ii) from Sec.~\ref{subsec:debiasing-strategies} is applied. Then, under mild regularity conditions, the $L_2$-concentration error of the debiased PFN, $\tilde q_{\theta^*}$, satisfies 
    \begingroup\makeatletter\def\f@size{9}\check@mathfonts
    \vspace{-0.12cm}
    {\setlength{\belowdisplayskip}{-0cm} \begin{align} \label{eq:debiased-rates}
        \norm{\tilde{\psi}-\psi_0}_{L_2(\tilde q_{\theta^*}, \mathcal{D}_n)} \lesssim \underbrace{\tilde{F}_{\psi}(P_0, \mathcal{D}_n, \star)}_{= O_{P_0}(n^{-1/2})}  + R_{2,\psi}(P_0, \mathcal{D}_n, \star), \quad \star = \begin{cases}
            \Pi \text{, (i) label correction,}\\
            \widetilde \Pi \text{, (ii) prior correction,}
        \end{cases}
    \end{align}}
    \endgroup
    where $R_{2,\psi}(P_0, \mathcal{D}_n, \star)$ is the corresponding posterior-averaged second-order remainder. 
\end{theorem}

Theorem~\ref{thrm:debiased-rates} shows that our debiasing strategies (i) and (ii) effectively replace a potentially slowly concentrating error term, $F_{\psi}(P_0, \mathcal{D}_n, \Pi)$, with the term that concentrates at the fast parametric speed. This finally allows us to establish a strong connection between the debiased ABI with PFNs and the classical frequentist bias-corrected estimators (e.g., A-IPTW / TMLE). As the concentration rate \emph{matches up to the second-order remainder} with the convergence rate of the classical frequentist bias-corrected estimators, we consider two cases: (1)~when the second-order remainder is negligible, and (2)~when it is not. 

\textbf{(1) Second-order remainder is $o_{P_0}(n^{-1/2})$.} In this case, we establish the Bernstein--von Mises property, namely, the asymptotic normality of the posterior given by the debiased PFN $\tilde q_{\theta^*}$. 

\begin{theorem}[Bernstein--von Mises property for debiased PFNs] \label{thrm:bbvm}
    Assume the conditions of Theorem~\ref{thrm:debiased-rates}, $R_{2,\psi}(P_0, \mathcal D_n,\star)=o_{P_0}(n^{-1/2})$, and standard regularity conditions for semi-parametric Bernstein--von Mises convergence. Then, for either debiasing strategy (i) or (ii), the debiased PFN posterior distribution is asymptotically normal and attains the semi-parametric efficiency bound:
    \vspace{-0.2cm}
    \begingroup\makeatletter\def\f@size{9}\check@mathfonts
    {\setlength{\belowdisplayskip}{-0cm} \begin{equation}
        d_{\mathrm{TV}}\left[\widetilde Q_{\theta^*}\left(\sqrt n(\psi-\psi_0) \mid\mathcal D_n\right); N\left(\Delta_n, V_{\text{eff}}(P_0): = \Var_{P_0}\left[\phi_\psi(\cdot;P_0)\right] \right)\right] \xrightarrow{P_0} 0, 
    \end{equation}}
    \endgroup
    where $d_{\mathrm{TV}}$ is total variation, and $\Delta_n := \sqrt{n} P_n [\phi_\psi(\cdot;P_0)]$ is the asymptotically efficient random centering term. 
\end{theorem}

Theorem~\ref{thrm:bbvm} also establishes a \textbf{local optimality of the debiased PFNs} around $P_0$: 
Their rescaled target posteriors converge to a normal distribution with variance equal to the semi-parametric efficiency bound, $V_{\mathrm{eff}}(P_0)$, yielding asymptotically efficient uncertainty quantification.

\textbf{(2) Second-order remainder is non-negligible.} When the second-order remainder concentrates slower than the parametric rate $n^{-1/2}$, its contribution to the posterior is non-negligible, and, in general, the asymptotic shape of the posterior cannot be established. Still, in this case, we can infer the \emph{posterior concentration rates} for each term in Eq.~\eqref{eq:debiased-rates} in Theorem~\ref{thrm:debiased-rates}. This, in turn, allows us to formulate two following global decision-theoretic guarantees for the debiased PFN, i.e., one for $\Pi$-prior-averaged and another for worst-case squared $l_2/L_2$ risks. 

\begin{theorem}[Rate optimality of debiased PFNs] \label{thrm:minimax}
    Assume the conditions of Theorem~\ref{thrm:debiased-rates}, and suppose that, on an effective prior-supported class $\mathcal P_n^\Pi:= \operatorname{supp}(\Pi) \cap \mathcal{P}^{\text{SA}}_n$ with $\Pi(\mathcal P_n^\Pi)\to1$ (where $\mathcal{P}^{\text{SA}}_n$ is a structure-agnostic class of distributions that depends on the nuisance function estimators), the second-order contribution satisfies uniformly: $\sup \!{}_{P\in\mathcal P_n^\Pi} \E_{\mathcal{D}_n\sim P ^{\otimes n}} [R_{2,\psi}^2(P,\mathcal D_n, \star)]\lesssim r_{2,n}^2$.
    Then, for either debiasing strategy $\star$ (i.e., label or prior correction), the following hold:
    \begingroup\makeatletter\def\f@size{8}\check@mathfonts
    \vspace{-0.1cm}
    {\setlength{\belowdisplayskip}{0.05cm} \begin{align}
        \underbrace{\mathcal{R}_{l_2^2}(\hat{\psi}_{\theta^*}; \Pi,n)}_{\text{Bayes optimal risk, see Remark~\ref{rem:bayes-optimality}}} \!\!\!\!\!\!\!\!\! = \frac{1}{2} \mathcal{R}_{L_2^2}(\psi;\Pi,n) \le \mathcal{R}_{l_2^2}(\hat{\tilde{\psi}}_{\theta^*}; \Pi,n) &\le \mathcal{R}_{L_2^2}({\tilde{\psi}}; \Pi,n) \lesssim n^{-1}+r_{2,n}^2,
        \label{eq:bayes-non-optimality} \\[-2em] 
        \rotatebox{-90}{\textcolor{BrickRed}{$\le$}} \qquad & \qquad \,\,\, \rotatebox{-90}{\textcolor{BrickRed}{$\le$}} \nonumber \\[-0.4em] 
        \underbrace{\mathfrak M_n(\mathcal P_n^\Pi)}_{\text{minimax risk}} \le \mathcal{W}_{l_2^2}(\hat{\tilde{\psi}}_{\theta^*}; \mathcal P_n^\Pi,n) &\le \mathcal{W}_{L_2^2}({\tilde{\psi}}; \mathcal P_n^\Pi,n) \lesssim n^{-1}+r_{2,n}^2, \label{eq:minimax-optimality}
    \end{align}}
    \endgroup
    where inequalities in \textcolor{BrickRed}{red} hold if $\Pi$ is fully supported on $\mathcal P_n^\Pi$; $\mathcal{R}_{l_2^2}(\hat{{\psi}}; \Pi,n): = \E_{P\sim\Pi} \E_{\mathcal{D}_n\sim P^{\otimes n}}  [\hat{{\psi}}(\mathcal{D}_n) - \psi(P)]^2$, $\mathcal{R}_{L_2^2}({\tilde{\psi}}; \Pi,n) := \E_{P\sim\Pi} \E_{\mathcal{D}_n\sim P^{\otimes n}} \lvert\lvert{\tilde{\psi}-\psi}\rvert\rvert^2_{L_2(\tilde q_{\theta^*}, \mathcal{D}_n)}$ are $\Pi$-prior-averaged squared $l_2/L_2$ risks; $\mathcal{W}_{l_2^2}(\hat{{\psi}}; \mathcal P_n^\Pi,n) := \sup_{P \in \mathcal P_n^\Pi} \E_{\mathcal{D}_n\sim P^{\otimes n}}[\hat{{\psi}}(\mathcal{D}_n) - \psi(P)]^2$  and $\mathcal{W}_{L_2^2}({\tilde{\psi}}; \mathcal P_n^\Pi,n):=\sup_{P \in \mathcal P_n^\Pi} \E_{\mathcal{D}_n\sim P^{\otimes n}} \rvert\rvert{\tilde{\psi}-\psi}\rvert\rvert^2_{L_2(\tilde q_{\theta^*}, \mathcal{D}_n)}$ are worst-case squared $l_2/L_2$ risks; and $\mathfrak M_n(\mathcal P_n^\Pi):= \inf_{\hat{\psi}: \mathcal{Z}^{n} \to \Psi} \mathcal{W}_{l_2^2}(\hat{{\psi}}; \mathcal P_n^\Pi,n)$ denotes the minimax risk over $\mathcal P_n^\Pi$.
\end{theorem}

\begin{wrapfigure}{r}{0.48\textwidth}
    \vspace{-0.9cm}
    \centering
    \includegraphics[width=\linewidth]{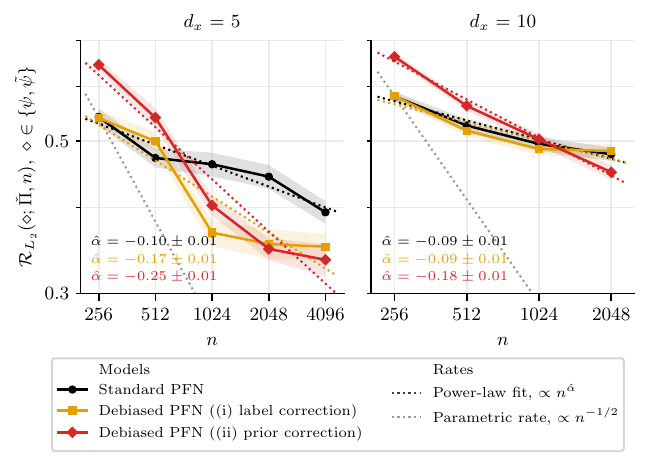}
    \vspace{-0.85cm}
    \caption{\textbf{$L_2$ risk on the evaluation prior $\check \Pi$} (based on Example~\ref{ex:c} with $w\equiv1$; axes are log-scaled). Shown: mean $\pm$ SE $\check \Pi$-averaged $L_2$ concentration errors (solid lines)
    and the corresponding power-law regression fits (dotted lines) depending on dataset sizes $n$ and covariate dimensionality $d_x$ (over $M=200$ ground-truth $\check P$, each with $D=200$ datasets; lower is better). Here, $\diamond \in \{\psi, \tilde\psi\}$ are the target posteriors from the standard and debiased PFNs, respectively.}
    \label{fig:part2-exampleC-ATE}
    \vspace{-1.3cm}
\end{wrapfigure} 

Theorem~\ref{thrm:minimax} establishes a \emph{rate}, rather than an exact notion of optimality. Notably, both strategies (i) and (ii) are generally \emph{not Bayes-optimal under the original prior} $\Pi$ (see Eq.~\eqref{eq:bayes-non-optimality}): as (i)~modifies the training labels, and (ii) is only Bayes-optimal wrt. the corrected prior $\widetilde \Pi$. Nevertheless, whenever the matching minimax risk satisfies $\mathfrak M_n(\mathcal P_n^\Pi)\gtrsim n^{-1}+r_{2,n}^2$ (which, in particular, holds for Examples~\ref{ex:a}--\ref{ex:c}, see Appendix~\ref{app:minimax-examples}), the worst-case risk of the debiased PFN matches the minimax rate, while its original-prior-averaged risk remains bounded by the same rate (see Eq.~\eqref{eq:minimax-optimality}). Hence, the debiased PFNs are \textbf{globally minimax-rate optimal over the class $\mathcal P_n^\Pi$}. On the other hand, the same upper bound in Eq.~\eqref{eq:minimax-optimality} \textbf{does not generally hold for the original PFNs} (as a result of Theorem~\ref{thrm:slow-rates}). 

%%%%%%%%%%%%%%%%%%%%%%%%%%%%%%%%%%%%%%%%%%%%%%%%%%%%%%%%%%%%%%%%%%%%%%%%%%%%%%%%%%%%%%%%%%
\vspace{-0.1cm}
\section{Experiments} \label{sec:experiments}
\vspace{-0.1cm}
%%%%%%%%%%%%%%%%%%%%%%%%%%%%%%%%%%%%%%%%%%%%%%%%%%%%%%%%%%%%%%%%%%%%%%%%%%%%%%%%%%%%%%%%%%

\textbf{Setup.} To validate our theory, we evaluate the debiased PFN with \textbf{\emph{(i)~label correction}} and \textbf{\emph{(ii)~prior correction}} (Sec.~\ref{subsec:debiasing-strategies}) for the ATE (Example~\ref{ex:c} with $w \equiv 1$). In Appendix~\ref{app:additional-results-A}, we further provide the results for mean value estimation (Example~\ref{ex:a}). For ATE estimation, we use an adapted synthetic prior $\Pi$ over nonlinear structural causal models from TabPFN $\mathrm{v1}$ \citep{hollmann2023tabpfn}. For each $d_x \in \{5, 10\}$ and $n \in \{256, 512, 1024, 2048\}$, we train a \emph{standard PFN} (Eq.~\eqref{eq:pfn-nll-semiparam}) and two \emph{debiased PFNs (i)-(ii)} (Eq.~\eqref{eq:label-corr},\eqref{eq:prior-corr}) with the same architecture, training data, and optimization budget (thus, all three PFN variants are comparable). We provide all the implementation details in Appendix~\ref{app:implementation-details}.

\textbf{Evaluation.} Given the Bayes-optimality of the idealized PFNs (Remark~\ref{rem:bayes-optimality}), we want to evaluate the performance of PFNs based on a \emph{modified prior} $\check \Pi$ that consists of preselected worst-case distributions (\ie, those that yield worst-case posterior in Theorem~\ref{thrm:slow-rates}; see further details in Appendix~\ref{app:implementation-details}). While the modified evaluation prior $\check \Pi$ has approximately the same support, now the Bayes-optimality does not apply, and, hence, we can empirically validate our theoretical findings. Our evaluation then proceeds in three parts. \textbf{(1)}~As a sanity check, we evaluate the amount of observed confounding in sampled datasets (with the training/ evaluation priors $\Pi / \check \Pi$, and after debiasing applied to the training labels and prior). Following Remark~\ref{rem:pib} in Appendix~\ref{app:theory}, this allows to check the potential amount of the first-order bias ($F_\psi$) in different datasets. Then, \textbf{(2)}~to verify the results of Theorems~\ref{thrm:debiased-rates} and \ref{thrm:minimax}, we evaluate the $L_2$ risk for different PFNs based on the evaluation prior $\check \Pi$, namely, $\mathcal{R}_{L_2}(\psi; \check \Pi,n) = \E_{\check P\sim \check\Pi} \E_{\mathcal{D}_n\sim \check P^{\otimes n}} \lvert\lvert{{\psi}-\check\psi}\rvert\rvert_{L_2(q_{\theta^*}, \mathcal{D}_n)}$. Finally, \textbf{(3)}~to showcase Theorem~\ref{thrm:bbvm}, we measure $\Pi/\check \Pi$-prior-averaged empirical total variation between the PFN-based posteriors and a frequentist A-IPTW estimator with the ground-truth nuisance functions.

\textbf{Results for part (1).} We defer the results for part \textbf{(1)} to Appendix~\ref{app:additional-results-C}. Therein, (a) the evaluation prior $\check \Pi$ contains a significantly larger spread of the observed confounding $\Delta$ than the training prior $\Pi$ (hence making it ``worst-case''); and (b)~both debiasing strategies target this mismatch, yet in different ways: (ii)~prior correction spreads $\Delta$ of the training targets at least as widely as under $\check\Pi$, whereas (i)~label correction leaves the spread nearly unchanged. The latter is expected, as (i)~corrects the training labels with a data-dependent one-step term rather than modifying the prior.

\textbf{Results for part (2).} Fig.~\ref{fig:part2-exampleC-ATE} shows the results. Here, our both debiased PFNs often outperform the standard PFN for sufficiently large $n$ (\ie, $n \ge 1024$ for $d_x=5$ and $n \ge 2048$ for $d_x=10$), and achieve much faster concentration rates $\hat{\alpha}$ (especially for $d_x=5$). This confirms the asymptotic optimality of both debiasing strategies based on a worst-case evaluation prior $\check \Pi$ (as suggested by Theorems~\ref{thrm:debiased-rates} and \ref{thrm:minimax}).  

\textbf{Results for part (3).} The results are in Appendix~\ref{app:additional-results-C}. There, based on the evaluation prior $\check \Pi$, the total variation of all the PFNs grows with $n$, which suggests that the second-order remainder and amortization errors are not negligible for the considered $n$. Still, \emph{our debiased PFNs are closer to the efficient normal approximation} (A-IPTW estimator) than the standard PFN (especially, (ii)~prior correction). 

\textbf{Conclusion.} Our work provides two generic strategies for debiasing PFNs for semi-parametric inference with rigorous and provable asymptotic optimality guarantees.

%%%%%%%%%%%%%%%%%%%%%%%%%%%%%%%%%%%%%%%%%%%%%%%%%%%%%%%%%%%%%%%%%%%%%%%%%%%%%%%%%%%%%%%%%%
% \vspace{-0.1cm}
% \section{Conclusion}
% \vspace{-0.1cm}
%%%%%%%%%%%%%%%%%%%%%%%%%%%%%%%%%%%%%%%%%%%%%%%%%%%%%%%%%%%%%%%%%%%%%%%%%%%%%%%%%%%%%%%%%%

% \newpage
\subsection*{AI use statement}

In this work, we used generative AI tools (LLM-based coding assistants and agents) to implement methods, to provide feedback on the design of the experiments, and to assist in interpreting experimental results.
We have also used generative AI tools to aid the formulation of mathematical claims, proofs, and hypotheses.
All synthetic data were produced by our own simulation code, not generated by AI models; translation, dataset cleaning, and qualitative data analysis do not apply to this work.
Additionally, we used generative AI tools to create and edit software code and scientific figures and to draft and edit parts of the paper.
We have reviewed all AI-assisted work.
All theoretical results and proofs were checked and verified by the authors.
We take responsibility for the final content of this work, including text, claims, and artifacts produced with the aid of generative AI.

\subsection*{Ethics statement}

This work is methodological and uses only simulated data; it involves no human subjects, personal data, or data releases. We do not foresee specific ethical concerns beyond the general caveat that uncertainty estimates from amortized models should be validated before use in high-stakes decisions.

\subsection*{Reproducibility statement}

The assumptions for all theoretical results are stated in the main text, and complete proofs are given in Appendix~\ref{app:theory}. The data-generating prior, the training procedure of all PFN variants, and the evaluation protocol are described in Sec.~\ref{sec:experiments} and Appendix~\ref{app:implementation-details}, with additional results in Appendix~\ref{app:additional-results}. Anonymous source code covering data generation, training, inference, and metric computation, including a pipeline script with all random seeds fixed, is available at \url{https://anonymous.4open.science/r/Ortho-PFN-Anonymized-E4B7}.

\subsubsection*{Acknowledgments} This paper is supported by the DAAD programme Konrad Zuse Schools of Excellence in Artificial Intelligence, sponsored by the Federal Ministry of Research, Technology and Space. S.F. acknowledges funding via Swiss National Science Foundation Grant 186932. This work has been supported by the German Federal Ministry of Education and Research (Grant: 01IS24082).

\bibliography{bibliography}

@inproceedings{alaa2018limits,
  title={Limits of estimating heterogeneous treatment effects: Guidelines for practical algorithm design},
  author={Alaa, Ahmed M. and van der Schaar, Mihaela},
  booktitle={International Conference on Machine Learning},
  year={2018}
}

@inproceedings{alaa2017bayesian,
  title={Bayesian inference of individualized treatment effects using multi-task {Gaussian} processes},
  author={Alaa, Ahmed M. and van der Schaar, Mihaela},
  booktitle={Advances in Neural Information Processing Systems},
  year={2017}
}

@article{chipman2010bart,
  title={{BART}: {B}ayesian additive regression trees},
  author={Chipman, Hugh A. and George, Edward I. and McCulloch, Robert E.},
  journal={The Annals of Applied Statistics},
  volume={4},
  number={1},
  year={2010},
  publisher={Institute of Mathematical Statistics}
}

@article{shi2019adapting,
  title={Adapting neural networks for the estimation of treatment effects},
  author={Shi, Claudia and Blei, David and Veitch, Victor},
  journal={Advances in Neural Information Processing Systems},
  year={2019}
}

@inproceedings{garnelo2018conditional,
  title={Conditional neural processes},
  author={Garnelo, Marta and Rosenbaum, Dan and Maddison, Christopher and Ramalho, Tiago and Saxton, David and Shanahan, Murray and Teh, Yee Whye and Rezende, Danilo and Eslami, S.M. Ali},
  booktitle={International Conference on Machine Learning},
  year={2018},
}

@inproceedings{brown2020language,
  title={Language models are few-shot learners},
  author={Brown, Tom and Mann, Benjamin and Ryder, Nick and Subbiah, Melanie and Kaplan, Jared D. and Dhariwal, Prafulla and Neelakantan, Arvind and Shyam, Pranav and Sastry, Girish and Askell, Amanda and others},
  booktitle={Advances in Neural Information Processing Systems},
  year={2020}
}

@inproceedings{balazadeh2025causalpfn,
  title={{CausalPFN}: Amortized causal effect estimation via in-context learning},
  author={Balazadeh, Vahid and Kamkari, Hamidreza and Thomas, Valentin and Li, Benson and Ma, Junwei and Cresswell, Jesse C. and Krishnan, Rahul G.},
  booktitle={Advances in Neural Information Processing Systems},
  year={2025}
}

@inproceedings{robertson2025pfn,
  title={{Do-PFN}: In-context learning for causal effect estimation},
  author={Robertson, Jake and Reuter, Arik and Guo, Siyuan and Hollmann, Noah and Hutter, Frank and Sch{\"o}lkopf, Bernhard},
  booktitle={Advances in Neural Information Processing Systems},
  year={2025}
}

@article{hollmann2025accurate,
  title={Accurate predictions on small data with a tabular foundation model},
  author={Hollmann, Noah and M{\"u}ller, Samuel and Purucker, Lennart and Krishnakumar, Arjun and K{\"o}rfer, Max and Hoo, Shi Bin and Schirrmeister, Robin Tibor and Hutter, Frank},
  journal={Nature},
  volume={637},
  number={8045},
  pages={319--326},
  year={2025},
  publisher={Nature Publishing Group UK London}
}

@inproceedings{ma2026foundation,
    title={Foundation Models for Causal Inference via Prior-Data Fitted Networks},
    author={Yuchen Ma and Dennis Frauen and Emil Javurek and Stefan Feuerriegel},
    booktitle={International Conference on Learning Representations},
    year={2026},
}

@inproceedings{dong2024survey,
  title={A survey on in-context learning},
  author={Dong, Qingxiu and Li, Lei and Dai, Damai and Zheng, Ce and Ma, Jingyuan and Li, Rui and Xia, Heming and Xu, Jingjing and Wu, Zhiyong and Chang, Baobao and others},
  booktitle={Conference on Empirical Methods in Natural Language Processing},
  year={2024}
}

@inproceedings{kim2019attentive,
  title={Attentive Neural Processes},
  author={Kim, Hyunjik and Mnih, Andriy and Schwarz, Jonathan and Garnelo, Marta and Eslami, Ali and Rosenbaum, Dan and Vinyals, Oriol and Teh, Yee Whye},
  booktitle={International Conference on Learning Representations},
  year={2019}
}

@inproceedings{xie2022an,
    title={An Explanation of In-context Learning as Implicit {Bayesian} Inference},
    author={Sang Michael Xie and Aditi Raghunathan and Percy Liang and Tengyu Ma},
    booktitle={International Conference on Learning Representations},
    year={2022},
}

@inproceedings{muller2022transformers,
    title={Transformers Can Do {Bayesian} Inference},
    author={Samuel M{\"u}ller and Noah Hollmann and Sebastian Pineda Arango and Josif Grabocka and Frank Hutter},
    booktitle={International Conference on Learning Representations},
    year={2022},
}

@inproceedings{hollmann2023tabpfn,
    title={Tab{PFN}: A Transformer That Solves Small Tabular Classification Problems in a Second},
    author={Noah Hollmann and Samuel M{\"u}ller and Katharina Eggensperger and Frank Hutter},
    booktitle={International Conference on Learning Representations},
    year={2023}
}

@article{nagler2025uncertainty,
  title={Uncertainty Quantification for Prior-Data Fitted Networks using Martingale Posteriors},
  author={Nagler, Thomas and R{\"u}gamer, David},
  journal={arXiv preprint arXiv:2505.11325},
  year={2025}
}

@article{linero2023and,
  title={The how and why of {Bayesian} nonparametric causal inference},
  author={Linero, Antonio R. and Antonelli, Joseph L.},
  journal={Wiley Interdisciplinary Reviews: Computational Statistics},
  volume={15},
  number={1},
  pages={e1583},
  year={2023},
  publisher={Wiley Online Library}
}

@article{hahn2018regularization,
  title={Regularization and Confounding in Linear Regression for Treatment Effect Estimation},
  author={Hahn, P. Richard and Carvalho, Carlos M and Puelz, David and others},
  journal={Bayesian Analysis},
  volume={13},
  number={1},
  year={2018}
}

@article{yiu2025semiparametric,
  title={Semiparametric posterior corrections},
  author={Yiu, Andrew and Fong, Edwin and Holmes, Chris and Rousseau, Judith},
  journal={Journal of the Royal Statistical Society Series B: Statistical Methodology},
  volume={87},
  number={4},
  pages={1025--1054},
  year={2025},
  publisher={Oxford University Press UK}
}

@article{ray2020semiparametric,
  title={Semiparametric {Bayesian} Causal Inference},
  author={Ray, Kolyan and van der Vaart, Aad},
  journal={The Annals of Statistics},
  volume={48},
  number={5},
  pages={2999--3020},
  year={2020},
  publisher={JSTOR}
}

@article{ng2025tabmgp,
  title={{TabMGP}: Martingale Posterior with {TabPFN}},
  author={Ng, Kenyon and Fong, Edwin and Frazier, David T. and Knoblauch, Jeremias and Wei, Susan},
  journal={arXiv preprint arXiv:2510.25154},
  year={2025}
}

@article{li2023bayesian,
  title={Bayesian causal inference: {A} critical review},
  author={Li, Fan and Ding, Peng and Mealli, Fabrizia},
  journal={Philosophical Transactions of the Royal Society A},
  volume={381},
  number={2247},
  pages={20220153},
  year={2023},
  publisher={The Royal Society}
}

@inproceedings{ray2019debiased,
  title={Debiased {Bayesian} inference for average treatment effects},
  author={Ray, Kolyan and Szab{\'o}, Botond},
  booktitle={Advances in Neural Information Processing Systems},
  year={2019}
}

@inproceedings{hatt2021estimating,
  title={Estimating average treatment effects via orthogonal regularization},
  author={Hatt, Tobias and Feuerriegel, Stefan},
  booktitle={International Conference on Information and Knowledge Management},
  year={2021}
}

@inproceedings{robins2000robust,
  title={Robust estimation in sequentially ignorable missing data and causal inference models},
  author={Robins, James M.},
  booktitle={Proceedings of the American Statistical Association},
  year={2000}
}

@article{newey1994asymptotic,
  title={The asymptotic variance of semiparametric estimators},
  author={Newey, Whitney K.},
  journal={Econometrica},
  volume={62},
  number={6},
  pages={1349--1382},
  year={1994},
  publisher={JSTOR}
}

@inproceedings{jin2025structure,
  title={Structure-agnostic Optimality of Doubly Robust Learning for Treatment Effect Estimation},
  author={Jin, Jikai and Syrgkanis, Vasilis},
  booktitle={Annual Conference on Learning Theory},
  year={2025},
}

@book{bickel1993efficient,
  title={Efficient and adaptive estimation for semiparametric models},
  author={Bickel, Peter J. and Klaassen, Chris A.J. and Ritov, Ya’acov and Wellner, Jon A.},
  year={1993},
  publisher={Springer New York}
}

@book{tsiatis2006semiparametric,
  title={Semiparametric theory and missing data},
  author={Tsiatis, Anastasios A.},
  year={2006},
  publisher={Springer New York}
}

@article{van2006targeted,
  title={Targeted Maximum Likelihood Learning},
  author={Mark J. van der Laan and Daniel B. Rubin},
  journal={The International Journal of Biostatistics},
  year={2006},
  volume={2},
}

@inproceedings{chernozhukov2022riesznet,
  title={{RieszNet} and {ForestRiesz}: Automatic debiased machine learning with neural nets and random forests},
  author={Chernozhukov, Victor and Newey, Whitney and Quintas-Mart{\i}nez, V{\i}ctor M. and Syrgkanis, Vasilis},
  booktitle={International Conference on Machine Learning},
  year={2022},
}

@inproceedings{hines2025automatic,
    title={Automatic debiasing of neural networks via moment-constrained learning},
    author={Christian L. Hines and Oliver J. Hines},
    booktitle={Conference on Causal Learning and Reasoning},
    year={2025},
}

@article{bickel2012semiparametric,
  title={The semiparametric {Bernstein}-von {M}ises theorem},
  author={Bickel, PJ and Kleijn, BJK},
  journal={The Annals of Statistics},
  pages={206--237},
  year={2012},
  publisher={JSTOR}
}

@incollection{hartigan1983asymptotic,
  title={Asymptotic normality of posterior distributions},
  author={Hartigan, J.A.},
  booktitle={Bayes theory},
  pages={107--118},
  year={1983},
  publisher={Springer}
}

@article{kennedy2024semiparametric,
  title={Semiparametric doubly robust targeted double machine learning: a review},
  author={Kennedy, Edward H.},
  journal={Handbook of Statistical Methods for Precision Medicine},
  pages={207--236},
  year={2024},
  publisher={Chapman and Hall/CRC}
}

@book{van2000asymptotic,
  title={Asymptotic statistics},
  author={van der Vaart, Aad W.},
  volume={3},
  year={2000},
  address={Cambridge, United Kingdom},
  publisher={Cambridge University Press}
}

@inproceedings{melnychuk2026frequentist,
  title={Frequentist Consistency of Prior-Data Fitted Networks for Causal Inference},
  author={Melnychuk, Valentyn and Balazadeh, Vahid and Feuerriegel, Stefan and Krishnan, Rahul G.},
  booktitle={International Conference on Machine Learning},
  year={2026}
}

@inproceedings{gershman2014amortized,
  title={Amortized inference in probabilistic reasoning},
  author={Gershman, Samuel and Goodman, Noah},
  booktitle={The Annual Meeting of the Cognitive Science Society},
  year={2014}
}

@inproceedings{le2017inference,
  title={Inference compilation and universal probabilistic programming},
  author={Le, Tuan Anh and Baydin, Atilim Gunes and Wood, Frank},
  booktitle={International Conference on Artificial Intelligence and Statistics},
  year={2017}
}

@article{qi2026survivalpfn,
  title={{SurvivalPFN}: Amortizing Survival Prediction via In-Context {B}ayesian Inference},
  author={Qi, Shi-ang and Balazadeh, Vahid and Cooper, Michael and Greiner, Russell and Krishnan, Rahul G.},
  journal={arXiv preprint arXiv:2605.15488},
  year={2026}
}

@inproceedings{muller2023pfns4bo,
  title={{PFNs4BO}: In-context learning for {B}ayesian optimization},
  author={M{\"u}ller, Samuel and Feurer, Matthias and Hollmann, Noah and Hutter, Frank},
  booktitle={International Conference on Machine Learning},
  year={2023}
}

@article{javurek2026amortizing,
  title={Amortizing Causal Sensitivity Analysis via Prior Data-Fitted Networks},
  author={Javurek, Emil and Frauen, Dennis and Brockschmidt, Marie and Schweisthal, Jonas and Feuerriegel, Stefan},
  journal={arXiv preprint arXiv:2605.10590},
  year={2026}
}

@article{dooley2023forecastpfn,
  title={{ForecastPFN}: Synthetically-trained zero-shot forecasting},
  author={Dooley, Samuel and Khurana, Gurnoor Singh and Mohapatra, Chirag and Naidu, Siddartha V. and White, Colin},
  journal={Advances in Neural Information Processing Systems},
  year={2023}
}

@article{ansari2024chronos,
    title={Chronos: Learning the Language of Time Series},
    author={Abdul Fatir Ansari and Lorenzo Stella and Ali Caner Turkmen and Xiyuan Zhang and Pedro Mercado and Huibin Shen and Oleksandr Shchur and Syama Sundar Rangapuram and Sebastian Pineda Arango and Shubham Kapoor and Jasper Zschiegner and Danielle C. Maddix and Hao Wang and Michael W. Mahoney and Kari Torkkola and Andrew Gordon Wilson and Michael Bohlke-Schneider and Bernie Wang},
    journal={Transactions on Machine Learning Research},
    year={2024},
}

@inproceedings{reuter2026use,
    title={Use What You Know: Causal Foundation Models with Partial Graphs},
    author={Arik Reuter and Anish Dhir and Cristiana Diaconu and Jake Robertson and Ole Ossen and Frank Hutter and Adrian Weller and Mark van der Wilk and Bernhard Sch{\"o}lkopf},
    booktitle={International Conference on Machine Learning},
    year={2026},
}

@article{seletkov2026survival,
  title={Survival In-Context: Amortized {B}ayesian Survival Analysis via Prior-Fitted Networks},
  author={Seletkov, Dmitrii and Hager, Paul and Kaissis, Georgios and Braren, Rickmer and Rueckert, Daniel and Rehms, Raphael},
  journal={arXiv preprint arXiv:2603.29475},
  year={2026}
}

@article{mahajan2025amortized,
    title={Amortized Inference of Causal Models via Conditional Fixed-Point Iterations},
    author={Divyat Mahajan and Jannes Gladrow and Agrin Hilmkil and Cheng Zhang and Meyer Scetbon},
    journal={Transactions on Machine Learning Research},
    year={2025}
}

@article{bohm2026survpfn,
  title={{SurvPFN}: Towards Foundation Models for Survival Predictions},
  author={B{\"o}hm, Samuel and Purucker, Lennart and Hutter, Frank and Schlosser, Pascal},
  journal={arXiv preprint arXiv:2606.04564},
  year={2026}
}

@inproceedings{taga2025timepfn,
  title={{TimePFN}: Effective multivariate time series forecasting with synthetic data},
  author={Taga, Ege Onur and Ildiz, Muhammed Emrullah and Oymak, Samet},
  booktitle={AAAI Conference on Artificial Intelligence},
  year={2025}
}

@inproceedings{nagler2023statistical,
  title={Statistical foundations of prior-data fitted networks},
  author={Nagler, Thomas},
  booktitle={International Conference on Machine Learning},
  year={2023}
}

@inproceedings{hulskamp2026uncertainty,
  title={On the Uncertainty in Prior-Data Fitted Network Pretraining},
  author={H{\"u}lskamp, Manuel and Kobialka, Julius and Sommer, Emanuel and R{\"u}gamer, David},
  year={2026},
  booktitle={ICML Workshop on Foundation Models for Structured Data}
}

@inproceedings{ching2026efficient,
    title={Efficient and Minimax Optimal In-context Nonparametric Regression with Transformers},
    author={Michelle Ching and Ioana Popescu and Nico Smith and Tianyi Ma and William G. Underwood and Richard J. Samworth},
    booktitle={International Conference on Machine Learning},
    year={2026},
}

@article{ma2025optimal,
  title={Optimal in-context adaptivity and distributional robustness of transformers},
  author={Ma, Tianyi and Wang, Tengyao and Samworth, Richard J.},
  journal={arXiv preprint arXiv:2510.23254},
  year={2025}
}

@inproceedings{muller2025position,
    title={Position: {T}he Future of {B}ayesian Prediction Is Prior-Fitted},
    author={Samuel M{\"u}ller and Arik Reuter and Noah Hollmann and David R{\"u}gamer and Frank Hutter},
    booktitle={International Conference on Machine Learning Position Paper Track},
    year={2025},
}

@inproceedings{rueter2025can,
    title={Can Transformers Learn Full {B}ayesian Inference in Context?},
    author={Reuter, Arik and Rudner, Tim G.J. and Fortuin, Vincent and R{\"u}gamer, David},
    booktitle={International Conference on Machine Learning Position Paper Track},
    year={2025},
}

@article{brockschmidt2026confoundingshap,
  title={Confounding{SHAP}: Quantifying Confounding Strength in Causal Inference},
  author={Brockschmidt, Marie and Thies, Santo M. A. R. and Schr{\"o}der, Maresa and Frauen, Dennis and Melnychuk, Valentyn and Muschalik, Maximilian and H{\"u}llermeier, Eyke and Feuerriegel, Stefan},
  journal={arXiv preprint arXiv:2605.10533},
  year={2026}
}

@article{fonseca2026explainerpfn,
  title={{ExplainerPFN}: Towards tabular foundation models for model-free zero-shot feature importance estimations},
  author={Fonseca, Joao and Stoyanovich, Julia},
  journal={arXiv preprint arXiv:2601.23068},
  year={2026}
}

@article{freedman1999wald,
  title={Wald Lecture: On the {Bernstein-von Mises} theorem with infinite-dimensional parameters},
  author={Freedman, David},
  journal={The Annals of Statistics},
  volume={27},
  number={4},
  pages={1119--1141},
  year={1999},
  publisher={Institute of Mathematical Statistics}
}

@article{balakrishnan2026fundamental,
  title={The fundamental limits of structure-agnostic functional estimation},
  author={Balakrishnan, Sivaraman and Kennedy, Edward and Wasserman, Larry},
  journal={Statistical Science},
  volume={41},
  number={3},
  pages={659--670},
  year={2026},
  publisher={Institute of Mathematical Statistics}
}

@article{angelopoulos2023prediction,
  title={Prediction-powered inference},
  author={Angelopoulos, Anastasios N. and Bates, Stephen and Fannjiang, Clara and Jordan, Michael I. and Zrnic, Tijana},
  journal={Science},
  volume={382},
  number={6671},
  pages={669--674},
  year={2023},
  publisher={American Association for the Advancement of Science}
}

@article{hirano2003efficient,
  title={Efficient estimation of average treatment effects using the estimated propensity score},
  author={Hirano, Keisuke and Imbens, Guido W. and Ridder, Geert},
  journal={Econometrica},
  volume={71},
  number={4},
  pages={1161--1189},
  year={2003},
  publisher={Wiley Online Library}
}

@article{kim2024transformers,
  title={Transformers are minimax optimal nonparametric in-context learners},
  author={Kim, Juno and Nakamaki, Tai and Suzuki, Taiji},
  journal={Advances in Neural Information Processing Systems},
  year={2024}
}

@article{van2026calibeating,
  title={Calibeating prediction-powered inference},
  author={van der Laan, Lars and van der Laan, Mark},
  journal={arXiv preprint arXiv:2604.21260},
  year={2026}
}

@article{grinsztajn2025tabpfn,
  title={{TabPFN-2.5}: Advancing the State of the Art in Tabular Foundation Models},
  author={Grinsztajn, L{\'e}o and Fl{\"o}ge, Klemens and Key, Oscar and Birkel, Felix and Jund, Philipp and Roof, Brendan and J{\"a}ger, Benjamin and Safaric, Dominik and Alessi, Simone and Hayler, Adrian and others},
  journal={arXiv preprint arXiv:2511.08667},
  year={2025}
}

@article{grinsztajn2026tabpfn,
  title={{TabPFN-3}: Technical report},
  author={Grinsztajn, L{\'e}o and Fl{\"o}ge, Klemens and Key, Oscar and Birkel, Felix and Jund, Philipp and Roof, Brendan and Manium, Mihir and Hoo, Shi Bin and B{\"u}hler, Magnus and Garg, Anurag and others},
  journal={arXiv preprint arXiv:2605.13986},
  year={2026}
}

@article{robins1995analysis,
  title={Analysis of semiparametric regression models for repeated outcomes in the presence of missing data},
  author={Robins, James M. and Rotnitzky, Andrea and Zhao, Lue Ping},
  journal={Journal of the American Statistical Association},
  volume={90},
  number={429},
  pages={106--121},
  year={1995},
  publisher={Taylor \& Francis}
}

@article{vaswani2017attention,
  title={Attention is all you need},
  author={Vaswani, Ashish and Shazeer, Noam and Parmar, Niki and Uszkoreit, Jakob and Jones, Llion and Gomez, Aidan N. and Kaiser, {\L}ukasz and Polosukhin, Illia},
  journal={Advances in Neural Information Processing Systems},
  year={2017}
}

@article{dawid1982well,
  title={The well-calibrated {B}ayesian},
  author={Dawid, A. Philip},
  journal={Journal of the American Statistical Association},
  volume={77},
  number={379},
  pages={605--610},
  year={1982},
  publisher={Taylor \& Francis}
}

@article{van2011information,
  title={Information Rates of Nonparametric {G}aussian Process Methods},
  author={van der Vaart, Aad and van Zanten, Harry},
  journal={Journal of Machine Learning Research},
  volume={12},
  number={6},
  year={2011}
}

@book{ghosal2017fundamentals,
  title={Fundamentals of Nonparametric {B}ayesian Inference},
  author={Ghosal, Subhashis and van der Vaart, Aad},
  series={Cambridge Series in Statistical and Probabilistic Mathematics},
  year={2017},
  publisher={Cambridge University Press}
}

@article{yang2015semiparametric,
  title={Semiparametric {Bernstein-von Mises} theorem: Second order studies},
  author={Yang, Yun and Cheng, Guang and Dunson, David B.},
  journal={arXiv preprint arXiv:1503.04493},
  year={2015}
}

@article{diaz2015targeted,
  title={Targeted maximum likelihood estimation using exponential families},
  author={D{\'\i}az, Iv{\'a}n and Rosenblum, Michael},
  journal={The International Journal of Biostatistics},
  volume={11},
  number={2},
  pages={233--251},
  year={2015},
  publisher={De Gruyter}
}

@article{gruber2010targeted,
  title={A targeted maximum likelihood estimator of a causal effect on a bounded continuous outcome},
  author={Gruber, Susan and van der Laan, Mark J.},
  journal={The International Journal of Biostatistics},
  volume={6},
  number={1},
  pages={26},
  year={2010}
}

@article{fortini2026principled,
  title={A principled framework for uncertainty decomposition in {TabPFN}},
  author={Fortini, Sandra and Ng, Kenyon and Petrone, Sonia and Rousseau, Judith and Wei, Susan},
  journal={arXiv preprint arXiv:2602.04596},
  year={2026}
}

@article{jin2025sharp,
  title={Sharp structure-agnostic lower bounds for general linear functional estimation},
  author={Jin, Jikai and Syrgkanis, Vasilis},
  journal={arXiv preprint arXiv:2512.17341},
  year={2025}
}

@book{le2012asymptotic,
  title={Asymptotic methods in statistical decision theory},
  author={Le Cam, Lucien},
  year={2012},
  publisher={Springer Science \& Business Media}
}

@book{tsybakov2009introduction,
  title={Introduction to nonparametric estimation},
  author={Tsybakov, Alexandre B. and Zaiats, Vladimir},
  volume={11},
  year={2009},
  publisher={Springer}
}

@article{nabi2026semiparametric,
  title={Semiparametric Efficiency Theory as Differential Calculus on a Space of Probability Distributions},
  author={Nabi, Razieh},
  journal={arXiv preprint arXiv:2606.22784},
  year={2026}
}
\bibliographystyle{iclr2027_conference}

\appendix

%%%%%%%%%%%%%%%%%%%%%%%%%%%%%%%%%%%%%%%%%%%%%%%%%%%%%%%%%%%%%%%%%%%%%%%%%%%%%%%%%%%%%%%%%%
\newpage
\section{Extended Related Work} \label{app:extended-rw}
%%%%%%%%%%%%%%%%%%%%%%%%%%%%%%%%%%%%%%%%%%%%%%%%%%%%%%%%%%%%%%%%%%%%%%%%%%%%%%%%%%%%%%%%%%

\subsection{Prior-data Fitted Networks}

\textbf{Amortized Bayesian inference \& in-context learning.} Amortized Bayesian inference (ABI) replaces repeated, dataset-specific posterior computation by a learned inference map that shares computation across related inference problems \citep{gershman2014amortized,le2017inference}. Closely related ideas appear in neural processes, which learn distributions over functions conditional on observed context sets \citep{garnelo2018conditional,kim2019attentive}. More recently, the emergence of \emph{in-context learning} (ICL) in transformers has provided a particularly flexible mechanism for such amortization: A pretrained model adapts to a new task through its context alone, without updating its parameters \citep{brown2020language,dong2024survey}. Several theoretical perspectives have been proposed to explain ICL, including implicit Bayesian inference \citep{xie2022an}, meta-learning, and the in-context implementation of statistical learning algorithms. The Bayesian and ICL viewpoints become especially closely connected when pretraining tasks are themselves sampled from a prior: conditioning on the in-context dataset then corresponds to learning a posterior predictive rule \citep{muller2022transformers}. 

\textbf{PFNs.} Prior-data fitted networks (PFNs) instantiate this connection between ABI and ICL explicitly \citep{muller2022transformers}. Rather than fitting a model separately to each observed dataset, a PFN is pretrained on synthetic datasets sampled from a user-specified prior over data-generating processes. By predicting held-out quantities conditional on the remaining observations, the network learns to approximate the corresponding Bayesian posterior-predictive distribution; after pretraining, inference on a new dataset reduces to a forward pass through the network. A prominent realization is TabPFN \citep{hollmann2023tabpfn}, which brought the PFN paradigm to tabular classification. Its substantially extended successor supports both classification and regression, categorical and missing features, and datasets with up to thousands of observations, while achieving strong predictive performance on small- and medium-sized tabular datasets \citep{hollmann2025accurate}. The PFN paradigm has since been extended beyond generic tabular prediction, including Bayesian optimization \citep{muller2023pfns4bo}, time-series forecasting \citep{dooley2023forecastpfn,ansari2024chronos,taga2025timepfn}, and, more recently, survival analysis \citep{seletkov2026survival,qi2026survivalpfn,bohm2026survpfn}; see \citet{muller2025position} for a broader perspective. 

\textbf{PFNs for semi-parametric inference.} To our knowledge, explicit applications of PFNs to semi-parametric target estimation have so far concentrated on \emph{causal inference}. \citet{balazadeh2025causalpfn} introduced CausalPFN, which pretrains on observational datasets generated under unconfoundedness and directly amortizes estimation of average and heterogeneous treatment effects. Furthermore, \citet{robertson2025pfn} proposed Do-PFN, which instead pretrains across heterogeneous structural causal models and interventions, thereby learning to predict interventional outcomes without requiring the causal graph at test time. Going beyond these specific constructions, \citet{ma2026foundation} developed CausalFM, a general framework for constructing PFN priors from structural causal models covering back-door, front-door, and instrumental-variable settings. Importantly, they formalized conditions under which such priors yield identifiable and consistent Bayesian causal estimators, highlighting that an inappropriate causal pretraining prior may remain biased even asymptotically. Subsequent work further incorporated partial causal-graph information \citep{reuter2026use} or extended PFN-based amortization to causal sensitivity analysis \citep{javurek2026amortizing}. Again, all the above-mentioned works (at best) established the consistency of estimation \emph{without further studying the asymptotic properties of the PFNs}.

\subsection{Semi-parametric Inference}

\textbf{Frequentist estimation.} Classical semi-parametric inference constructs estimators of a low-dimensional target while treating other components of the data-generating process as infinite-dimensional nuisance functions \citep{kennedy2024semiparametric}. In causal inference, in particular, prominent examples include outcome-regression plug-in estimators, inverse-probability weighting, and estimators that combine the two. A central example is an \emph{augmented inverse probability treatment weighted} (A-IPTW) estimator. Rather than relying solely on estimated nuisance functions, A-IPTW augments a plug-in estimate by an influence-function-based correction. Under standard regularity conditions and sufficiently accurate nuisance estimation, this construction yields $\sqrt{n}$-consistent, asymptotically normal, and semi-parametrically efficient estimation of finite-dimensional functionals such as the ATE \citep{newey1994asymptotic,robins2000robust,bickel1993efficient,tsiatis2006semiparametric}. On the other hand, \emph{targeted maximum likelihood estimation} (TMLE) provides an alternative implementation of the same first-order debiasing principle and is asymptotically equivalent under suitable conditions \citep{van2006targeted}. Modern work increasingly combines such debiased (or orthogonal) estimators with flexible machine-learning estimates of the nuisance functions \citep[e.g.,][]{shi2019adapting,hatt2021estimating,chernozhukov2022riesznet,hines2025automatic}. This separation between flexible nuisance learning and a target-specific correction is particularly important for our setting: We design our debiasing strategies such that the flexibility of PFNs does not come at the cost of slower target estimation rates.

\textbf{Minimax optimality.} Minimax theory characterizes the fundamental statistical difficulty of an estimation problem by asking for the smallest worst-case error achievable over a specified model class \citep{le2012asymptotic,van2000asymptotic,tsybakov2009introduction}. A minimax lower bound thus establishes \emph{a rate below which no estimator can uniformly improve}. At the same time, an upper bound can be obtained by analyzing existing estimators (\eg, debiased estimators from above). When the two bounds match, the corresponding estimator is \emph{minimax optimal}. In semi-parametric problems, however, there is no ``one-size-fits-all'' optimal rate: The minimax rate generally depends on both the target functional and the restrictions imposed on the nuisance functions. Recent work has therefore derived lower bounds for specific classes of functionals: \eg, the integrated squared density and expected conditional covariance \citep{balakrishnan2026fundamental}; weighted average treatment effects \citep{jin2025structure}; and general linear functionals with mixed-bias property \citep{jin2025sharp}. These results motivate our target-specific analysis: In Appendix~\ref{app:minimax-examples}, for each estimand, we compare the rate achieved by our debiased PFN with the corresponding minimax lower bound.

\textbf{Bayesian estimation.} Unlike the frequentist estimators, Bayesian semi-parametric inference places a prior on the unknown components of the data-generating process and propagates the resulting posterior uncertainty to the target estimand. Flexible nuisance models used for this purpose include Gaussian processes \citep{alaa2017bayesian,alaa2018limits} and Bayesian additive regression trees \citep{chipman2010bart}; see \citet{li2023bayesian,linero2023and} for broader discussions. 

\textbf{Bernstein--von Mises property of Bayesian estimators}. For finite-dimensional functionals of an infinite-dimensional model, however, posterior consistency of the nuisance functions alone does not ensure valid uncertainty quantification for the target. The relevant asymptotic property is provided by \emph{semi-parametric Bernstein--von Mises (BvM) theory} \citep{hartigan1983asymptotic,bickel2012semiparametric}. A BvM result establishes that, after appropriate centering and $\sqrt{n}$ scaling, the marginal posterior for the target approaches a normal distribution whose variance coincides with the semi-parametric efficiency bound. Consequently, Bayesian credible sets acquire an \emph{asymptotically valid frequentist interpretation}. Establishing such a result is delicate because regularization or prior-induced bias \citep{hahn2018regularization} in the nuisance components can propagate to the target and dominate its $\sqrt{n}$ fluctuations. This interaction between nuisance contraction, prior bias, and target-level posterior behavior is precisely the aspect of semi-parametric Bayesian theory that is most closely connected to our analysis.

%%%%%%%%%%%%%%%%%%%%%%%%%%%%%%%%%%%%%%%%%%%%%%%%%%%%%%%%%%%%%%%%%%%%%%%%%%%%%%%%%%%%%%%%%%
\newpage
\section{Background Materials} \label{app:background}
%%%%%%%%%%%%%%%%%%%%%%%%%%%%%%%%%%%%%%%%%%%%%%%%%%%%%%%%%%%%%%%%%%%%%%%%%%%%%%%%%%%%%%%%%%

\subsection{Efficient Influence Functions}

We refer to \citet{kennedy2024semiparametric} for a detailed introduction to semi-parametric efficiency theory and to \citet{nabi2026semiparametric} for a geometric interpretation of efficient influence functions (EIFs). Below, we briefly introduce the concepts needed in our analysis; for more details, we refer to \citet{bickel1993efficient,tsiatis2006semiparametric,van2000asymptotic}.

\textbf{Parametric sub-models.} Let $P_0\in\mathcal{P}(\mathcal{Z})$ denote a fixed data-generating distribution. A one-dimensional regular parametric sub-model through $P_0$ is a collection $\{P_t: t \in(-\epsilon,\epsilon)\}\subseteq\mathcal{P}(\mathcal{Z})$ such that $P_{t=0}=P_0$ and the map $t\mapsto P_t$ is sufficiently smooth at $t=0$. The corresponding \emph{score} describes the infinitesimal direction in which the distribution is perturbed:
\begin{align}
    s(Z) := \left.\frac{\partial}{\partial t} \log P_t(Z) \right|_{t=0}.
\end{align}

Any regular score satisfies $P_0s=0$ and $P_0s^2<\infty$, and therefore belongs to the Hilbert space $L_2^0(P_0):=\{f: P_0f=0, P_0f^2<\infty\}$, equipped with inner product $\langle f,g\rangle_{P_0}:=P_0(fg)$. The collection of scores generated by all regular parametric sub-models determines the tangent space
\begin{align}
    \mathcal{T}(P_0) := \overline{\operatorname{span}} \left\{s:s=\left. \frac{\partial}{\partial t} \log P_t
    \right|_{t=0} \text{ for some regular } {P_t} \right\} \subseteq L_2^0(P_0).
\end{align}
Thus, although $\mathcal{P}(\mathcal{Z})$ may be infinite-dimensional, a parametric sub-model probes it along a single local direction. 

For a given target $\psi$, directions $s\in\mathcal{T}(P_0)$ along which $\psi(P_t)$ does not change to first order correspond to nuisance directions. Geometrically, the efficient influence function isolates the component of variation that is relevant for $\psi$ from these target-irrelevant perturbations.

\textbf{Pathwise derivatives.} The functional $\psi:\mathcal{P}(\mathcal{Z})\to\Psi$ is \emph{pathwise differentiable} at $P_0$ if, along every regular parametric sub-model ${P_t}$ through $P_0$, its directional derivative exists and defines a continuous linear functional of the corresponding score $s$:
\begin{align}
    \dot\psi_{P_0}(s) := \left. \frac{\partial}{\partial t} \psi(P_t) \right|_{t=0}.
\end{align}
By the Riesz representation theorem, there then exists a unique element of the tangent space, denoted by $\phi_\psi(\cdot; P_0)$, such that
\begin{align}
    \dot\psi_{P_0}(s) = \left\langle \phi_\psi(\cdot;P_0),s \right\rangle_{P_0} = P_0\left[ \phi_\psi(Z;P_0)\, s(Z) \right], \qquad s\in\mathcal{T}(P_0).
\end{align}

This function is the \emph{efficient influence function} (EIF), also called the canonical gradient, and satisfies $P_0\phi_\psi(\cdot; P_0)=0$. Equivalently, along a sufficiently smooth local path,
\begin{align}
    \psi(P_t)-\psi(P_0) = t\cdot P_0\left[\phi_\psi(Z;P_0)\, s(Z)\right] +o(t).
\end{align}
The EIF therefore describes the first-order sensitivity of the target to perturbations of the underlying distribution. This infinitesimal representation is the local counterpart of the von--Mises expansion used throughout the main text.

The geometry also gives a useful interpretation of the EIF norm. By Cauchy--Schwarz,
\begin{align}
    \norm{\dot\psi_{P_0}(s)} \le \norm{\phi_\psi(\cdot;P_0)}_{L_2(P_0)} \, \cdot \norm{s}_{L_2(P_0)}.
\end{align}
Hence, among unit-norm local perturbations, the EIF direction is the direction along which the target can change most rapidly, whenever such a score is attainable, or can be approximated, by regular parametric sub-models. This observation also motivates our prior correction, which introduces additional prior variation in a target-specific, EIF-aligned direction.

\subsection{Notions of Optimality}

We distinguish three notions of optimality that arise naturally in our analysis. \emph{Local frequentist optimality} compares estimators in shrinking neighborhoods of a fixed data-generating distribution $P_0$; \emph{global frequentist optimality} compares their worst-case behavior over an entire class of distributions; and \emph{Bayes optimality} averages performance over distributions drawn from a prior $\Pi$. These notions are conceptually distinct and, in general, none implies the others.

\textbf{Local frequentist optimality.} Consider first a fixed $P_0$ and $\psi_0=\psi(P_0)$. A regular asymptotically linear estimator $\hat{\psi}_n$ admits an expansion
\begin{align}
    \hat{\psi}_n-\psi_0 = P_n D(\cdot;P_0) + o_{P_0}(n^{-1/2}),
\end{align}
where $D(\cdot;P_0)\in L_2^0(P_0)$ is its influence function. Consequently,
\begin{align}
    \sqrt{n}\big(\hat{\psi}_n-\psi_0\big)\rightsquigarrow N\left(0, P_0D^2(\cdot;P_0)\right).
\end{align}

Regularity requires this first-order behavior to remain stable under local perturbations of $P_0$ of order $n^{-1/2}$. Semi-parametric efficiency theory implies that the EIF has the smallest possible variance among influence functions of regular estimators:
\begin{align}
    P_0D^2(\cdot;P_0) \ge P_0\phi_\psi^2(\cdot;P_0) = \Var_{P_0} \left[ \phi_\psi(Z;P_0) \right] =: V_{\mathrm{eff}}(P_0).
\end{align}
The quantity $V_{\mathrm{eff}}(P_0)$ is the \emph{semi-parametric efficiency bound}. An estimator attaining this bound is locally efficient at $P_0$. The qualifier ``local'' is important: the comparison concerns infinitesimal, typically $n^{-1/2}$-scale, perturbations around a fixed $P_0$, rather than uniform performance over the entire model class.

For Bayesian procedures, a semi-parametric Bernstein--von Mises property provides the corresponding local frequentist guarantee. If a target posterior is asymptotically normal with variance $V_{\mathrm{eff}}(P_0)/n$, its posterior uncertainty attains the same local efficiency bound and its credible sets have an asymptotically valid frequentist interpretation. This is the notion of local optimality later used for our debiased PFNs when the second-order remainder is $o_{P_0}(n^{-1/2})$.

\textbf{Global frequentist optimality.} Local efficiency does not characterize the behavior of an estimator uniformly over a potentially large class of data-generating distributions. For a class $\mathcal{P}'\subseteq\mathcal{P}(\mathcal{Z})$, we instead consider its worst-case squared $l_2$ risk
\begin{align}
    \mathcal{W}_{l_2^2}(\hat{\psi}; \mathcal{P}',n) := \sup_{P\in\mathcal{P}'} \E_{\mathcal{D}_n\sim P^{\otimes n}} \left[\hat{\psi}(\mathcal{D}_n)-\psi(P) \right]^2.
\end{align}
The corresponding minimax risk is
\begin{align}
    \mathfrak{M}_n(\mathcal{P}') := \inf_{\hat{\psi}:\mathcal{P}^{\otimes n}\to\Psi} \mathcal{W}_{l_2^2}(\hat{\psi}; \mathcal{P}',n).
\end{align}
The minimax risk is a property of the statistical problem itself: it gives the smallest worst-case error achievable by any estimator over $\mathcal{P}'$. If, for some sequence $r_n\to0$,
\begin{align}
    \mathfrak{M}_n(\mathcal{P}') \gtrsim r_n^2 \qquad\text{and}\qquad \mathcal{W}_{l_2^2}(\hat{\psi}; \mathcal{P}',n) \lesssim r_n^2,
\end{align}
then $\hat{\psi}$ is \emph{minimax-rate optimal} over $\mathcal{P}'$. Unlike local efficiency, global minimax optimality does not require a $\sqrt{n}$ rate or asymptotic normality: depending on the functional $\psi$ and the complexity of the nuisance functions $\eta(P)$, the optimal rate $r_n$ may be slower than $n^{-1/2}$.

In our analysis later, the relevant class is the effective prior-supported class $\mathcal{P}_n^\Pi$. Thus, the global question is whether the worst-case risk of a debiased PFN over $\mathcal{P}_n^\Pi$ matches a corresponding minimax lower bound. This notion is ``global'' relative to local efficiency because it controls the estimator uniformly across $\mathcal{P}_n^\Pi$, rather than only in shrinking neighborhoods of a single $P_0$.

\textbf{Bayes optimality.} Bayes optimality, on the other hand, evaluates an estimator by averaging its loss with respect to a prior $\Pi$ over data-generating distributions. Under squared loss, the Bayes risk of an estimator $\hat{\psi}$ is
\begin{align}
    \mathcal{R}_{l_2^2}(\hat{\psi}; \Pi,n) := \E_{P\sim\Pi} \E_{\mathcal{D}_n\sim P^{\otimes n}}
    \left[\hat{\psi}(\mathcal{D}_n)-\psi(P) \right]^2.
\end{align}
Conditioning on the observed dataset $\mathcal{D}_n$, the posterior mean
\begin{align}
    \hat{\psi}^{\Pi}_n(\mathcal{D}_n) := \E_{\Pi} \left[\psi(P)\mid\mathcal{D}_n\right]
\end{align}
minimizes the posterior expected squared loss and therefore also minimizes $\mathcal{R}_{l_2^2}(\hat{\psi}; \Pi,n)$ over all estimators.

The Bayes risk is naturally related to the minimax risk introduced above. In particular, for any prior $\Pi$ supported on a class $\mathcal P'\subseteq\mathcal P(\mathcal Z)$,
\begin{align}
    \mathcal R_{l_2^2} \left(\hat{\psi}^{\Pi}_n; \Pi,n\right) &= \inf_{\hat{\psi}} \mathcal R_{l_2^2}(\hat{\psi}; \Pi,n) \le \inf_{\hat{\psi}} \sup_{P\in\mathcal P'} \E_{\mathcal D_n\sim P^{\otimes n}} \left[\hat{\psi}(\mathcal D_n)-\psi(P) \right]^2 = \mathfrak M_n(\mathcal P').
\end{align}
Thus, the Bayes risk under any fixed prior provides a lower bound on the corresponding minimax risk. Equality may hold for a least-favorable prior, whereas a prior concentrating on statistically easier distributions can have a substantially smaller Bayes risk.

Bayes optimality is fundamentally an average-case criterion: it weights performance according to how likely different data-generating distributions are under $\Pi$. It therefore differs from local frequentist optimality, which concerns behavior in shrinking neighborhoods of a fixed $P_0$, and from global minimax optimality, which controls the worst-case risk over a class of distributions. In particular, an estimator can be Bayes optimal under $\Pi$ while concentrating slowly for some fixed $P_0$ or performing poorly on distributions that receive little prior mass. Conversely, modifying the prior or the PFN training objective to improve frequentist concentration rates may sacrifice Bayes optimality under the original prior, while remaining Bayes optimal with respect to the modified prior.

%%%%%%%%%%%%%%%%%%%%%%%%%%%%%%%%%%%%%%%%%%%%%%%%%%%%%%%%%%%%%%%%%%%%%%%%%%%%%%%%%%%%%%%%%%
\newpage
\section{Example C} \label{app:ex-c}
%%%%%%%%%%%%%%%%%%%%%%%%%%%%%%%%%%%%%%%%%%%%%%%%%%%%%%%%%%%%%%%%%%%%%%%%%%%%%%%%%%%%%%%%%%

\begin{examplec}[Weighted average treatment effect (WATE)] 
    The WATE is one of the central estimands in causal inference \citep{hirano2003efficient}. Here,
    $ Z = (X, A, Y = AY_1 + (1-A)Y_0)$ with pre-treatment covariates $X \in [0, 1]^{d_x}$, a treatment $ A \in \{0,1\}$, and the observed outcome $ Y \in [0, 1]$. Then, the WATE is defined as $\psi_w = \E_P[w(X; \pi) (Y_1 - Y_0)] / \E_P(w(X; \pi))$, where $w(\cdot; \pi) \ge 0$ is a user specified weighting function that depends on covariates $x$ through a propensity score $\pi(x) = \mathbb{P}(A=1 \mid X =x)$. For example, $w(x; \pi) =1$ yields an average treatment effect (ATE), and $w(x; \pi) = \pi(x) (1-\pi(x))$ yields the overlap-weighted ATE. Under standard causal assumptions (namely, unconfoundedness, $Y_0,Y_1 \ind A \mid X$, and strong overlap, $\kappa < \pi(x) < 1 - \kappa$, with $\kappa \in (0, 1/2)$), the WATE is identifiable as $\psi_w(P) = \E_P[w(X; \pi) (\mu_1(X) - \mu_0(X))] / \E_P [w(X; \pi)] = \int w(x; \pi)(\mu_1(x) - \mu_0(x)) \diff{P}(x) / \int w(x; \pi) \diff{P}(x)$, where $\mu_a(x) = \E_P[Y \mid A=a, X=x]$ is a conditional expectation. Here, the corresponding nuisance functions, the EIF, and the second-order remainder are as follows:  
    \begingroup\makeatletter\def\f@size{8}\check@mathfonts
    \begin{align}
        \eta(P) &= [\mu_a(x), \mathbbm{1}\{w' \not\equiv 1\} \pi(x), P(x) ], \nonumber \\ 
        \phi_{\psi,w}(Z; \green{P})  &= \frac{w(X; \green{\pi})}{\bar{w}(\green{P})}  \bigg[\frac{A -\green{\pi(}X\green{)}}{\green{\pi(}X\green{)} (1 - \green{\pi(}X\green{)} )}(Y-\green{\mu_A(}X\green{)}) 
        + \bigg((A -\green{\pi(}X\green{)}) \cdot \frac{w'(X; \green{\pi})}{w(X; \green{\pi})} + 1 \bigg)\big( \green{\tau(}X\green{)} - \psi_w(\green{P} )\Big) \bigg],  \nonumber \\
        R_{2, \psi, w}(P, P_0) &=  \frac{1}{\bar{w}(P)} \Bigg[ (\psi_w(P) - \psi_w(P_0)) \cdot \big( (P - P_0)w(\cdot; P) + P_0( \pi - \pi_0)w'(\cdot; \pi) \big) \label{eq:exc-eif} \\
        & \qquad   + P_0 \Big\{ \big( w(\cdot; \pi) - w(\cdot; \pi_0) - w'(\cdot; \pi)(\pi - \pi_0) \big) \cdot (\tau_0 - \psi_w(P_0)) \Big\} \nonumber \\
        & \qquad   + P_0 \bigg\{ \big(\pi - \pi_0 \big) \bigg( \bigg(\frac{w(\cdot; P)}{\pi} - w'(\cdot; P) \bigg) (\mu_1 - \mu_{1,0}) + \bigg(\frac{w(\cdot; P)}{1 - \pi} + w'(\cdot; P) \bigg) (\mu_0 - \mu_{0,0})  \bigg) \bigg\}, \nonumber
    \end{align}
    \endgroup
    where $\bar{w}(P) = {\E_P[} w(X; {\pi})]$, $w'(\cdot; \pi)$ is a derivative of $w$ wrt. to $\pi$, and $\tau(\cdot) = \mu_1(\cdot) - \mu_0(\cdot)$ is a conditional average treatment effect (CATE) \citep{hirano2003efficient}.
\end{examplec}

\subsection{Slow Concentration Rates}

By extending Example~\ref{ex:b-slow}, we analogously show slow concentration rates for the WATE.
\begin{examplec}[WATE]
    Assume that $w$ is uniformly bounded, Lipschitz in $\pi$, and that $\E_P[w(X)]$ is bounded away from zero. Then, for a fixed pair of distributions $P$ and $P_0$, it can be shown that
    \begin{align}
        \abs{\psi_w-\psi_{w,0}} &\lesssim \norm{P-P_0}_{L_2(x)} + \norm{\mu_1-\mu_{1,0}}_{L_2(P_0)} + \norm{\mu_0-\mu_{0,0}}_{L_2(P_0)} \nonumber\\[-0.2em]
        &\qquad + \mathbbm{1}\{w' \not\equiv 0\} \norm{\pi-\pi_0}_{L_2(P_0)}.
    \end{align}
    Hence, a corresponding nuisance-error norm is
    \begin{align}
        \norm{\eta(P)-\eta(P_0)}_{\mathcal H} &:= \norm{P-P_0}_{L_2(x)} + \norm{\mu_1-\mu_{1,0}}_{L_2(P_0)} + \norm{\mu_0-\mu_{0,0}}_{L_2(P_0)} \nonumber\\[-0.2em]
        &\qquad + \mathbbm{1}\{w' \not\equiv 0\} \norm{\pi-\pi_0}_{L_2(P_0)}.
    \end{align}
    Again, this bound is locally rate-sharp. In particular, consider the path $\mu_{1,t}(x)=\mu_{1,0}(x)+t$, while keeping $\mu_{0,t}=\mu_{0,0}$, $\pi_t=\pi_0$, and $P_t(x)=P_0(x)$. Then $w_t=w_0$ and
    \begin{align}
        \psi_w(P_t)-\psi_w(P_0)=t,
        \qquad
        \norm{\mu_{1,t}-\mu_{1,0}}_{L_2(P_0)}=\abs{t}.
    \end{align}
    Similar paths can be analogously found wrt. other nuisance functions. Thus, whenever the posterior assigns asymptotically non-negligible mass around such paths, the WATE concentration error can inherit the concentration rate of the latent nuisance functions in the same order.
\end{examplec}

\subsection{Debiasing Strategies}

\textbf{(i)~Label correction.} Compared to Example~\ref{ex:b-label-corr}, the label correction of the WATE additionally corrects for the propensity score errors.
\begin{examplec}[WATE]
    Applying Eq.~\eqref{eq:label-corr} with the EIF from Eq.~\eqref{eq:exc-eif}, the corrected label is
    \begin{align}
        \tilde{\psi_w}(P) &= \psi_w(P) + \operatorname{BB}_n\{\phi_{\psi,w}(Z;P)\} = \operatorname{BB}_n\{m_w(Z;P)\},
    \end{align}
    where the corresponding pseudo-outcome is
    \begingroup\makeatletter\def\f@size{9}\check@mathfonts
    \begin{align}
        m_w(Z;P) &= \psi_w(P) + \frac{1}{\bar w(P)} \Bigg[w(X;\pi) \, \frac{A - \pi(X)}{\pi(X) (1 - \pi(X))} \, \big(Y-\mu_A(X)\big)
        \\[-0.2em]
        &\qquad\qquad + w(X;\pi) (\tau(X)-\psi_w(P)) + w'(X;\pi)(\tau(X)-\psi_w(P))(A-\pi(X)) \Bigg]. \nonumber
    \end{align}
    \endgroup
    The first two terms inside the brackets correct the first-order errors arising from the outcome regressions and the covariate distribution, respectively, while the final term additionally corrects the first-order propensity-score error induced by the dependence of $w$ on $\pi$. In particular, for the ATE, $w\equiv1$ and hence $w'\equiv0$, so the last term vanishes, and the label reduces to the usual A-IPTW pseudo-outcome. For the overlap-weighted ATE, $w(\pi)=\pi(1-\pi)$ and $w'(\pi)=1-2\pi$, such that the additional propensity correction is generally non-zero.
\end{examplec}

\textbf{(ii)~Prior correction.} Again, the prior correction of the WATE extends the previous Example~\ref{ex:b-prior-corr} by additionally perturbing the prior for the propensity score.
\begin{examplec}[WATE]
    Assume that a prior draw factorizes as $P(z)=P(x)\pi(x)^a(1-\pi(x))^{1-a}P(y\mid x,a)$ and that $P(y\mid x,A=a)=N(\mu_a(x);\sigma_a^2(x))$. Let $\bar w(P)=P[w(X;\pi)]$ and $\tau(x)=\mu_1(x)-\mu_0(x)$. One possible prior correction jointly perturbs $P(x)$, $\mu_1(x)$, $\mu_0(x)$, and $\pi(x)$ using the same $\lambda$:
    \begingroup\makeatletter\def\f@size{9}\check@mathfonts
    \begin{align}
        \tilde P(x) &:= \frac{\exp\left\{\lambda\frac{w(x;\pi)}{\bar w(P)}\big(\tau(x)-\psi_w(P)\big)\right\}}{C_{\psi,w}(P)}P(x), \quad \tilde\mu_1(x):= \mu_1(x) + \lambda \, \frac{\sigma_1^2(x)w(x;\pi)}{\bar w(P)\pi(x)}, \nonumber\\
        \tilde\mu_0(x) &:= \mu_0(x) - \lambda \, \frac{\sigma_0^2(x)w(x;\pi)}{\bar w(P)(1-\pi(x))}, \quad \operatorname{logit}\tilde\pi(x) :=
        \operatorname{logit}\pi(x) + \lambda \, \frac{w'(x;\pi)}{\bar w(P)} \big(\tau(x)-\psi_w(P)\big),
    \end{align}
    where
    \begin{align}
        C_{\psi,w}(P, \lambda) &:= P\left[\exp\left\{\lambda \frac{w(X;\pi)}{\bar w(P)} \big(\tau(X)-\psi_w(P)\big) \right\} \right].
    \end{align}
    \endgroup
    Indeed, at $\lambda=0$, the three corresponding score contributions are
    \begingroup\makeatletter\def\f@size{9}\check@mathfonts
    \begin{align}
        S[\tilde P] &= \frac{w(X;\pi)}{\bar w(P)} \big(\tau(X)-\psi_w(P)\big), \nonumber\\
        S[\tilde P](Y\mid X,A) &= \frac{w(X;\pi)}{\bar w(P)} \frac{A-\pi(X)}{\pi(X)(1-\pi(X))} \big(Y-\mu_A(X)\big), \nonumber\\
        S[\tilde P](A\mid X) &= \big(A-\pi(X)\big) \frac{w'(X;\pi)}{\bar w(P)} \big(\tau(X)-\psi_w(P)\big),
    \end{align}
    \endgroup
    whose sum equals $\phi_{\psi,w}(Z;P)$. Hence, the construction satisfies Eq.~\eqref{eq:prior-corr}. Relative to Example~\ref{ex:b-prior-corr}, the WATE correction perturbs both outcome regressions and, whenever $w$ depends on $\pi$, additionally fluctuates the propensity score along its target-relevant direction. In particular, for the ATE with $w\equiv1$, $w'\equiv0$ and the propensity-score perturbation disappears.
\end{examplec}

%%%%%%%%%%%%%%%%%%%%%%%%%%%%%%%%%%%%%%%%%%%%%%%%%%%%%%%%%%%%%%%%%%%%%%%%%%%%%%%%%%%%%%%%%%
\newpage
\section{Theoretical Results} \label{app:theory}
%%%%%%%%%%%%%%%%%%%%%%%%%%%%%%%%%%%%%%%%%%%%%%%%%%%%%%%%%%%%%%%%%%%%%%%%%%%%%%%%%%%%%%%%%%
Here, we introduce additional notation. For a sequence of random vectors $X_n$ and a positive deterministic sequence $a_n$, we write $X_n=\Omega_P(a_n)$ if $\lVert X_n\rVert_2/a_n$ is bounded away from zero in $P^{\otimes n}$-probability, \ie, for every $\epsilon>0$, there exist constants $c_\epsilon>0$ and $n_\epsilon<\infty$ such that $P^{\otimes n}(\lVert X_n\rVert_2/a_n\ge c_\epsilon)\ge1-\epsilon$ for all $n\ge n_\epsilon$. We write $X_n=\Theta_P(a_n)$ if both $X_n=O_P(a_n)$ and $X_n=\Omega_P(a_n)$. Finally, for two non-negative deterministic sequences $a_n$ and $b_n$, we write $a_n\asymp b_n$ if both $a_n\lesssim b_n$ and $b_n\lesssim a_n$. $\rightsquigarrow$ stands for weak convergence.

\begin{numrem}{1}[Bayes-optimality of the idealized PFN] \label{rem:bayes-optimality-app}
    The posterior mean of an idealized PFN, i.e., $\hat{\psi}_{\theta^*}(\mathcal{D}_n) := {\int \psi \,  q_{\theta^*}(\psi \mid \mathcal{D}_n) \diff{\psi}}$, is Bayes-optimal under the squared $l_2$ risk $\mathcal{R}_{l_2^2}(\hat{\psi}; \Pi,n) := \E_{P\sim\Pi} \E_{\mathcal{D}_n\sim P^{\otimes n}} [\hat{\psi}(\mathcal{D}_n) - \psi(P)]^2$:
    \begingroup\makeatletter\def\f@size{9}\check@mathfonts
    \vspace{-0.1cm}
    {\setlength{\belowdisplayskip}{-0.1cm} \begin{equation}
        \mathcal{R}_{l_2^2}(\hat{\psi}_{\theta^*}; \Pi,n) = \inf_{\hat{\psi}: \mathcal{Z}^{n} \to \Psi} \mathcal{R}_{l_2^2}(\hat{\psi}; \Pi,n) = \E_{P\sim\Pi} \E_{\mathcal{D}_n\sim P^{\otimes n}}[\Var_{\Pi}[\psi(P) \mid \mathcal{D}_n]].
    \end{equation}}
    \endgroup
\end{numrem}
\begin{proof}
    By definition of an idealized PFN, $q_{\theta^*}(\psi \mid \mathcal{D}_n)$ is the density of $\Pi\bigl(\psi(P) \mid \mathcal{D}_n\bigr)$, and hence its posterior mean satisfies
    \begin{align}
        \hat{\psi}_{\theta^*}(\mathcal{D}_n) = \int \psi \, q_{\theta^*}(\psi \mid \mathcal{D}_n)\diff{\psi} = \E_{\Pi}\bigl[\psi(P)\mid\mathcal{D}_n\bigr].
    \end{align}
    Consider any estimator $\hat{\psi}: \mathcal{Z}^{n}\to\Psi$. Conditional on $\mathcal{D}_n$, the squared error admits the usual bias--variance decomposition:
    \begin{align}
        \E_{\Pi}\left[\bigl(\hat{\psi}(\mathcal{D}_n)-\psi(P)\bigr)^2 \middle| \mathcal{D}_n \right] = \left(\hat{\psi}(\mathcal{D}_n) - \E_{\Pi}\bigl[\psi(P)\mid\mathcal{D}_n\bigr] \right)^2 + \Var_{\Pi}\bigl[ \psi(P)\mid\mathcal{D}_n \bigr].
    \end{align}
    The second term does not depend on $\hat{\psi}$, while the first term is non-negative and is uniquely minimized by
    \begin{align}
        \hat{\psi}(\mathcal{D}_n) = \E_{\Pi}\bigl[\psi(P)\mid\mathcal{D}_n\bigr] = \hat{\psi}_{\theta^*}(\mathcal{D}_n).
    \end{align}
    Taking expectations with respect to the joint distribution $P\sim\Pi$ and $\mathcal{D}_n\sim P^{\otimes n}$ therefore yields
    \begin{align}
        \mathcal{R}_{l_2^2}(\hat{\psi}; \Pi,n) &= \mathcal{R}_{l_2^2}(\hat{\psi}_{\theta^*}; \Pi,n) + \E_{P\sim\Pi} \E_{\mathcal{D}_n\sim P^{\otimes n}} \left[\left(\hat{\psi}(\mathcal{D}_n) - \hat{\psi}_{\theta^*}(\mathcal{D}_n) \right)^2 \right] \ge \mathcal{R}_{l_2^2}(\hat{\psi}_{\theta^*}; \Pi,n).
    \end{align}
    Consequently,
    \begin{align}
        \mathcal{R}_{l_2^2}(\hat{\psi}_{\theta^*}; \Pi,n) &= \inf_{\hat{\psi}: \mathcal{Z}^{n}\to\Psi} \mathcal{R}_{l_2^2}(\hat{\psi}; \Pi,n) = \E_{P\sim\Pi} \E_{\mathcal{D}_n\sim P^{\otimes n}} \left[ \Var_{\Pi}\bigl[\psi(P)\mid\mathcal{D}_n\bigr]\right].
    \end{align}
    This proves the claim.
\end{proof}

\begin{numtheorem}{1}[Slow local $L_2$-concentration rates (full version)] \label{thrm:slow-rates-app}
    Let $q_{\theta^*}$ be an idealized PFN, such that $q_{\theta^*}(\psi\mid\mathcal{D}_n)$ equals the push-forward of $\Pi(P\mid\mathcal{D}_n)$ through $\psi(\cdot): P\to\psi(P)$. Fix $P_0\in\operatorname{supp}\Pi$, let $\psi_0:=\psi(P_0)$, and suppose that $\mathcal{D}_n\sim P_0^{\otimes n}$. Assume that 
    \begin{itemize}[leftmargin=0.6cm,itemsep=-0.55mm]
        \item[(a)] for every relevant $P\in\operatorname{supp}\Pi$, the target admits the von--Mises expansion from Eq.~\eqref{eq:von-mises};
        \item[(b)] the first-order term is locally Lipschitz in the nuisance functions: for some constant $C_\psi(P_0)<\infty$,
        \begin{align} \label{eq:lipschitz}
            \abs{P_0\phi_\psi(\cdot;P)} \le C_\psi(P_0) \cdot \norm{\eta(P)-\eta(P_0)}_{\mathcal{H}};
        \end{align}
        \item[(c)] the corresponding posterior second moments are finite.
    \end{itemize}
    Then,
    \begingroup\makeatletter\def\f@size{9}\check@mathfonts
    \begin{align} \label{eq:slow-rates-app}
        \norm{\psi-\psi_0}_{L_2(q_{\theta^*},\mathcal{D}_n)} \lesssim \underbrace{\bigg[\int\norm{\eta(P)-\eta(P_0)}_{\mathcal{H}}^2\, \Pi(\diff P\mid\mathcal{D}_n)\bigg]^{1/2}}_{F_\psi(P_0,\mathcal{D}_n,\Pi)} + \underbrace{\bigg[\int R_{2,\psi}^2(P,P_0) \, \Pi(\diff P\mid\mathcal{D}_n)\bigg]^{1/2}}_{R_{2,\psi}(P_0,\mathcal{D}_n,\Pi)}.
    \end{align}
    \endgroup
    In particular, if for deterministic sequences $\varepsilon_n\to0$ and $r_{2,n}\to0$,
    \begin{align} \label{eq:slow-rates-rate-app}
        F_\psi(P_0,\mathcal{D}_n,\Pi) &= O_{P_0}(\varepsilon_n), \qquad R_{2,\psi}(P_0,\mathcal{D}_n,\Pi) = O_{P_0}(r_{2,n}),
    \end{align}
    then $\norm{\psi-\psi_0}_{L_2(q_{\theta^*},\mathcal{D}_n)} = O_{P_0}(\varepsilon_n+r_{2,n})$.
    Hence, if $r_{2,n}=o(\varepsilon_n)$, the target posterior has concentration error upper-bounded by the first-order nuisance rate $O_{P_0}(\varepsilon_n)$.

    Moreover, this first-order bound is rate-sharp whenever the posterior places asymptotically non-negligible mass on target-relevant nuisance directions. In particular, suppose that 
    \begin{itemize}[leftmargin=0.6cm,itemsep=-0.55mm]
        \item[(d)] there exist constants $0<c_1<c_2<\infty$, $c_3>0$, and $\delta>0$, together with measurable sets $\mathcal{A}_n\subseteq\mathcal{P}(\mathcal{Z})$, such that, with $P_0^{\otimes n}$-probability tending to one,
        \begin{align} \label{eq:sharp-mass-app} 
            \Pi( \mathcal{A}_n \mid \mathcal{D}_n ) &\ge \delta;
        \end{align} 
        \item[(e)] every $P\in\mathcal{A}_n$ satisfies
        \begin{align} \label{eq:sharp-direction-app} 
            c_1\varepsilon_n \le \norm{ \eta(P)-\eta(P_0) }_{\mathcal{H}} \le c_2\varepsilon_n, \quad \text{ and } \quad 
            \abs{ \psi(P)-\psi_0 } \ge c_3 \norm{ \eta(P)-\eta(P_0) }_{\mathcal{H}}; 
        \end{align}
        \item[(f)] $F_\psi(P_0,\mathcal{D}_n,\Pi)=O_{P_0}(\varepsilon_n)$ and $R_{2,\psi}(P_0,\mathcal{D}_n,\Pi)=o_{P_0}(\varepsilon_n)$.
    \end{itemize}
    Then, the following holds: 
    \begin{align} \label{eq:slow-rates-sharp-app} 
        F_\psi(P_0,\mathcal{D}_n,\Pi) &= \Theta_{P_0}(\varepsilon_n), \qquad \text{and } \qquad \norm{ \psi-\psi_0 }_{L_2(q_{\theta^*},\mathcal{D}_n)} = \Theta_{P_0}(\varepsilon_n), 
    \end{align}
    namely, without an explicit first-order correction, the target posterior can inherit the concentration rate of the latent nuisance functions.
    
\end{numtheorem}
\begin{proof}
    Since $q_{\theta^*}$ is an idealized PFN, its target posterior is exactly the push-forward of $\Pi(P\mid\mathcal{D}_n)$ under the map $\psi(\cdot):P\to\psi(P)$. Hence,
    \begin{align} \label{eq:slow-rates-proof}
        \norm{\psi-\psi_0}_{L_2(q_{\theta^*},\mathcal{D}_n)}^2 &= \int \bigl(\psi-\psi_0\bigr)^2 q_{\theta^*}(\psi\mid\mathcal{D}_n) \diff{\psi} = \int \bigl(\psi(P)-\psi_0\bigr)^2 \Pi(\diff P\mid\mathcal{D}_n).
    \end{align}
    Applying the von--Mises expansion from Eq.~\eqref{eq:von-mises} gives
    \begin{align}
        \psi(P)-\psi_0 = -P_0\phi_\psi(\cdot;P) + R_{2,\psi}(P,P_0).
    \end{align}
    Therefore, by the triangle inequality in $L_2(\Pi(\cdot\mid\mathcal{D}_n))$,
    \begin{align} \label{eq:slow-rates-proof2}
        \norm{\psi-\psi_0}_{L_2(q_{\theta^*},\mathcal{D}_n)} &\le \bigg[\int \bigl\{P_0\phi_\psi(\cdot;P) \bigr\}^2 \Pi(\diff P\mid\mathcal{D}_n) \bigg]^{1/2} + \bigg[\int R_{2,\psi}^2(P,P_0) \Pi(\diff P\mid\mathcal{D}_n) \bigg]^{1/2}.
    \end{align}
    
    By Eq.~\eqref{eq:lipschitz},
    \begin{align}
        \bigg[\int \bigl\{P_0\phi_\psi(\cdot;P)\bigr\}^2 \Pi(\diff P\mid\mathcal{D}_n)\bigg]^{1/2} &\le C_\psi(P_0) \bigg[\int \norm{\eta(P)-\eta(P_0)}_{\mathcal{H}}^2 \Pi(\diff P\mid\mathcal{D}_n)\bigg]^{1/2} \nonumber\\
        &= C_\psi(P_0) F_\psi(P_0,\mathcal{D}_n,\Pi).
    \end{align}
    Combining this inequality with Eq.~\eqref{eq:slow-rates-proof2} proves Eq.~\eqref{eq:slow-rates-app}. The rate statement in Eq.~\eqref{eq:slow-rates-rate-app} follows immediately from
    $F_\psi(P_0,\mathcal{D}_n,\Pi)=O_{P_0}(\varepsilon_n)$ and
    $R_{2,\psi}(P_0,\mathcal{D}_n,\Pi)=O_{P_0}(r_{2,n})$.

    It remains to establish rate-sharpness. On the event in Eq.~\eqref{eq:sharp-mass-app}, the definition of $F_\psi$, and the first inequality in Eq.~\eqref{eq:sharp-direction-app} imply 
    \begin{align} 
        F_\psi^2( P_0,\mathcal{D}_n,\Pi ) &= \int \norm{ \eta(P)-\eta(P_0) }_{\mathcal{H}}^2 \Pi( \diff P\mid\mathcal{D}_n ) \nonumber\\ 
        &\ge \int_{\mathcal{A}_n} \norm{ \eta(P)-\eta(P_0) }_{\mathcal{H}}^2 \Pi( \diff P\mid\mathcal{D}_n ) \nonumber\\ 
        &\ge c_1^2 \varepsilon_n^2 \Pi( \mathcal{A}_n \mid \mathcal{D}_n ) \ge c_1^2 \delta \varepsilon_n^2. 
    \end{align}

    Together with the assumed upper bound $F_\psi(P_0,\mathcal{D}_n,\Pi)=O_{P_0}(\varepsilon_n)$, this gives 
    \begin{align} 
        F_\psi( P_0,\mathcal{D}_n,\Pi ) = \Theta_{P_0}(\varepsilon_n). 
    \end{align}

    Similarly, using the push-forward identity and the second inequality in Eq.~\eqref{eq:sharp-direction-app}, \begin{align} 
        \norm{ \psi-\psi_0 }_{L_2(q_{\theta^*},\mathcal{D}_n)}^2 &= \int \bigl( \psi(P)-\psi_0 \bigr)^2 \Pi( \diff P\mid\mathcal{D}_n ) \nonumber\\ 
        &\ge \int_{\mathcal{A}_n} \bigl( \psi(P)-\psi_0 \bigr)^2 \Pi( \diff P\mid\mathcal{D}_n ) \nonumber\\ 
        &\ge c_3^2 \int_{\mathcal{A}_n} \norm{ \eta(P)-\eta(P_0) }_{\mathcal{H}}^2 \Pi( \diff P\mid\mathcal{D}_n ) \nonumber\\ 
        &\ge c_1^2c_3^2 \delta \varepsilon_n^2.
    \end{align}

    Consequently, 
    \begin{align} 
        \norm{ \psi-\psi_0 }_{L_2(q_{\theta^*},\mathcal{D}_n)} = \Omega_{P_0}(\varepsilon_n). 
    \end{align}

    On the other hand, Eq.~\eqref{eq:slow-rates-app}, together with $F_\psi(P_0,\mathcal{D}_n,\Pi)=O_{P_0}(\varepsilon_n)$ and $R_{2,\psi}(P_0,\mathcal{D}_n,\Pi)=o_{P_0}(\varepsilon_n)$, gives 
    \begin{align} 
        \norm{ \psi-\psi_0 }_{L_2(q_{\theta^*},\mathcal{D}_n)} = O_{P_0}(\varepsilon_n).
    \end{align}

    Combining the lower and upper bounds proves Eq.~\eqref{eq:slow-rates-sharp-app}.
\end{proof}

\begin{rem}[Slow concentration rates due to a prior-induced bias (Examples \ref{ex:b}-\ref{ex:c})] \label{rem:pib}
    Consider Example~\ref{ex:b} with fixed $\pi(x)=\pi_0(x)$, while $\mu(x)$ and $P(x)$  remain unknown. Analogously to \citet{melnychuk2026frequentist}, define the degree of observed selection bias as
    $\Delta(P):=\psi(P)-m(P)$, where $m(P):=\E_P[RY\mid R=1]=P[\pi\mu]/P[\pi]$.
    Then, for any fixed $P_0$ and $\mathcal{D}_n$, we have
    \begin{equation} \label{eq:pib-first-order}
        F_\psi(P_0,\mathcal{D}_n,\Pi)
        \geq
        \frac{\kappa}{\kappa+2}
        \bigg[ \abs{\Delta(P_0)} - \int \abs{\Delta(P)}\,\Pi(\diff P\mid\mathcal{D}_n) \bigg]_+,
    \end{equation}
    where $[a]_+:=\max\{a,0\}$. Hence, if the prior induces a posterior concentrated on nearly unconfounded distributions (\ie, $\int \abs{\Delta(P)}\,\Pi(\diff P\mid\mathcal{D}_n) \approx 0$) despite substantial selection bias under $P_0$ (\ie, $\abs{\Delta(P_0)} \gg 0$), the first-order error cannot vanish.
\end{rem}
\begin{proof}
    Since $\pi(x)=\pi_0(x)$ is known and fixed, we have the following equalities:
    \begin{align}
        \Delta(P)-\Delta(P_0) &= \psi(P) - \psi(P_0) -\bigg(  \frac{P[\pi_0 \mu ]}{P[\pi_0]} - \frac{P_0[\pi_0 \mu_0]}{P_0[\pi_0]} \bigg), \\
        \psi(P) - \psi(P_0) &= (P - P_0)\mu + P_0[\mu - \mu_0], 
    \end{align}
    where the second follows from Eq.~\eqref{eq:ex-b-slow-rates} and the fact that $R_{2, \psi}(P, P_0) = 0$ when the labeling score is known. Also, by the Cauchy–Schwarz inequality and boundedness of the outcome, the following inequalities hold:
    \begin{align}
        \abs{\frac{P[\pi_0 \mu ]}{P[\pi_0]} - \frac{P_0[\pi_0 \mu_0]}{P_0[\pi_0]}} &\le {\frac{\abs{(P-P_0)[\pi_0 \mu] + P_0[\pi_0(\mu - \mu_0)]}}{P[\pi_0]}} + \frac{P_0[\pi_0\mu_0]}{P_0[\pi_0]} \frac{\abs{(P-P_0)[\pi_0]}}{P[\pi_0]} \\
        & \le \frac{2 \norm{P - P_0}_{L_2(x)} + \norm{\mu-\mu_0}_{L_2(P_0)}}{\kappa}, \nonumber \\
        \abs{\psi(P) - \psi(P_0)} &\le \norm{P - P_0}_{L_2(x)} + \norm{\mu-\mu_0}_{L_2(P_0)}.
    \end{align}
    Thus, we have the following:
    \begin{align}
        \abs{\Delta(P)-\Delta(P_0)} &\le \Big(1 + \frac{2}{\kappa}\Big) \norm{P - P_0}_{L_2(x)} +\Big(1 + \frac{1}{\kappa}\Big) \norm{\mu-\mu_0}_{L_2(P_0)} \\
        &\le \frac{\kappa + 2}{\kappa} \cdot \big(\norm{P - P_0}_{L_2(x)} + \norm{\mu-\mu_0}_{L_2(P_0)} \big).
    \end{align}
    
    Integrating over $\Pi(\diff P\mid\mathcal{D}_n)$ and applying triangle inequality gives
    \begin{align}
        \abs{\Delta(P_0)}
        &\leq \int\abs{\Delta(P)}\,\Pi(\diff P\mid\mathcal{D}_n) + \frac{\kappa + 2}{\kappa} F_\psi(P_0,\mathcal{D}_n,\Pi).
    \end{align}
    Finally, by rearranging the terms of the latter, we get the desired Eq.~\eqref{eq:pib-first-order}.
\end{proof}

\begin{numtheorem}{2}[$L_2$-concentration rates of debiased PFNs (full version)] \label{thrm:debiased-rates-app}
    Assume the conditions (a)-(c) of Theorem~\ref{thrm:slow-rates-app}. Fix $P_0\in\operatorname{supp}\Pi$, let $\psi_0:=\psi(P_0)$, and suppose that $\mathcal{D}_n\sim P_0^{\otimes n}$. Write $P_n f:=n^{-1}\sum_{i=1}^n f(Z_i)$ and $\phi_P:=\phi_\psi(\cdot;P)$, with $\phi_0:=\phi_\psi(\cdot;P_0)$. Assume that $\phi_0\in L_2(P_0)$ and that 
    \begin{itemize}[leftmargin=0.6cm,itemsep=-0.55mm]
        \item[(g)] the posterior is stochastically equicontinuous in the sense that 
        \begin{align} \label{eq:eqcont}
            \bigg[\int\big\{(P_n-P_0)(\phi_P-\phi_0)\big\}^2 \Pi_\star(\diff P\mid\mathcal{D}_n)\bigg]^{1/2} = o_{P_0}(n^{-1/2}),
        \end{align}
        where $\Pi_\star=\Pi$ for the label correction and $\Pi_\star$ denotes the $P$-marginal of the corrected posterior for the prior correction;
        \item[(h-i)] for strategy (i), the following holds 
        \begin{align} \label{eq:bb-moment-app} 
            \int P_n\left[ \phi_P^2 \right] \Pi( \diff P\mid\mathcal{D}_n ) = O_{P_0}(1); 
        \end{align} 
        \item[(h-ii)] for strategy (ii), let $\tilde P=\operatorname{LF}_\psi(P;\lambda)$, define $V_\psi(P) := P\left[ \phi_P^2 \right]$,
        and suppose that the least-favorable path is locally regular so that a Taylor expansion wrt. $\lambda$ holds:
        \begin{align} \label{eq:lf-target-expansion-app} 
            \psi\big( \operatorname{LF}_\psi(P;\lambda) \big) - \psi(P) = \lambda V_\psi(P) + r_{\operatorname{LF},\psi}(P,\lambda), 
        \end{align}
        where $r_{\operatorname{LF},\psi}(P,\lambda)$ is a second-order remainder; and the standard local asymptotic normality and prior-flatness conditions hold along this path, including $n\sigma_n^2\to\infty$, such that the joint corrected posterior satisfies 
        \begin{align} \label{eq:lambda-posterior-app} 
            \bigg[ \int \big\{ \lambda V_\psi(P) - P_n\phi_P \big\}^2 \, \widetilde\Pi( \diff P\times\diff\lambda \mid \mathcal{D}_n ) \bigg]^{1/2} = O_{P_0}(n^{-1/2}); 
        \end{align}
        
    \end{itemize}
    Then, for either debiasing strategy (i)--(ii), the $L_2$-concentration error of the corresponding idealized debiased PFN satisfies 
    \begin{align} \label{eq:debiased-ratesa-app} 
        \norm{ \tilde{\psi}-\psi_0 }_{L_2(\tilde q_{\theta^*},\mathcal{D}_n)} &\le \underbrace{ \widetilde F_\psi( P_0, \mathcal{D}_n, \star ) }_{ O_{P_0}(n^{-1/2}) } + R_{2,\psi}( P_0, \mathcal{D}_n, \star ), \qquad \star= \begin{cases} \Pi, & \text{(i) label correction},\\ \widetilde\Pi, & \text{(ii) prior correction}, \end{cases} 
    \end{align}
    where the corrected first-order terms are 
    \begin{align} \label{eq:corrected-first-order-app} 
        \widetilde F_\psi( P_0, \mathcal{D}_n, \Pi ) &:= \bigg[ \int \E_W\left[ \big\{ (\operatorname{BB}_n-P_0)\phi_P \big\}^2 \,\middle|\, \mathcal{D}_n,P \right] \Pi( \diff P\mid\mathcal{D}_n ) \bigg]^{1/2}, \\ 
        \widetilde F_\psi( P_0, \mathcal{D}_n, \widetilde\Pi ) &:= \bigg[ \int \big\{ (P_n-P_0)\phi_P + \lambda V_\psi(P) - P_n\phi_P \big\}^2 \, \widetilde\Pi( \diff P \times \diff\lambda \mid \mathcal{D}_n ) \bigg]^{1/2}. 
    \end{align}

    The corresponding posterior-averaged second-order remainders are 
    \begin{align} \label{eq:debiased-remainders-app} 
        R_{2,\psi}( P_0, \mathcal{D}_n, \Pi ) &:= \bigg[ \int R_{2,\psi}^2(P,P_0) \Pi( \diff P\mid\mathcal{D}_n ) \bigg]^{1/2}, \\ 
        R_{2,\psi}( P_0, \mathcal{D}_n, \widetilde\Pi ) &:= \bigg[ \int \big\{ R_{2,\psi}(P,P_0) + r_{\operatorname{LF},\psi}(P,\lambda) \big\}^2 \, \widetilde\Pi( \diff P\times \diff\lambda \mid \mathcal{D}_n ) \bigg]^{1/2}.
    \end{align}
\end{numtheorem}

\begin{proof}
    We consider the two correction strategies separately.

    \textbf{(i) Label correction.}
    Recall from Eq.~\eqref{eq:label-corr} that
    \begin{align}
        \tilde{\psi}(P) = \psi(P) + \operatorname{BB}_n \left\{\phi_P \right\}.
    \end{align}
    Since the target functional satisfies the von--Mises expansion,
    \begin{align}
        \psi(P)-\psi_0 = -P_0\phi_P + R_{2,\psi}(P,P_0),
    \end{align}
    we obtain the exact decomposition
    \begin{align} \label{eq:label-decomposition-proof}
        \tilde{\psi}(P)-\psi_0 = (\operatorname{BB}_n-P_0)\phi_P + R_{2,\psi}(P,P_0).
    \end{align}
    For an idealized label-corrected PFN, $\tilde q_{\theta^*}$ is the conditional distribution induced by drawing $P\sim\Pi(P\mid\mathcal{D}_n)$ and independent Bayesian-bootstrap weights $W$. Hence, by Eq.~\eqref{eq:label-decomposition-proof} and the triangle inequality in the corresponding $L_2$ space,
    \begin{align}
        \norm{\tilde{\psi}-\psi_0}_{L_2(\tilde q_{\theta^*},\mathcal{D}_n)} &\le \widetilde F_\psi(P_0, \mathcal{D}_n, \Pi) + R_{2,\psi}(P_0, \mathcal{D}_n,\Pi).
    \end{align}

    It remains to show that the first term is of order $n^{-1/2}$. Decompose
    \begin{align} \label{eq:bb-first-order-decomp}
        (\operatorname{BB}_n-P_0)\phi_P &= (\operatorname{BB}_n-P_n)\phi_P + (P_n-P_0)\phi_0 + (P_n-P_0)(\phi_P-\phi_0).
    \end{align}
    Applying the $L_2$ triangle inequality to Eq.~\eqref{eq:bb-first-order-decomp} gives
    \begin{align}
        \widetilde F_\psi(P_0, \mathcal{D}_n, \Pi) &\le \bigg[\int \E_W\left[ \big\{ (\operatorname{BB}_n-P_n)\phi_P \big\}^2 \,\middle|\, \mathcal{D}_n,P \right] \Pi( \diff P\mid\mathcal{D}_n)\bigg]^{1/2} \nonumber\\
        & \qquad + \abs{(P_n-P_0)\phi_0} \nonumber\\
        & \qquad + \bigg[\int \big\{(P_n-P_0)(\phi_P-\phi_0)\big\}^2 \Pi(\diff P\mid\mathcal{D}_n)\bigg]^{1/2}.
    \end{align}
    For Bayesian-bootstrap weights $W\sim\operatorname{Dir}(n;1,\ldots,1)$,
    \begin{align}
        \E_W\left[\big\{(\operatorname{BB}_n-P_n)\phi_P\big\}^2 \,\middle|\, \mathcal{D}_n,P \right] = \frac{ P_n\left[\big( \phi_P-P_n\phi_P \big)^2\right]}{n+1} \le \frac{P_n[\phi_P^2]}{n+1}.
    \end{align}
    Therefore, Eq.~\eqref{eq:bb-moment-app} implies that the first term is $O_{P_0}(n^{-1/2})$. Since $\phi_0\in L_2(P_0)$,
    \begin{align}
        (P_n-P_0)\phi_0 = O_{P_0}(n^{-1/2}),
    \end{align}
    by the central limit theorem, while the final term is $o_{P_0}(n^{-1/2})$ by Eq.~\eqref{eq:eqcont}. Consequently,
    \begin{align}
        \widetilde F_\psi(P_0, \mathcal{D}_n, \Pi) = O_{P_0}(n^{-1/2}).
    \end{align}

    \textbf{(ii) Prior correction.}
    Write $\tilde P = \operatorname{LF}_\psi(P;\lambda)$.
    By construction of the least-favorable path, 
    \begin{align} 
        \left. \frac{\partial}{ \partial\lambda } \log \operatorname{LF}_\psi(P;\lambda) \right|_{\lambda=0} = \phi_P. 
    \end{align} 
    Pathwise differentiability of $\psi$ therefore gives 
    \begin{align} 
        \left. \frac{\partial}{ \partial\lambda } \psi\big( \operatorname{LF}_\psi(P;\lambda) \big) \right|_{\lambda=0} &= P\left[ \phi_P \left. \frac{\partial}{ \partial\lambda } \log \operatorname{LF}_\psi(P;\lambda) \right|_{\lambda=0} \right] = P[ \phi_P^2 ] = V_\psi(P). 
    \end{align} 
    Hence Eq.~\eqref{eq:lf-target-expansion-app} is the first-order expansion of the target along the correction path. Combining this expansion with the von--Mises expansion around $P_0$ yields 
    \begin{align} 
        \psi(\tilde P)-\psi_0 &= -P_0\phi_P + \lambda V_\psi(P) + R_{2,\psi}(P,P_0) + r_{\operatorname{LF},\psi}(P,\lambda) \nonumber\\ &= (P_n-P_0)\phi_P + \big\{ \lambda V_\psi(P) - P_n\phi_P \big\} + R_{2,\psi}(P,P_0) + r_{\operatorname{LF},\psi}(P,\lambda). 
    \end{align} 
    Since an idealized prior-corrected PFN recovers the push-forward of the joint posterior $\widetilde\Pi(P,\lambda\mid\mathcal{D}_n)$ through $(P,\lambda)\mapsto\psi(\operatorname{LF}_\psi(P;\lambda))$, the $L_2$ triangle inequality implies 
    \begin{align} 
        \norm{ \tilde{\psi}-\psi_0 }_{L_2(\tilde q_{\theta^*},\mathcal{D}_n)} &\le \widetilde F_\psi( P_0, \mathcal{D}_n, \widetilde\Pi ) + R_{2,\psi}( P_0, \mathcal{D}_n, \widetilde\Pi ). 
    \end{align}
    To bound the first-order contribution, write 
    \begin{align} 
        (P_n-P_0)\phi_P &= (P_n-P_0)\phi_0 + (P_n-P_0)(\phi_P-\phi_0). 
    \end{align} 
    Hence, 
    \begin{align} 
        \widetilde F_\psi( P_0, \mathcal{D}_n, \widetilde\Pi ) &\le \abs{ (P_n-P_0)\phi_0 } + \bigg[ \int \big\{ (P_n-P_0)(\phi_P-\phi_0) \big\}^2 \widetilde\Pi( \diff P\times\diff\lambda \mid \mathcal{D}_n ) \bigg]^{1/2} \nonumber\\ &\qquad+ \bigg[ \int \big\{ \lambda V_\psi(P) - P_n\phi_P \big\}^2 \widetilde\Pi( \diff P\times\diff\lambda \mid \mathcal{D}_n ) \bigg]^{1/2}. 
    \end{align}
    The first term is $O_{P_0}(n^{-1/2})$ by the central limit theorem, the second is $o_{P_0}(n^{-1/2})$ by Eq.~\eqref{eq:eqcont}, and the third is $O_{P_0}(n^{-1/2})$ by Eq.~\eqref{eq:lambda-posterior-app}. Thus, 
    \begin{align} 
        \widetilde F_\psi( P_0, \mathcal{D}_n, \widetilde\Pi ) = O_{P_0}(n^{-1/2}). 
    \end{align} 
    This establishes Eq.~\eqref{eq:debiased-ratesa-app} for both correction strategies.
\end{proof}

\begin{numtheorem}{3}[Bernstein--von Mises property for debiased PFNs] \label{thrm:bbvm-app}
    Assume the conditions (a)-(c) of Theorem~\ref{thrm:slow-rates-app} and conditions (g)-(h) of Theorem~\ref{thrm:debiased-rates-app}. Let
    $\phi_P:=\phi_\psi(\cdot;P)$, $\phi_0:=\phi_\psi(\cdot;P_0)$, and
    $V_0 := V_{\mathrm{eff}}(P_0) = P_0[\phi_0^2] = \Var_{P_0}[\phi_0] \in(0,\infty)$.
    Suppose that 
    \begin{itemize}[leftmargin=0.6cm,itemsep=-0.55mm]
        \item[(j)] the corresponding posterior-averaged second-order remainder is negligible at the parametric scale,
        \begin{align} \label{eq:second-order-small}
            R_{2,\psi}(P_0,\mathcal{D}_n,\star) = o_{P_0}(n^{-1/2}), \qquad \star=
            \begin{cases}
                \Pi, & \text{(i) label correction},\\
                \widetilde\Pi, & \text{(ii) prior correction};
            \end{cases}
        \end{align}
        \item[(k-i)] for strategy (i), the posterior is consistent
        in the empirical variance of the efficient influence function:
        \begin{align}\label{eq:eif-bvm-app}
            \int \operatorname{Var}_{P_n}[\phi_P-\phi_0]\,
            \Pi(\diff P\mid\mathcal D_n)
            = o_{P_0}(1).
        \end{align}
        \item[(k-ii)] for strategy (ii), with a slight abuse of notation, let $\widetilde \Pi(\diff P\mid\mathcal D_n)$ simultaneously denote the $P$-marginal of $\widetilde\Pi$, and define
        $\ell_n(P,\lambda) := \sum_{i=1}^n \log \operatorname{LF}_\psi(P;\lambda)(Z_i), \widehat\lambda_n(P) = \frac{P_n[\phi_P]}{V_\psi(P)}$. Assume there exist measurable sets $\mathcal H_n\subseteq\mathcal P$ such that
        \begin{align}\label{eq:bvm-localization-app}
            \widetilde \Pi(\mathcal H_n^c\mid\mathcal D_n) =o_{P_0}(1), \quad \text{ and } \quad  
            \sup_{P\in\mathcal H_n} |V_\psi(P)-V_0| =o_{P_0}(1).
        \end{align}
        Furthermore, for every fixed $M<\infty$, assume uniform local asymptotic normality along the least-favorable paths:
        \begin{align}\label{eq:bvm-lan-app}
            \sup_{\substack{P\in\mathcal H_n\\ |t|\le M}} \bigg| \ell_n\left(P,\widehat\lambda_n(P) +\frac{t}{\sqrt n}\right) -\ell_n(P,\widehat\lambda_n(P)) +\frac{t^2}{2}V_\psi(P)\bigg| =o_{P_0}(1).
        \end{align}
        Finally, assume local flatness of the normal perturbation prior:
        \begin{align}\label{eq:bvm-flatness-app}
            n\sigma_n^2&\to\infty, \qquad \sup_{P\in\mathcal H_n} \frac{|\widehat\lambda_n(P)|}{\sqrt n\,\sigma_n^2} =o_{P_0}(1).
        \end{align}
    \end{itemize}

    Then, for either debiasing strategy (i) or (ii),
    \begin{align} \label{eq:bvm-app}
        d_{\mathrm{TV}} \left[\widetilde Q_{\theta^*} \left(\sqrt{n}(\psi-\psi_0) \mid \mathcal{D}_n \right); N\left(\Delta_n, V_{\mathrm{eff}}(P_0)\right) \right] \xrightarrow{P_0} 0,
    \end{align}
    where $\Delta_n:=\sqrt{n}P_n[\phi_\psi(\cdot;P_0)]$ is the asymptotically efficient random centering term. Hence, the debiased PFN posterior asymptotically \textbf{attains the semi-parametric efficiency bound}.
    Moreover, given the asymptotic normality of $\Delta_n$ ($\Delta_n \rightsquigarrow N(0,V_{\mathrm{eff}}(P_0))$), a posterior draw $\tilde{\psi} \sim \widetilde Q_{\theta^*}(\cdot \mid \mathcal{D}_n)$ is also asymptotically normal under repeated sampling of $\mathcal{D}_n \sim P_0^{\otimes n}$:
    \begin{align} \label{eq:bvm-as-app}
        \sqrt{n}(\tilde{\psi}-\psi_0)
        \rightsquigarrow N(0,2V_{\mathrm{eff}}(P_0)),
    \end{align}
    where convergence is with respect to the joint law of $\mathcal{D}_n$ and $\tilde{\psi}$. The factor two arises from adding sampling variation in the random posterior center $\Delta_n$ and conditional posterior variation $\tilde{\psi}$.
\end{numtheorem}

\begin{proof}
    Since $\phi_0$ is an efficient influence function, $P_0[\phi_0]=0$. Thus, $\Delta_n=\sqrt n(P_n-P_0)\phi_0$.

    We first establish a common decomposition for both correction strategies. By the von--Mises expansion and Eq.~\eqref{eq:lf-target-expansion-app}, for $\tilde{\psi} \sim \widetilde Q_{\theta^*}(\psi \mid \mathcal{D}_n)$:
    \begin{align}\label{eq:bvm-common-decomp-app}
        \sqrt n(\tilde\psi-\psi_0)
        =\Delta_n+U_n^\star+A_n^\star,
    \end{align}
    where
    \begin{align}
        U_n^\star &:=
        \begin{cases}
            \sqrt n(\operatorname{BB}_n-P_n)\phi_P, &\text{(i)},\\
            \sqrt n\{\lambda V_\psi(P)-P_n\phi_P\}, &\text{(ii)},
        \end{cases}\\
        A_n^\star &:=\sqrt n(P_n-P_0)(\phi_P-\phi_0) +\sqrt n\,\mathfrak r_\psi^\star,\\
        \mathfrak r_\psi^\star &:=
        \begin{cases}
            R_{2,\psi}(P,P_0),&\text{(i)},\\
            R_{2,\psi}(P,P_0) +r_{\operatorname{LF},\psi}(P,\lambda), &\text{(ii)}.
        \end{cases}
    \end{align}

    Write $\mathbb E_\star[\cdot\mid\mathcal D_n]$ for expectation under the corresponding corrected posterior, including the independent Bayesian-bootstrap weights for strategy (i).

    By the $L_2$ triangle inequality,
    \begin{align}\label{eq:bvm-remainder-proof}
        &\left\{\mathbb E_\star[(A_n^\star)^2\mid\mathcal D_n]\right\}^{1/2}\le \sqrt n\bigg[\int \{(P_n-P_0)(\phi_P-\phi_0)\}^2 \Pi_\star(\diff P\mid\mathcal D_n)\bigg]^{1/2} \\
        &\qquad + \sqrt n R_{2,\psi}(P_0,\mathcal D_n,\star) = o_{P_0}(1), \nonumber
    \end{align}
    by condition (g) and Eq.~\eqref{eq:second-order-small}. In particular, $\mathbb E_\star[|A_n^\star|\mid\mathcal D_n] =o_{P_0}(1)$.

    \textbf{(i) Label correction.}
    Define
    \begin{align}
        B_n &:=\sqrt n(\operatorname{BB}_n-P_n)\phi_0, \quad \text{ and } \quad E_n(P) :=\sqrt n(\operatorname{BB}_n-P_n)(\phi_P-\phi_0).
    \end{align}
    Then $U_n^\Pi=B_n+E_n(P)$.

    The conditional Bayesian-bootstrap central limit theorem gives
    \begin{align}\label{eq:bvm-bb-clt-proof}
        d_{\mathrm{BL}}\left[\mathcal L_W(B_n\mid\mathcal D_n); N(0,V_0) \right]=o_{P_0}(1),
    \end{align}
    where $d_{\mathrm{BL}}$ is a bounded-Lipschitz distance, and $\mathcal L_W$ is a probability distribution of $B_n$.
    Indeed, writing the Dirichlet weights as normalized independent standard exponential variables, this follows from the conditional Lindeberg central limit theorem, since $\phi_0\in L_2(P_0)$ and $\operatorname{Var}_{P_n}[\phi_0]\to V_0$. Furthermore, by Eq.~\eqref{eq:eif-bvm-app}, the Dirichlet variance identity implies
    \begin{align}
        &\int \mathbb E_W[E_n(P)^2\mid P,\mathcal D_n]\,\Pi(\diff P\mid\mathcal D_n) = \frac{n}{n+1}
        \int \operatorname{Var}_{P_n}[\phi_P-\phi_0] \Pi(\diff P\mid\mathcal D_n) =o_{P_0}(1).
    \end{align}

    For each fixed $P$, let $\nu_{n,P}$ denote
    the conditional law, over the bootstrap weights, of
    $\sqrt n(\tilde\psi(P)-\psi_0)-\Delta_n$.
    Combining the preceding bounds with
    Eq.~\eqref{eq:bvm-remainder-proof} gives
    \begin{align}
        &\int d_{\mathrm{BL}}\left[\nu_{n,P}; N(0,V_0) \right]\Pi(\diff P\mid\mathcal D_n) \le d_{\mathrm{BL}}\left[ \mathcal L_W(B_n\mid\mathcal D_n); N(0,V_0)\right] + o_{P_0}(1) =o_{P_0}(1).
    \end{align}

    Now, we note that, conditionally on $P$, the corrected target $\tilde\psi(P)=\psi(P)+\operatorname{BB}_n\{\phi_P\}$ is an affine transformation of the Dirichlet weights. Since $\operatorname{Dirichlet}(1,\ldots,1)$ is uniform on the probability simplex, $\nu_{n,P}$ is log-concave. Then, weak convergence of log-concave laws to a nondegenerate normal distribution implies convergence in
    total variation. This implication is uniform in a neighborhood of the fixed normal distribution $N(0, V_0)$. Consequently, the preceding posterior-averaged bounded-Lipschitz convergence yields
    \begin{align}\label{eq:bvm-label-tv-proof}
        \int d_{\mathrm{TV}}\left[\nu_{n,P}; N(0,V_0)\right]\Pi(\diff P\mid\mathcal D_n) =o_{P_0}(1).
    \end{align}
    By convexity of total variation under mixtures,
    \begin{align}
        d_{\mathrm{TV}}\left[\widetilde Q_{\theta^*}(U_n^\Pi+A_n^\Pi\mid\mathcal D_n); N(0,V_0)\right] = o_{P_0}(1).
    \end{align}
    Together with Eq.~\eqref{eq:bvm-common-decomp-app}, this proves Eq.~\eqref{eq:bvm-app} for strategy (i).

    \textbf{(ii) Prior correction.} Define the local perturbation parameter
    \begin{align}
        T_n &:=\sqrt n\{\lambda-\widehat\lambda_n(P)\}, \quad \text{ and } \quad \widehat\lambda_n(P) :=\frac{P_n[\phi_P]}{V_\psi(P)}.
    \end{align}
    Then
    \begin{align}\label{eq:bvm-lambda-reparam-proof}
        U_n^{\widetilde\Pi}=V_\psi(P) \cdot T_n.
    \end{align}

    Let $g_n$ denote the density of $N(0,\sigma_n^2)$. By the independent prior specification in Eq.~\eqref{eq:prior-corr}, the conditional posterior density of $T_n$ given $P$ and $\mathcal D_n$ is proportional to
    \begin{align}
        \exp\bigg\{\ell_n\left(P,\widehat\lambda_n(P) +\frac{t}{\sqrt n} \right) -\ell_n(P,\widehat\lambda_n(P))\bigg\}
        \cdot \frac{g_n(\widehat\lambda_n(P)+t/\sqrt n)}{g_n(\widehat\lambda_n(P))}.
    \end{align}
    For the normal perturbation prior,
    \begin{align}
        \log \frac{g_n(\widehat\lambda_n(P)+t/\sqrt n)}{g_n(\widehat\lambda_n(P))} &= -\frac{t\widehat\lambda_n(P)}{\sqrt n\sigma_n^2} -\frac{t^2}{2n\sigma_n^2}.
    \end{align}

    Therefore, Eq.~\eqref{eq:bvm-lan-app} and~\eqref{eq:bvm-flatness-app} imply, uniformly over $P\in\mathcal H_n$ and bounded $t$, that the conditional posterior density of $T_n$ is proportional to $\exp\left\{-\frac{t^2}{2}V_\psi(P) +o_{P_0}(1) \right\}$.

    We next establish posterior localization. By Eq.~\eqref{eq:bvm-localization-app}, $V_\psi(P)\ge V_0/2$ on $\mathcal H_n$ with probability tending to one. Hence, for every $M>0$,
    \begin{align}
        &\widetilde\Pi(|T_n|>M\mid\mathcal D_n) \le \widetilde \Pi(\mathcal H_n^c\mid\mathcal D_n) + \frac{4}{M^2V_0^2} \int (U_n^{\widetilde\Pi})^2 \widetilde\Pi(\diff P\times\diff\lambda \mid\mathcal D_n) \nonumber\\
        &\quad=o_{P_0}(1)+O_{P_0}(M^{-2}),
    \end{align}
    where the last step follows from Eq.~\eqref{eq:lambda-posterior-app}. Thus, the local posterior is tight.

    Combining this localization with the uniform quadratic likelihood expansion and prior flatness, the standard truncated-density Laplace argument gives
    \begin{align}\label{eq:bvm-lambda-tv-proof}
        &\int d_{\mathrm{TV}}\bigg[\mathcal L_{\widetilde\Pi}(T_n\mid P,\mathcal D_n); N(0,V_\psi(P)^{-1}) \bigg]\, \widetilde \Pi(\diff P\mid\mathcal D_n) =o_{P_0}(1),
    \end{align}
    where $ L_{\widetilde\Pi}$ is a probability distribution of $T_n$.
    Together with $\sup_{P\in\mathcal H_n}|V_\psi(P)-V_0|=o_{P_0}(1)$ and Eq.~\eqref{eq:bvm-lambda-reparam-proof}, this establishes
    \begin{align}
        d_{\mathrm{TV}}\left[\mathcal L_{\widetilde\Pi}(U_n^{\widetilde\Pi}\mid\mathcal D_n); N(0,V_0)\right] =o_{P_0}(1).
    \end{align}

    It remains to transfer this convergence to the corrected target, accounting for the
    possibly $\lambda$-dependent remainder. For each $P$, define
    \begin{align}
        b_{n,P}(t) :=\sqrt n\bigg\{&(P_n-P_0)(\phi_P-\phi_0) + R_{2,\psi}(P,P_0) + r_{\operatorname{LF},\psi} \left(P,\widehat\lambda_n(P) +\frac{t}{\sqrt n} \right) \bigg\}.
    \end{align}
    Then the common decomposition becomes
    \begin{align}
        \sqrt n(\tilde\psi-\psi_0) =\Delta_n+V_\psi(P) \cdot T_n+b_{n,P}(T_n).
    \end{align}

    By Eq.~\eqref{eq:bvm-remainder-proof},
    \begin{align}
        \int b_{n,P}(T_n)^2\, \widetilde\Pi(\diff P\times\diff\lambda \mid\mathcal D_n) =o_{P_0}(1).
    \end{align}
    Moreover, Eq.~\eqref{eq:bvm-lan-app} implies, for every fixed $M<\infty$,
    \begin{align}
        \sup_{\substack{P\in\mathcal H_n\\|t|\le M}} |b_{n,P}'(t)| =o_{P_0}(1).
    \end{align}

    We claim that, for every fixed $M,\epsilon>0$,
    \begin{align}\label{eq:bvm-uniform-rem-proof}
        \widetilde \Pi\left(\sup_{|t|\le M}|b_{n,P}(t)|>\epsilon \,\middle|\,\mathcal D_n \right)=o_{P_0}(1).
    \end{align}
    To see this, choose a fixed $M_0>0$. Eq.~\eqref{eq:bvm-lambda-tv-proof} and the positivity of $V_0$ imply that there exist $c>0$ and sets $\mathcal G_n\subseteq\mathcal H_n$ with posterior probability tending to one, such that
    \begin{align}
        \widetilde\Pi(|T_n|\le M_0 \mid P,\mathcal D_n )\ge c
    \end{align}
    for every $P\in\mathcal G_n$. By the mean-value theorem, for such $P$,
    \begin{align}
        |b_{n,P}(0)| &\le \frac{1}{c} \mathbb E_{\widetilde\Pi}\left[|b_{n,P}(T_n)| \mid P,\mathcal D_n \right] + M_0\sup_{|t|\le M_0}|b_{n,P}'(t)|.
    \end{align}
    Integrating over $\mathcal G_n$ and using the posterior $L_2$ bound shows that $b_{n,P}(0)=o_{P_0}(1)$ in $\widetilde \Pi$-posterior probability. The derivative bound then proves Eq.~\eqref{eq:bvm-uniform-rem-proof}.

    Now, consider the transformation $h_{n,P}(t) :=V_\psi(P) \cdot t+b_{n,P}(t)$. On posterior-relevant $P$, this transformation converges locally in $C^1$ to $t\mapsto V_\psi(P) \cdot t$. Since $V_\psi(P)$ is bounded away from zero, $h_{n, P}$ is locally strictly increasing. The change-of-variables formula for densities, together with normal tail truncation, therefore implies that the push-forward of $N(0, V_\psi(P)^{-1})$ through $h_{n, P}$ approaches $N(0, V_\psi(P))$ in total variation, in $\widetilde \Pi$-posterior probability.

    Combining this fact with Eq.~\eqref{eq:bvm-lambda-tv-proof}, using contraction of total variation under measurable transformations,
    and then using uniform convergence $V_\psi(P)\to V_0$, gives
    \begin{align}
        d_{\mathrm{TV}}\left[\widetilde Q_{\theta^*}(U_n^{\widetilde\Pi} + A_n^{\widetilde\Pi} \mid\mathcal D_n); N(0,V_0) \right] =o_{P_0}(1).
    \end{align}
    In view of Eq.~\eqref{eq:bvm-common-decomp-app}, this proves Eq.~\eqref{eq:bvm-app} for strategy (ii).

    \textbf{Repeated-sampling asymptotic normality.}
    By the classical central limit theorem, $\Delta_n\rightsquigarrow N(0, V_0)$. Together with Eq.~\eqref{eq:bvm-app}, this implies the joint convergence
    \begin{align}
        \left(\Delta_n,\sqrt n(\tilde\psi-\psi_0)-\Delta_n\right) \rightsquigarrow (Z_1,Z_2),
    \end{align}
    where $Z_1,Z_2$ are independent $N(0,V_0)$ random variables. Hence, by the continuous mapping theorem,
    \begin{align}
        \sqrt n(\tilde\psi-\psi_0) \rightsquigarrow Z_1+Z_2 \sim N(0,2V_0),
    \end{align}
    which establishes Eq.~\eqref{eq:bvm-as-app}.
\end{proof}

\begin{numtheorem}{4}[Rate optimality of debiased PFNs] \label{thrm:minimaxa-app}
    Assume the conditions (a)-(c) of Theorem~\ref{thrm:slow-rates-app} and conditions (g)-(h) of Theorem~\ref{thrm:debiased-rates-app}, and consider an effective prior-supported class
    $\mathcal P_n^\Pi:=\operatorname{supp}(\Pi)\cap\mathcal P_n^{\mathrm{SA}}$
    with $\Pi(\mathcal P_n^\Pi)\to1$. Suppose that 
    \begin{itemize}[leftmargin=0.6cm,itemsep=-0.55mm]
        \item[(l)] the first- and second-order terms in Eq.~\eqref{eq:debiased-ratesa-app} of Theorem~\ref{thrm:debiased-rates-app} satisfy uniformly over $\mathcal P_n^\Pi$,
        \begin{align} \label{eq:uniform-first-second}
            \sup_{P\in\mathcal P_n^\Pi} \E_{\mathcal D_n\sim P^{\otimes n}} \left[ \widetilde F_\psi^2(P,\mathcal D_n,\star) \right] &\lesssim
            n^{-1}, \quad \sup_{P\in\mathcal P_n^\Pi} \E_{\mathcal D_n\sim P^{\otimes n}} \left[ R_{2,\psi}^2(P,\mathcal D_n,\star) \right] \lesssim r_{2,n}^2,
        \end{align}
        where $\star = \Pi$ for (i) label correction, and $\star = \widetilde\Pi$ for (ii) prior correction;
        \item[(m)] if $\Pi$ is not fully supported on $\mathcal P_n^\Pi$, the contribution outside the effective class is negligible at the same rate:
        \begin{align} \label{eq:prior-tail-risk}
            \E_{P\sim\Pi} \left[ \mathbbm{1}\{P\notin\mathcal P_n^\Pi\} \, \E_{\mathcal D_n\sim P^{\otimes n}} \norm{\tilde{\psi}-\psi(P)}_{L_2(\tilde q_{\theta^*},\mathcal D_n)}^2 \right] \lesssim n^{-1}+r_{2,n}^2. 
        \end{align}
    \end{itemize}
    
    Then, for either debiasing strategy,
    \begingroup\makeatletter\def\f@size{9}\check@mathfonts
    {\begin{align}
         \!\!\!\underbrace{\mathcal{R}_{l_2^2}(\hat{\psi}_{\theta^*}; \Pi,n)}_{\text{Bayes optimal risk, see Remark~\ref{rem:bayes-optimality-app}}} \!\!\!\!\!\!\! = \frac{1}{2} \mathcal{R}_{L_2^2}(\psi;\Pi,n) \le \mathcal{R}_{l_2^2}(\hat{\tilde{\psi}}_{\theta^*}; \Pi,n) &\le \mathcal{R}_{L_2^2}({\tilde{\psi}}; \Pi,n) \lesssim n^{-1}+r_{2,n}^2, \label{eq:bayes-non-optimality-app} \\[-2em] 
        \rotatebox{-90}{\textcolor{BrickRed}{$\le$}} \qquad & \qquad \,\,\, \rotatebox{-90}{\textcolor{BrickRed}{$\le$}} \nonumber \\[-0.4em] 
        \underbrace{\mathfrak M_n(\mathcal P_n^\Pi)}_{\text{minimax risk}} \le \mathcal{W}_{l_2^2}(\hat{\tilde{\psi}}_{\theta^*}; \mathcal P_n^\Pi,n) &\le \mathcal{W}_{L_2^2}({\tilde{\psi}}; \mathcal P_n^\Pi,n) \lesssim n^{-1}+r_{2,n}^2, \label{eq:minimax-optimality-app}
    \end{align}}
    \endgroup
    where the inequalities in \textcolor{BrickRed}{red} hold whenever
    $\Pi(\mathcal P_n^\Pi)=1$; $\mathcal{R}_{l_2^2}(\hat{{\psi}}; \Pi,n): = \E_{P\sim\Pi} \E_{\mathcal{D}_n\sim P^{\otimes n}}  [\hat{{\psi}}(\mathcal{D}_n) - \psi(P)]^2$, $\mathcal{R}_{L_2^2}({\tilde{\psi}}; \Pi,n) := \E_{P\sim\Pi} \E_{\mathcal{D}_n\sim P^{\otimes n}} \lvert\lvert{\tilde{\psi}-\psi}\rvert\rvert^2_{L_2(\tilde q_{\theta^*}, \mathcal{D}_n)}$ are $\Pi$-prior-averaged squared $l_2$ risks; $\mathcal{W}_{l_2^2}(\hat{{\psi}}; \mathcal P_n^\Pi,n) := \sup_{P \in \mathcal P_n^\Pi} \E_{\mathcal{D}_n\sim P^{\otimes n}}[\hat{{\psi}}(\mathcal{D}_n) - \psi(P)]^2$  and $\mathcal{W}_{L_2^2}({\tilde{\psi}}; \mathcal P_n^\Pi,n):=\sup_{P \in \mathcal P_n^\Pi} \E_{\mathcal{D}_n\sim P^{\otimes n}} \rvert\rvert{\tilde{\psi}-\psi}\rvert\rvert^2_{L_2(\tilde q_{\theta^*}, \mathcal{D}_n)}$ are worst-case squared $l_2$ risks; and $\mathfrak M_n(\mathcal P_n^\Pi):= \inf_{\hat{\psi}: \mathcal{Z}^{n} \to \Psi} \mathcal{W}_{l_2^2}(\hat{{\psi}}; \mathcal P_n^\Pi,n)$ denotes the minimax risk over $\mathcal P_n^\Pi$.

    If, additionally: 
    \begin{itemize}[leftmargin=0.6cm,itemsep=-0.55mm]
        \item[(n)] a matching minimax risk is available as 
        \begin{align} \label{eq:minimax-lower-rate} 
            \mathfrak M_n(\mathcal P_n^\Pi) \gtrsim n^{-1}+r_{2,n}^2;
        \end{align} 
    \end{itemize}
    then the debiased PFN posterior mean is minimax-rate optimal: 
    \begin{align} \label{eq:minimax-rate-optimal} 
        \mathcal W_{l_2^2}( \hat{\tilde{\psi}}_{\theta^*} ; \mathcal P_n^\Pi,n) \asymp \mathfrak M_n(\mathcal P_n^\Pi) \asymp n^{-1}+r_{2,n}^2. 
    \end{align}
\end{numtheorem}
\begin{proof}
    We first relate the squared error of the debiased PFN posterior mean to its posterior $L_2$-concentration error. For every fixed $P$ and $\mathcal D_n$, Jensen's inequality gives
    \begin{align} \label{eq:jensen-risk-proof}
        \left[\hat{\tilde{\psi}}_{\theta^*}(\mathcal D_n) - \psi(P) \right]^2 &= \left[ \int \big(\tilde{\psi}-\psi(P) \big) \tilde q_{\theta^*} (\tilde{\psi}\mid\mathcal D_n) \diff{\tilde{\psi}} \right]^2 \\ 
        &\le \int \big(\tilde{\psi}-\psi(P)\big)^2 \tilde q_{\theta^*} (\tilde{\psi}\mid\mathcal D_n) \diff{\tilde{\psi}} = \norm{ \tilde{\psi}-\psi(P)}_{L_2(\tilde q_{\theta^*},\mathcal D_n)}^2.
    \end{align}
    
    Taking expectations and either averaging over $P\sim\Pi$ or taking the supremum over $P\in\mathcal P_n^\Pi$ therefore yields
    \begin{align}
        \mathcal R_{l_2^2} (\hat{\tilde{\psi}}_{\theta^*}; \Pi,n) &\le \mathcal R_{L_2^2}(\tilde{\psi}; \Pi,n), \quad 
        \mathcal W_{l_2^2} (\hat{\tilde{\psi}}_{\theta^*}; \mathcal P_n^\Pi,n) \le
        \mathcal W_{L_2^2} (\tilde{\psi}; \mathcal P_n^\Pi,n).
    \end{align}
    
    Next, Theorem~\ref{thrm:debiased-rates-app} gives, for every $P\in\mathcal P_n^\Pi$, 
    \begin{align} 
        \norm{ \tilde{\psi}-\psi(P) }_{L_2(\tilde q_{\theta^*},\mathcal D_n)} &\le \widetilde F_\psi( P,\mathcal D_n,\star ) + R_{2,\psi}( P,\mathcal D_n,\star ). 
    \end{align} 
    Using $(a+b)^2\le2a^2+2b^2$, 
    \begin{align} 
        \norm{ \tilde{\psi}-\psi(P) }_{L_2(\tilde q_{\theta^*},\mathcal D_n)}^2 &\le 2 \widetilde F_\psi^2( P,\mathcal D_n,\star ) + 2 R_{2,\psi}^2( P,\mathcal D_n,\star ). 
    \end{align}
    
    Taking expectation over $\mathcal D_n\sim P^{\otimes n}$ and then the supremum over $P\in\mathcal P_n^\Pi$, Eq.~\eqref{eq:uniform-first-second} implies 
    \begin{align} 
        \mathcal W_{L_2^2} ( \tilde{\psi} ; \mathcal P_n^\Pi,n) &\lesssim n^{-1}+r_{2,n}^2. \label{eq:worst-case-upper-proof} 
    \end{align} 
    
    Together with Eq.~\eqref{eq:jensen-risk-proof}, this proves the two upper bounds in Eq.~\eqref{eq:minimax-optimality-app}.
    
    For the prior-averaged risk, split the expectation according to whether $P$ lies in the effective class: 
    \begin{align} 
        \mathcal R_{L_2^2}(\tilde{\psi}; \Pi,n) &= \E_{P\sim\Pi} \left[ \mathbbm{1}\{P\in\mathcal P_n^\Pi\} \E_{\mathcal D_n\sim P^{\otimes n}} \norm{ \tilde{\psi}-\psi(P) }_{L_2(\tilde q_{\theta^*},\mathcal D_n)}^2 \right] \nonumber\\ 
        &\qquad+ \E_{P\sim\Pi} \left[ \mathbbm{1}\{P\notin\mathcal P_n^\Pi\} \E_{\mathcal D_n\sim P^{\otimes n}} \norm{ \tilde{\psi}-\psi(P) }_{L_2(\tilde q_{\theta^*},\mathcal D_n)}^2 \right]. 
    \end{align} 
    The first term is bounded by $\mathcal W_{L_2}(\tilde{\psi}; \mathcal P_n^\Pi,n)$ and hence by Eq.~\eqref{eq:worst-case-upper-proof}; the second is controlled by Eq.~\eqref{eq:prior-tail-risk}. Therefore, 
    \begin{align} 
        \mathcal R_{L_2^2} ( \tilde{\psi}; \Pi,n) \lesssim n^{-1}+r_{2,n}^2. 
    \end{align} 
    Together with Eq.~\eqref{eq:jensen-risk-proof}, this establishes the upper bounds in Eq.~\eqref{eq:bayes-non-optimality}. 

    By posterior self-consistency, $P_0\mid\mathcal D_n \sim\Pi(\cdot\mid\mathcal D_n)$. Hence,
    Hence,
    \begin{align}
        \mathcal R_{L_2^2}(\psi; \Pi,n) &= \E_{P_0\sim\Pi}\E_{\mathcal D_n\sim P_0^{\otimes n}} \left[\int(\psi(P)-\psi(P_0))^2\, \Pi(\diff P\mid\mathcal D_n)\right] \\
        &= \E_{P_0\sim\Pi}\E_{\mathcal D_n\sim P_0^{\otimes n}} \left[\iint(\psi(P)-\psi(P'))^2\, \Pi(\diff P\mid\mathcal D_n) \Pi(\diff P'\mid\mathcal D_n)\right] \\
        &= 2\E_{P_0\sim\Pi}\E_{\mathcal D_n\sim P_0^{\otimes n}} \left[\E_\Pi[\psi(P)^2\mid\mathcal D_n] -\E_\Pi[\psi(P)\mid\mathcal D_n]^2 \right] \\
        &= 2\E_{P_0\sim\Pi}\E_{\mathcal D_n\sim P_0^{\otimes n}} \left[\Var_\Pi[\psi(P)\mid\mathcal D_n]\right] \\
        &= 2\, \mathcal R_{l_2^2} (\hat\psi_{\theta^*}; \Pi,n).
    \end{align}

    By Remark~\ref{rem:bayes-optimality-app}, $\hat{\psi}_{\theta^*}$ is the Bayes estimator under the original prior $\Pi$ and squared $l_2$ loss. Since $\hat{\tilde{\psi}}_{\theta^*}$ is another measurable estimator based on $\mathcal D_n$, 
    \begin{align} 
        \mathcal R_{l_2^2} ( \hat{\psi}_{\theta^*}; \Pi,n) &\le \mathcal R_{l_2^2} ( \hat{\tilde{\psi}}_{\theta^*}; \Pi,n), 
    \end{align} which proves the first inequality in Eq.~\eqref{eq:bayes-non-optimality-app}.
    
    Likewise, by definition of the minimax risk, 
    \begin{align} 
        \mathfrak M_n( \mathcal P_n^\Pi ) &= \inf_{\hat{\psi}} \mathcal W_{l_2^2} ( \hat{\psi}; \mathcal P_n^\Pi,n) \le \mathcal W_{l_2^2}( \hat{\tilde{\psi}}_{\theta^*} ; \mathcal P_n^\Pi,n), 
    \end{align} establishing the first inequality in Eq.~\eqref{eq:minimax-optimality-app}.
    
    If $\Pi(\mathcal P_n^\Pi)=1$, then for any non-negative risk function $g(P)$, \begin{align} 
        \E_{P\sim\Pi}[g(P)] &\le \sup_{P\in\mathcal P_n^\Pi}g(P). 
    \end{align} 
    Applying this identity to the point-estimation and posterior $L_2$ risks gives the two \textcolor{BrickRed}{red} inequalities. 
    
    Finally, if the matching minimax risk in Eq.~\eqref{eq:minimax-lower-rate} holds, then 
    \begin{align} 
        n^{-1}+r_{2,n}^2 &\lesssim \mathfrak M_n( \mathcal P_n^\Pi ) \le \mathcal W_{l_2^2}( \hat{\tilde{\psi}}_{\theta^*} ; \mathcal P_n^\Pi,n) \lesssim n^{-1}+r_{2,n}^2, 
    \end{align} which proves Eq.~\eqref{eq:minimax-rate-optimal}.
    
\end{proof}

%%%%%%%%%%%%%%%%%%%%%%%%%%%%%%%%%%%%%%%%%%%%%%%%%%%%%%%%%%%%%%%%%%%%%%%%%%%%%%%%%%%%%%%%%%
\newpage
\section{Matching Minimax Risks for Examples~\ref{ex:a}--\ref{ex:c}} \label{app:minimax-examples}
%%%%%%%%%%%%%%%%%%%%%%%%%%%%%%%%%%%%%%%%%%%%%%%%%%%%%%%%%%%%%%%%%%%%%%%%%%%%%%%%%%%%%%%%%%

We briefly verify when the matching lower-bound condition in Theorem~\ref{thrm:minimaxa-app} holds for our running examples. Structure-agnostic lower bounds are typically formulated over local classes defined through the accuracy of black-box nuisance estimators \citep{balakrishnan2026fundamental,jin2025structure,jin2025sharp}. Since our effective class additionally intersects such a class with the support of the PFN prior, these results apply provided that $\mathcal P_n^\Pi$ contains the corresponding hard structure-agnostic sub-model. More precisely, if $\mathcal H_n^\psi$ denotes a hard sub-model from one of the lower-bound constructions below and
\begin{align}
    \mathcal H_n^\psi \subseteq \mathcal P_n^\Pi,
\end{align}
then, by monotonicity of the minimax risk,
\begin{align}
    \mathfrak M_n(\mathcal P_n^\Pi) \ge \mathfrak M_n(\mathcal H_n^\psi).
\end{align}
Thus, the results below should be understood subject to this mild prior-support requirement.

\begin{examplea}[Mean value]
    For $\psi(P)=\E_P [Z]$, there is no non-parametric nuisance estimation and $R_{2,\psi}\equiv0$, such that $r_{2,n}=0$. The structure-agnostic restriction is therefore vacuous. If $\mathcal P_n^\Pi$ contains a regular one-dimensional sub-model with non-zero variance (for example, the Bernoulli family with success probabilities in a neighborhood of $1/2$), the classical two-point or local asymptotic minimax lower bound gives \citep{van2000asymptotic}:
    \begin{align}
        \mathfrak M_n(\mathcal P_n^\Pi) \gtrsim n^{-1} = n^{-1}+r_{2,n}^2.
    \end{align}
    Hence, the $n^{-1}$ squared-risk rate obtained by the debiased PFN is minimax-rate optimal.
\end{examplea}

\begin{exampleb}[Mean value under MAR]
    Let $e_{\mu,n}$ and $e_{\pi,n}$ denote squared $L_2(P_X)$ radii for black-box estimators of the outcome regression and labeling score, respectively:
    \begin{align}
        \norm{\mu-\hat{\mu}_n}_{L_2(P_X)}^2 &\le e_{\mu,n}, & \norm{\pi-\hat{\pi}_n}_{L_2(P_X)}^2 &\le e_{\pi,n}.
    \end{align}
    Under strong overlap, the corresponding second-order remainder satisfies
    \begin{align}
        \abs{R_{2,\psi}(P,P_0)} &= \abs{P_0 \left[\frac{\pi-\pi_0}{\pi} (\mu-\mu_0)\right]} \lesssim \norm{\pi-\pi_0}_{L_2(P_0)} \norm{ \mu-\mu_0}_{L_2(P_0)}.
    \end{align}
    Consequently, over a structure-agnostic class with the above nuisance radii, the squared second-order contribution is of order
    \begin{align}
        r_{2,n}^2 \asymp e_{\mu,n}e_{\pi,n}.
    \end{align}

The MAR mean is a linear functional of $\mu$ with an affine regression score and Riesz representer proportional to $1/\pi$. Under strong overlap, the $L_2$ error of this Riesz representer is equivalent up to constants to the $L_2$ error of $\pi$. Hence, the affine-score structure-agnostic lower bound of \citet{jin2025sharp} yields
\begin{align}
    \mathfrak M_n(\mathcal P_n^\Pi) \gtrsim n^{-1} + e_{\mu,n}e_{\pi,n} \asymp n^{-1} + r_{2,n}^2.
\end{align}
This is the same characteristic ``parametric variance plus product of nuisance errors'' lower bound established for canonical mixed-bias functionals by \citet{balakrishnan2026fundamental}. Thus, whenever the corresponding hard local class is contained in $\mathcal P_n^\Pi$, the rate of our debiased PFN is structure-agnostic minimax optimal.
\end{exampleb}

\begin{examplec}[WATE]
    We first consider a fixed weight $w=w(x)$ and let
    \begin{align}
        \norm{\mu_a-\hat{\mu}_{a,n}}_{L_2(P_X)}^2 &\le e_{a,n}, \,  a\in\{0,1\}, & \norm{\pi-\hat{\pi}_n}_{L_2(P_X)}^2 &\le f_n.
    \end{align}

    For the unnormalized WATE, \citet{jin2025structure} establish a structure-agnostic lower bound of the form
    \begin{align}
        \mathfrak M_n^{\mathrm{WATE}} \gtrsim \max\{e_{0,n},e_{1,n}\} f_n \norm{w}_{L_\infty(P)}^2 + \frac{\norm{w}_{L_2(P)}^2}{n}.
    \end{align}
    Their result is stated as a quantile lower bound and therefore also implies the corresponding expected squared-risk lower bound. If $w$ is uniformly bounded and non-degenerate and $\bar w(P)=P[w(X)]$ is bounded away from zero, the same rate applies to our normalized WATE.

    In particular, one may restrict the lower-bound construction to a sub-model with fixed $P_X$, for which $\bar w(P)$ is constant. Absorbing weight-dependent constants gives
    \begin{align}
        \mathfrak M_n(\mathcal P_n^\Pi) \gtrsim n^{-1} + \max\{e_{0,n}, e_{1,n}\} f_n.
    \end{align}
    At the same time, the second-order remainder of the first-order corrected WATE contains products between propensity-score and outcome-regression errors, such that, up to constants,
    \begin{align}
        r_{2,n}^2 \asymp f_n \left(e_{0,n} + e_{1,n} \right) \asymp f_n \max\{e_{0,n}, e_{1,n}\}.
    \end{align}

    Thus,
    \begin{align}
        \mathfrak M_n(\mathcal P_n^\Pi) \gtrsim n^{-1} + r_{2,n}^2.
    \end{align}
    This result directly covers the ATE, $w\equiv1$, as a special case. \citet{jin2025structure} additionally establish the analogous structure-agnostic lower bound for the ATT, which corresponds to a normalized propensity-dependent weighting of the treatment effect.

    For the overlap-weighted ATE, $w(x;\pi)=\pi(x)(1-\pi(x))$, a complementary lower bound follows from \citet{balakrishnan2026fundamental}. Indeed, its numerator can be written exactly as the expected conditional covariance,
    \begin{align}
        P\left[\pi(X) \big(1-\pi(X)\big) \tau(X) \right] &= P\left[\operatorname{Cov}(A,Y \mid X)\right].
    \end{align}
    For squared $L_2$ nuisance radii $r_n$ and $s_n$ of the corresponding outcome regression and propensity score, respectively, \citet{balakrishnan2026fundamental} show that the structure-agnostic minimax risk for this functional satisfies
    \begin{align}
        \mathfrak M_n^{\mathrm{cov}} \gtrsim r_ns_n + n^{-1}.
    \end{align}
    Restricting to a hard sub-model on which $P[\pi(X)(1-\pi(X))]$ is fixed, or varies only at a smaller order, transfers the same lower bound to the normalized overlap-weighted ATE. Since its second-order remainder contains the same product-order nuisance contribution, $r_{2,n}^2\asymp r_ns_n$ on this sub-model, yielding again
    \begin{align}
        \mathfrak M_n(\mathcal P_n^\Pi) \gtrsim n^{-1} + r_{2,n}^2.
    \end{align}
\end{examplec}

Taken together, these lower bounds verify the matching-rate condition from Theorem~\ref{thrm:minimaxa-app} for Example~\ref{ex:a}, Example~\ref{ex:b}, and the canonical fixed-weight, ATE, and overlap-weighted instances of Example~\ref{ex:c}, whenever the effective prior-supported class contains the corresponding structure-agnostic hard sub-models. 

%%%%%%%%%%%%%%%%%%%%%%%%%%%%%%%%%%%%%%%%%%%%%%%%%%%%%%%%%%%%%%%%%%%%%%%%%%%%%%%%%%%%%%%%%%
\newpage
\section{Implementation Details \& Experimental Setup} \label{app:implementation-details}
%%%%%%%%%%%%%%%%%%%%%%%%%%%%%%%%%%%%%%%%%%%%%%%%%%%%%%%%%%%%%%%%%%%%%%%%%%%%%%%%%%%%%%%%%%

In the following, we provide the implementation details on prior datasets generation for training and evaluation, and on the implementation of PFNs. Code is available at \mbox{\url{https://anonymous.4open.science/r/Ortho-PFN-Anonymized-E4B7}}.

\subsection{Prior processes}

\subsubsection{Example C (ATE)}

\textbf{Training prior.} We adapt the prior of the TabPFN $\mathrm{v1}$ \citep{hollmann2023tabpfn} for $P(x)$ and $\pi(x)$, combined with Gaussian processes with a Mat\'ern kernel ($\beta = 1.5$) for $\mu_a(x)$. Specifically, for each prior draw $P \sim \Pi$, the covariates $X \in \R^{d_x}$ (with $d_x \in \{5, 10\}$) follow the cause distribution of a random draw from the released multilayer-perceptron-based structural causal model prior of TabPFN $\mathrm{v1}$, standardized on a reference sample (and we redraw pathological draws with the rejection rule of TabPFN $\mathrm{v1}$). Then, the propensity score is given by an independent draw from the same prior: its standardized output, scaled by $1.5$, is used as the logit of $\pi(x)$, and $\pi(x)$ is clipped to $[0.02, 0.98]$ (\ie, strong overlap with $\kappa = 0.02$). The outcome regressions are $\mu_0(x) = f(x)$ and $\mu_1(x) = f(x) + b + 1.5\, t(x)$, where $f$ and $t$ are independent unit-variance Gaussian process paths with a Mat\'ern kernel ($\beta = 1.5$, lengthscale $\sqrt{d_x}$, approximated with $4096$ random Fourier features) and $b \sim N(0, 5^2)$. Finally, $A \mid X \sim \operatorname{Bernoulli}(\pi(X))$ and $Y = \mu_A(X) + \varepsilon$ with $\varepsilon \sim N(0, 1)$. Hence, under the training prior $\Pi$, the propensity score is drawn independently of the outcome regressions. For every $P$, the ground-truth ATE and the efficient variance $V_{\mathrm{eff}}(P)$ are computed on a reference sample of $2^{20}$ covariate draws (or $2^{22}$, if the Monte Carlo standard error exceeds $5\%$ of the efficient standard deviation at $n = 2048$). We draw $69{,}632$ populations ($65{,}536$ for training and $4096$ for validation), each with one dataset of $4096$ rows; for $n < 4096$, the datasets are the first $n$ rows of the stored datasets, so the training data are nested across $n$.

\textbf{Evaluation prior.} The evaluation prior $\check \Pi$ is obtained from $\Pi$ by a targeted selection: For each population, the propensity logit is additionally shifted by $\pm 1.5\, \mu_0(x)$, where the sign alternates over populations (\ie, an equal mixture of both directions). Hence, the treatment assignment strongly depends on the outcome regression, which yields a substantially larger amount of observed confounding than under $\Pi$ (as we will see later in Fig.~\ref{fig:part1-exampleC-ATE}), while the support stays the same. For both the training and the evaluation priors, we draw $M = 200$ populations (disjoint from training and validation), with the same population and dataset seeds for $\Pi$ and $\check \Pi$, and $D = 200$ datasets of $4096$ rows per population (nested across $n$). The ground-truth ATE and $V_{\mathrm{eff}}(P_0)$ of each population are computed on a reference sample of $2^{21}$ covariate draws.

\subsubsection{Example A (Mean value)}

\textbf{Training prior.} Similarly to Fig.~\ref{fig:debiasing-example-a}, we use a logistic Gaussian process with a squared-exponential kernel ($\beta = \infty$), namely, $P(z) \propto \exp\{s_f f(z)\}$ on $[0, 1]$, where $f$ is a unit-variance Gaussian process path with lengthscale $0.2$ (approximated with $1024$ random Fourier features), and the amplitude $s_f = 0.1$ makes the prior over-smoothing. Every density is normalized on a grid of $4097$ points, and the datasets are sampled with the inverse cumulative distribution function. Hence, the ground-truth mean $\psi(P)$ and the variance $\Var_P(Z)$ are exact (up to the grid). Unlike Example~\ref{ex:c}, we draw fresh prior samples at every training update (\ie, without a saved training bank).

\textbf{Evaluation prior.} The worst-case evaluation prior $\check \Pi$ tilts the same paths along the EIF direction, namely, $\check P(z) \propto \exp\{s_f f(z) \pm \gamma (z - 1/2)\}$, where the sign alternates over populations, and $\gamma = 0.544$ equals three prior standard deviations of the slope of $s_f f$ along $z - 1/2$. Hence, the ground-truth $\check P$ is in the support of $\Pi$ but atypical for it (similarly to the rate-sharp worst-case perturbation $P_0 + t (z - 1/2)$ from Example~\ref{ex:a}), yet such datasets still occur under the training prior. For both the training and the evaluation priors, we draw $M = 100$ populations with $D = 100$ datasets of $1024$ rows (nested across $n$); the populations of $\Pi$ and $\check \Pi$ share the same paths $f$ and the same random numbers for sampling the datasets.

\subsection{PFN implementation}

\subsubsection{Example C (ATE)}

\textbf{Architecture.} All the PFNs share the same architecture: A transformer that embeds every scalar cell of $\mathcal{D}_n$ as a token and alternates attention across features and across rows, similar to TabPFN~v2 \citep{hollmann2025accurate}. A learned query row, which the observed rows do not attend to, is pooled across features and mapped to a $5$-component normal mixture density $q_\theta(\psi \mid \mathcal{D}_n)$. For $d_x=5$, we use $3$ layers, embedding dimension $64$, $4$ attention heads, and feed-forward width $128$ (hence, the number of trainable parameters totals to approx. $149$K). For $d_x=10$, we increased the capacity of the PFN ans set the number of layers to 3, the embedding dimension to $128$, the number of attention heads to $8$, and the feed-forward width $256$ (the number of trainable  parameters is approx. $594$K). Inputs and labels are normalized per dataset.

\textbf{Debiased PFN ((i) label correction).} The debiased PFN is trained with $K = 4$ independent Bayesian-bootstrap draws $(W_1, \dots, W_n) \sim \operatorname{Dir}(n; 1, \dots, 1)$ per dataset. Here, the labels $\tilde\psi(P)$ are the Bayesian-bootstrap averages of the A-IPTW pseudo-outcomes, computed with the ground-truth nuisance functions of each prior draw $P$. The final loss from Eq.~\eqref{eq:label-corr} then averages $K$ labels $\tilde\psi(P)$.

\textbf{Debiased PFN ((ii) prior correction).} We implement the prior correction of Example~\ref{ex:c} for the ATE ($w \equiv 1$, so that $\pi(x)$ is not perturbed) with a bounded version of the covariate tilt. Namely, we replace the exponential tilt of $P(x)$ with the weight $1 + \tanh(2 \lambda \tau(x))/2 \in [1/2, 3/2]$, which yields the same score at $\lambda = 0$ and allows exact sampling of the covariates via rejection sampling. The outcome regressions are perturbed as in Example~\ref{ex:c}, namely, $\tilde\mu_1(x) = \mu_1(x) + \lambda / \pi(x)$ and $\tilde\mu_0(x) = \mu_0(x) - \lambda / (1 - \pi(x))$ (as $\sigma_a^2(x) = 1$). We use $\lambda \sim N(0, \sigma_n^2 = n^{-1/2})$, with one standard normal draw per population that is shared across $n$. Hence, for every $n$, we generate a separate training bank with one $n$-row dataset per perturbed population $\tilde P$, and the training label is the ATE of $\tilde P$ (computed on the tilted reference sample).

\textbf{Training.} We train a separate network for every $n$ and method with AdamW (learning rate $3 \cdot 10^{-4}$, weight decay $10^{-5}$), batch size $31$ datasets, gradient-norm clipping at $1$, and mixed precision for $126{,}900$ updates ($60$ epochs). The learning rate is halved when the validation loss has not improved for $5$ validation cycles (every $2{,}115$ updates), and we keep the checkpoint with the lowest validation loss. All the methods use the same initialization, data order, and seed.

\textbf{Compute.} All the networks were trained on 2 NVIDIA H200 GPUs. For $d_x = 5$, one fit takes approximately $2$--$6.5$ hours for $n \leq 1024$ and approximately $15$ hours for $n = 2048$; for $d_x = 10$, approximately $2$--$8.5$ hours and $34$ hours, respectively (with three fits simultaneously sharing one GPU).

\subsubsection{Example A (Mean value)}

\textbf{Architecture.} We use the same architecture as for Example~\ref{ex:c} ($d_x = 5$), with a single input column $Z$ (here, the number of trainable  parameters is approx. $149$K). Unlike Example~\ref{ex:c}, inputs and labels are not normalized per dataset but transformed with fixed constants (namely, $Z - 1/2$ and $(\psi - 1/2)/0.02$, where $0.02$ is approximately the prior standard deviation of $\psi$), which are inverted at evaluation. This is necessary because the posterior standard deviations are small compared to $1/2$ (\ie, down to $0.009$ for $n = 1024$). At evaluation, we drop the mixture components with weights below $10^{-3}$.

\textbf{Debiased PFN ((i) label correction).} As for Example~\ref{ex:c}, the debiased PFN is trained with $K = 4$ independent Bayesian-bootstrap draws per dataset, where the labels are the Bayesian-bootstrap means $\tilde\psi(P) = \operatorname{BB}_n\{Z\}$ (Eq.~\eqref{eq:mean-label-corr}).

\textbf{Debiased PFN ((ii) prior correction).} We use the exponential tilting from Example~\ref{ex:a} with $\lambda \sim N(0, \sigma_n^2 = n^{-1/2})$, drawn independently for every prior draw. Here, the tilt is applied exactly on the grid (\ie, as an additional linear term of the log-density), and both the dataset and the training label $\psi(\tilde P)$ are taken from the tilted distribution $\tilde P$.

\textbf{Training.} We train a separate network for every $n \in \{16, 32, \dots, 1024\}$ and method with AdamW (learning rate $3 \cdot 10^{-4}$, weight decay $10^{-5}$), batch size $32$ datasets, gradient-norm clipping at $1$, and mixed precision for $12{,}000$ updates (with $500$ warm-up updates and a cosine decay of the learning rate). The validation loss is computed every $1{,}000$ updates on a fixed set of $512$ validation datasets, and we keep the checkpoint with the lowest validation loss.

\textbf{Compute.} All the networks were trained on 2 NVIDIA H200 GPUs. One fit takes approximately $12$--$50$ minutes, depending on $n$ (with two fits simultaneously sharing one GPU).

\subsection{Experimental Limitations} \label{app:exp-limitations}

Our experiments are designed to empirically verify our theory in a controlled setting where the ground-truth is known, rather than to benchmark against existing estimators. Hence, we compare the PFNs that differ only in the training priors / training labels and, thus, do not include frequentist estimators with estimated nuisance functions (\eg, A-IPTW or TMLE) or post-hoc corrections of PFN posteriors \citep{melnychuk2026frequentist}. For computational reasons, we train a single PFN of each type (\ie, standard, debiased with (i) label correction, and debiased with (ii) prior correction), so our error intervals are conditional on the trained networks. 
Finally, in our experiments, we only cover the ATE (Example~\ref{ex:c}) and the mean value (Example~\ref{ex:a}) estimation under a single synthetic prior $\Pi$, yet the former is sufficiently complex to empirically validate our theoretical results.

%%%%%%%%%%%%%%%%%%%%%%%%%%%%%%%%%%%%%%%%%%%%%%%%%%%%%%%%%%%%%%%%%%%%%%%%%%%%%%%%%%%%%%%%%%
\newpage
\section{Additional Results} \label{app:additional-results}
%%%%%%%%%%%%%%%%%%%%%%%%%%%%%%%%%%%%%%%%%%%%%%%%%%%%%%%%%%%%%%%%%%%%%%%%%%%%%%%%%%%%%%%%%%

This Appendix provides additional experimental results for the standard and our debiased PFNs.

\subsection{Example~\ref{ex:c} (ATE)} \label{app:additional-results-C}

\begin{figure}[H]
    \vspace{-0.4cm}
    \centering
    \includegraphics[width=\linewidth]{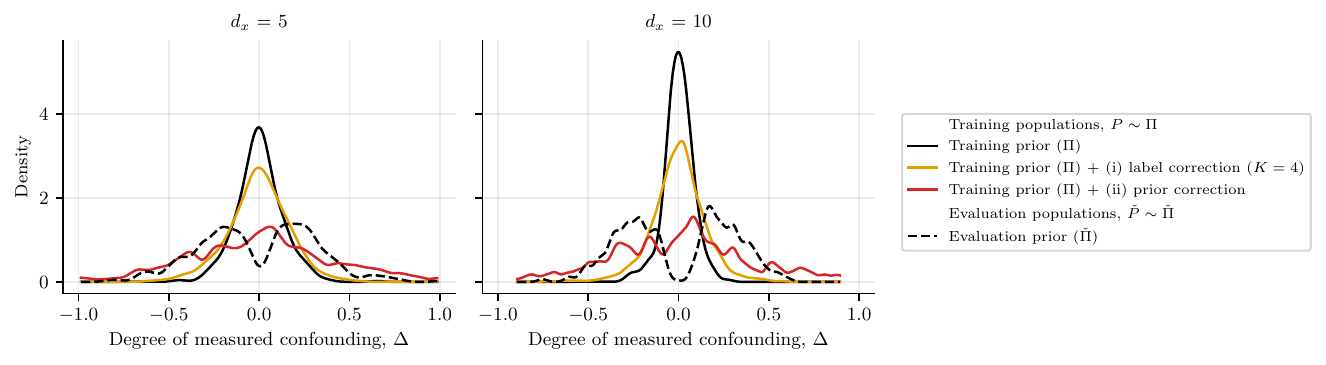}
    \vspace{-0.5cm}
    \caption{\textbf{Prior-induced confounding bias of the training targets for the ATE} (based on Example~\ref{ex:c} with $w \equiv 1$; see Remark~\ref{rem:pib}). Here, we show the densities of the standardized degree of observed confounding $\Delta$ (= differences between the training targets, $\diamond \in \{\psi(P), \tilde\psi(P), \psi(\tilde P)\}$, and the unadjusted treated--control contrast, $m(P) := \E_P[Y \mid A = 1] - \E_P[Y \mid A = 0]$, namely, $\Delta := \{\diamond - m\}/\mathrm{sd}[Y]$) depending on the covariate dimensionality $d_x$.
    The densities are estimated over $M = 512$ draws of ground-truth $P \sim \Pi$ from the original prior or their debiased counterparts: (i)~label correction, $\tilde\psi(P)$, with $D=1$ dataset $\mathcal{D}_n$ and $K=4$ Bayesian bootstrap draw per $P$; and (ii)~prior correction, $\psi(\tilde P)$ with $\tilde P = \operatorname{LF}_\psi(P; \lambda)$ and $\lambda \sim N(0, \sigma_n^2 = n^{-1/2})$ (both for $n = 256$; $m$ and $\mathrm{sd}[Y]$ are taken under $\tilde P$ for (ii)). For reference, we also show $\Delta$ under the worst-case evaluation prior $\check \Pi$.
    }
    \vspace{-0.5cm}
    \label{fig:part1-exampleC-ATE}
\end{figure}

\textbf{Results for part (1).} Fig.~\ref{fig:part1-exampleC-ATE} demonstrates that the worst-case evaluation prior $\check\Pi$ contains a substantially larger spread of the observed confounding than the training prior $\Pi$. Hence, a standard PFN trained on $\Pi$ implicitly assumes nearly unconfounded distributions $P$, while many ground-truth $\check P \sim \check\Pi$ have $\abs{\Delta(\check P)} \gg 0$. This means (by Remark~\ref{rem:pib}) that the first-order error of the standard PFN cannot vanish when evaluated at the worst-case prior. Then, (ii)~prior correction removes this mismatch at the level of the prior: the least-favorable perturbation spreads $\Delta$ at least as widely as under $\check\Pi$. In contrast, (i)~label correction only slightly increases the spread. Importantly, this is expected: (i)~does not modify the prior $\Pi$ but only the training label whose correction term has mean zero under $P$ and a standard deviation of order $n^{-1/2}$ (thus, its contribution to the spread vanishes for growing $n$). Instead, (i)~debiases the target through the data-dependent one-step correction, which removes the first-order error regardless of how much confounding is contained in $\Pi$.

\begin{figure}[H]
    \vspace{-0.6cm}
    \centering
    \includegraphics[width=\linewidth]{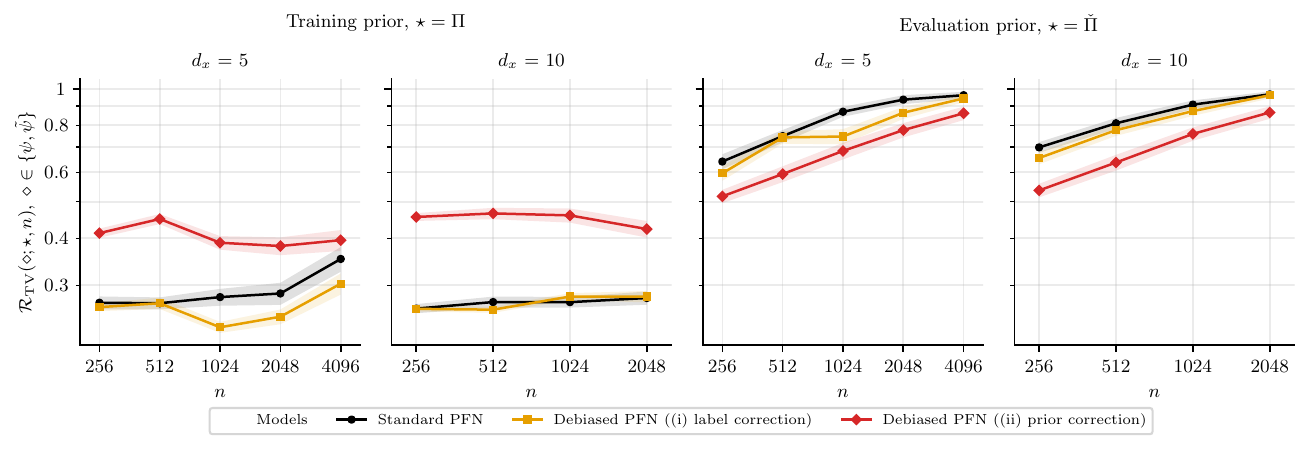}
    \vspace{-0.5cm}
    \caption{\textbf{Total variation, averaged over the training and evaluation priors, $\Pi$ and $\check \Pi$, respectively} (based on Example~\ref{ex:a} with $w \equiv 1$). Shown: mean $\pm$ 3SE $\Pi/\check \Pi$-averaged total variation between the PFN posterior and the asymptotic normal distribution given by the efficient frequentist estimator (A-IPTW w/ ground-truth nuisance functions), $\mathcal{R}_\mathrm{TV}(\diamond; \star,n)$, depending on dataset sizes $n$ and covariate dimensionality $d_x$ (over $M=200$ ground-truth $\check P$, each with $D=200$ datasets; lower is better; axes are log-scaled). Here, $\diamond \in \{\psi, \tilde\psi\}$ are the target posteriors from the standard and debiased PFNs, respectively.}
    \label{fig:part3-exampleC-ATE}
    \vspace{-0.3cm}
\end{figure}

\textbf{Results for part (3).} Fig.~\ref{fig:part3-exampleC-ATE} shows the results. Under the training prior $\Pi$, the total variation of all the PFNs stays approximately constant for growing $n$ (\ie, around $0.25$--$0.3$ for the standard PFN and (i)~label correction, and around $0.4$ for (ii)~prior correction). Hence, even under the training prior, the trained PFN posteriors remain at a non-vanishing distance from the efficient normal approximation for the considered $n$, which is likely due to amortization errors. Here, the larger distance of (ii)~prior correction is expected, as it is trained on the corrected prior $\widetilde \Pi$ rather than on $\Pi$ (and, hence, is not Bayes-optimal under $\Pi$; see Eq.~\eqref{eq:bayes-non-optimality}). Under the evaluation prior $\check \Pi$, on the other hand, the total variation grows with $n$ for all the PFNs (up to $0.86$--$0.97$ for the largest $n$): Here, the efficient normal approximation concentrates at the parametric rate, while the remaining errors of the PFN posteriors decay more slowly, so that both distributions increasingly separate. Nevertheless, our debiased PFNs are consistently closer to the efficient normal approximation than the standard PFN, in particular, (ii)~prior correction (\eg, $0.86$ vs. $0.96$ for $d_x = 5$ and $0.87$ vs. $0.97$ for $d_x = 10$ for the largest $n$), whereas (i)~label correction is only slightly closer. This is in line with Theorem~\ref{thrm:bbvm}, which additionally requires a negligible second-order remainder $R_{2,\psi}$: Under the targeted selection of $\check \Pi$, the PFNs have to implicitly infer propensity scores that are atypical for $\Pi$, and, hence, the second-order remainder (together with the amortization errors) does not vanish fast enough for the considered $n$.

\subsection{Example~\ref{ex:a} (Mean Value)} \label{app:additional-results-A}

In the following, we report the experimental results of estimating the mean value (Example~\ref{ex:a}) with the standard and our debiased PFNs: We show the results for parts (2) and (3). Note, that there is no natural interpretation of the prior-induced confounding bias in the case of the mean value, hence part (1) is omitted.  We provide the corresponding implementation details in Appendix~\ref{app:implementation-details}.

\begin{figure}[H]
    \vspace{-0.4cm}
    \centering
    \includegraphics[width=0.75\linewidth]{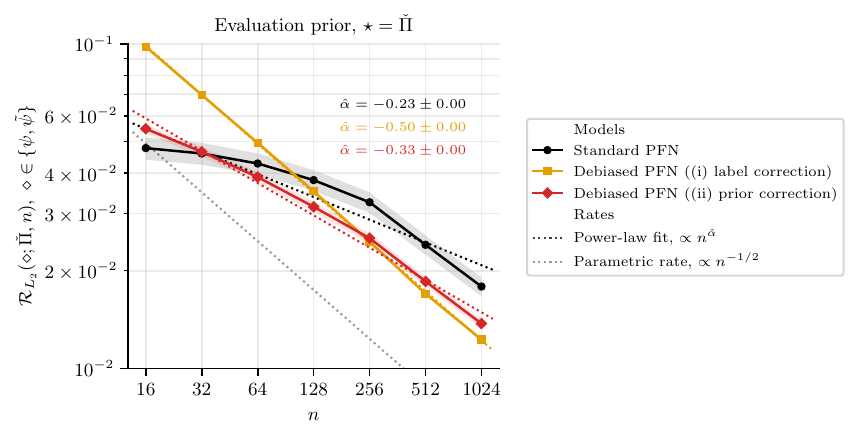}
    \vspace{-0.5cm}
    \caption{\textbf{$L_2$ risk on the evaluation prior $\check \Pi$} (based on Example~\ref{ex:a}). Shown: mean $\pm$ 3SE $\check \Pi$-averaged $L_2$ concentration errors (solid lines) and the corresponding power-law regression fits
    (dotted lines) depending on dataset sizes $n$ (over $M=200$ ground-truth $\check P$, each with $D=200$ datasets; lower is better; axes are log-scaled). Here, $\diamond \in \{\psi, \tilde\psi\}$ are the target posteriors from the standard and debiased PFNs, respectively.}
    \label{fig:part2-exampleA-mean}
    \vspace{-0.3cm}
\end{figure}

\textbf{Results for part (2).} Fig.~\ref{fig:part2-exampleA-mean} demonstrates the results. Therein, the standard PFN concentrates \emph{markedly slower than the parametric rate} ($\hat{\alpha} = -0.47$). Similarly to Fig.~\ref{fig:debiasing-example-a}, the oversmoothing training prior $\Pi$ pulls the posterior of $\psi$ towards $1/2$, and this first-order error does not vanish for the worst-case ground-truth $\check P \sim \check \Pi$ (as suggested by Theorem~\ref{thrm:slow-rates}). In contrast, (i)~label correction recovers the parametric rate ($\hat{\alpha} = -1.01$), and (ii)~prior correction substantially speeds up the concentration ($\hat{\alpha} = -0.66$). For small $n$ (\ie, $n \le 64$), the standard PFN still achieves a lower risk, as the informative training prior $\Pi$ dominates its posterior. Yet, for sufficiently large $n$ (\ie, $n \ge 128$), both debiased PFNs outperform the standard PFN, and the gap grows with $n$ (\eg, at $n = 1024$, the risk of the standard PFN is more than twice that of (i)~label correction). The remaining gap between (ii) and (i) is due to the scale of the perturbation: For the mean value, $\sigma_n^2 = n^{-1/2}$ is small relative to the variance of $Z$, and, hence, the corrected prior $\widetilde \Pi$ still mildly pulls the posterior towards $1/2$. Overall, this confirms that both debiasing strategies speed up the target concentration rates at the worst-case evaluation prior $\check \Pi$ (as suggested by Theorems~\ref{thrm:debiased-rates} and \ref{thrm:minimax}).

\begin{figure}[H]
    \vspace{-0.3cm}
    \centering
    \includegraphics[width=0.9\linewidth]{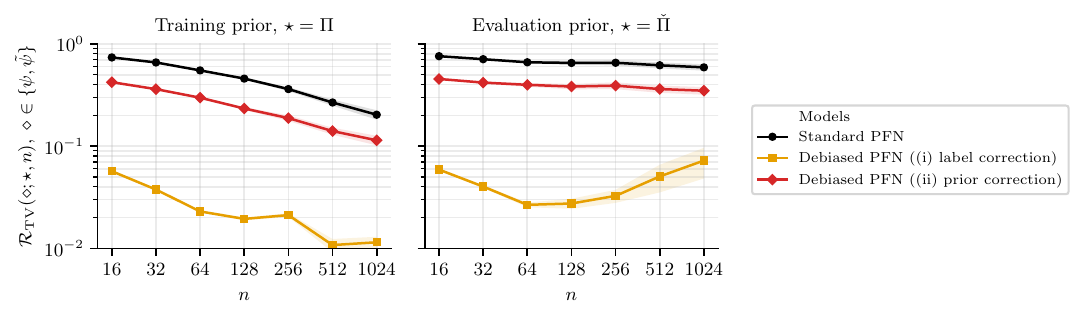}
    \vspace{-0.5cm}
    \caption{\textbf{Total variation, averaged over the training and evaluation priors, $\Pi$ and $\check \Pi$, respectively} (based on Example~\ref{ex:a}). Shown: mean $\pm$ 3SE $\Pi/\check \Pi$-averaged total variation between the PFN posterior and the asymptotic normal distribution given by the efficient frequentist estimator (sample mean), $\mathcal{R}_\mathrm{TV}(\diamond; \star,n)$, depending on dataset sizes $n$ (over $M=200$ ground-truth $\check P$, each with $D=200$ datasets; lower is better; axes are log-scaled). Here, $\diamond \in \{\psi, \tilde\psi\}$ are the target posteriors from the standard and debiased PFNs, respectively.}
    \label{fig:part3-exampleA-mean}
    \vspace{-0.3cm}
\end{figure}

\textbf{Results for part (3).} Fig.~\ref{fig:part3-exampleA-mean} shows the results. Here, the posterior of (i)~label correction is close to the efficient normal approximation of the sample mean for all $n$, both under the training and under the evaluation prior (\ie, the total variation is at most $0.07$), as suggested by Theorem~\ref{thrm:bbvm}. In contrast, the posterior of the standard PFN deviates substantially. Under the training prior $\Pi$, its total variation decreases with $n$ (from $0.74$ to $0.20$), as its posterior is initially much narrower than the efficient one due to the informative prior. Yet, under the evaluation prior $\check\Pi$, it barely decreases (from $0.76$ to $0.59$), as the posterior remains shifted towards $1/2$ (\ie, the first-order error from part (2)). Then, (ii)~prior correction lies in between: Its total variation decreases under both priors, but more slowly under $\check\Pi$ (from $0.45$ to $0.35$), in line with the remaining pull towards $1/2$ from part (2). Finally, the slight increase of the total variation of (i)~label correction under $\check\Pi$ for $n \ge 512$ is likely due to amortization errors, as datasets from $\check\Pi$ are atypical for the training prior $\Pi$.

\end{document}